\documentclass[sigconf]{acmart}
\usepackage{algorithm}
\usepackage{algorithmic}
\usepackage{subcaption}
\RequirePackage{multirow} 
\RequirePackage{tabularx}
\renewcommand\footnotetextcopyrightpermission[1]{}
\AtBeginDocument{%
  \providecommand\BibTeX{{%
    \normalfont B\kern-0.5em{\scshape i\kern-0.25em b}\kern-0.8em\TeX}}}
\acmConference[]{April}{2023}{Toward Scalable Image Feature Compression}
\acmPrice{15.00}
\acmISBN{978-1-4503-XXXX-X/18/06}
\begin{document}
\title{Toward Scalable Image Feature Compression: A Content-Adaptive and Diffusion-Based Approach}
 
\author{Sha Guo}
\orcid{0009-0008-9111-4084}
\affiliation{%
  \institution{School of Computer Science}
  \institution{Peking University}
    \institution{Peng Cheng Laboratory}
  \city{Beijing}
  \country{China}
}
\email{sandykwokcs@stu.pku.edu.cn}

\author{Zhuo Chen}
\authornotemark[1]
\orcid{0000-0003-0563-1760}
\affiliation{%
  \institution{Peng Cheng Laboratory}
  \city{Shenzhen}
  \country{China}
}
\email{chenzh08@pcl.ac.cn}

\author{Yang Zhao}
\orcid{0000-0002-4032-8049}
\affiliation{
  \institution{School of Computer and Information}
  \institution{Hefei University of Technology}
  \city{Hefei}
  \country{China}
}
\email{yzhao@hfut.edu.cn}

\author{Ning Zhang}
\orcid{0000-0003-3985-4305}
\affiliation{%
  \institution{ School of Computer Science}
  \institution{Peking University}
    \city{Beijing}
  \country{China}
}
\email{zhangn77@pku.edu.cn}

\author{Xiaotong Li}
\orcid{0000-0001-8219-4176}
\affiliation{%
  \institution{School of Computer Science}
  \institution{Peking University}
    \city{Beijing}
  \country{China}
}
\email{lixiaotong@stu.pku.edu.cn}

\author{Lingyu Duan}
\authornote{ Corresponding author. }

\orcid{0000-0002-4491-2023}
\affiliation{%
  \institution{School of Computer Science}
  \institution{Peking University}
  \institution{Peng Cheng Laboratory}
  \city{Beijing}
  \country{China}
}
\email{lingyu@pku.edu.cn}

\renewcommand{\shortauthors}{Sha Guo et al.}

\begin{abstract}
Traditional image codecs emphasize signal fidelity and human perception, often at the expense of machine vision tasks. Deep learning methods have demonstrated promising coding performance by utilizing rich semantic embeddings optimized for both human and machine vision. However, these compact embeddings struggle to capture fine details such as contours and textures, resulting in imperfect reconstructions. Furthermore, existing learning-based codecs lack scalability.
To address these limitations, this paper introduces a content-adaptive diffusion model for scalable image compression. The proposed method encodes fine textures through a diffusion process, enhancing perceptual quality while preserving essential features for machine vision tasks. The approach employs a Markov palette diffusion model combined with widely used feature extractors and image generators, enabling efficient data compression. By leveraging collaborative texture-semantic feature extraction and pseudo-label generation, the method accurately captures texture information. A content-adaptive Markov palette diffusion model is then applied to represent both low-level textures and high-level semantic content in a scalable manner.
This framework offers flexible control over compression ratios by selecting intermediate diffusion states, eliminating the need for retraining deep learning models at different operating points. Extensive experiments demonstrate the effectiveness of the proposed framework in both image reconstruction and downstream machine vision tasks such as object detection, segmentation, and facial landmark detection, achieving superior perceptual quality compared to state-of-the-art methods.
\end{abstract}





\maketitle
\section{Introduction}
\begin{figure}[ht]
  \centering
\includegraphics[width=\linewidth]{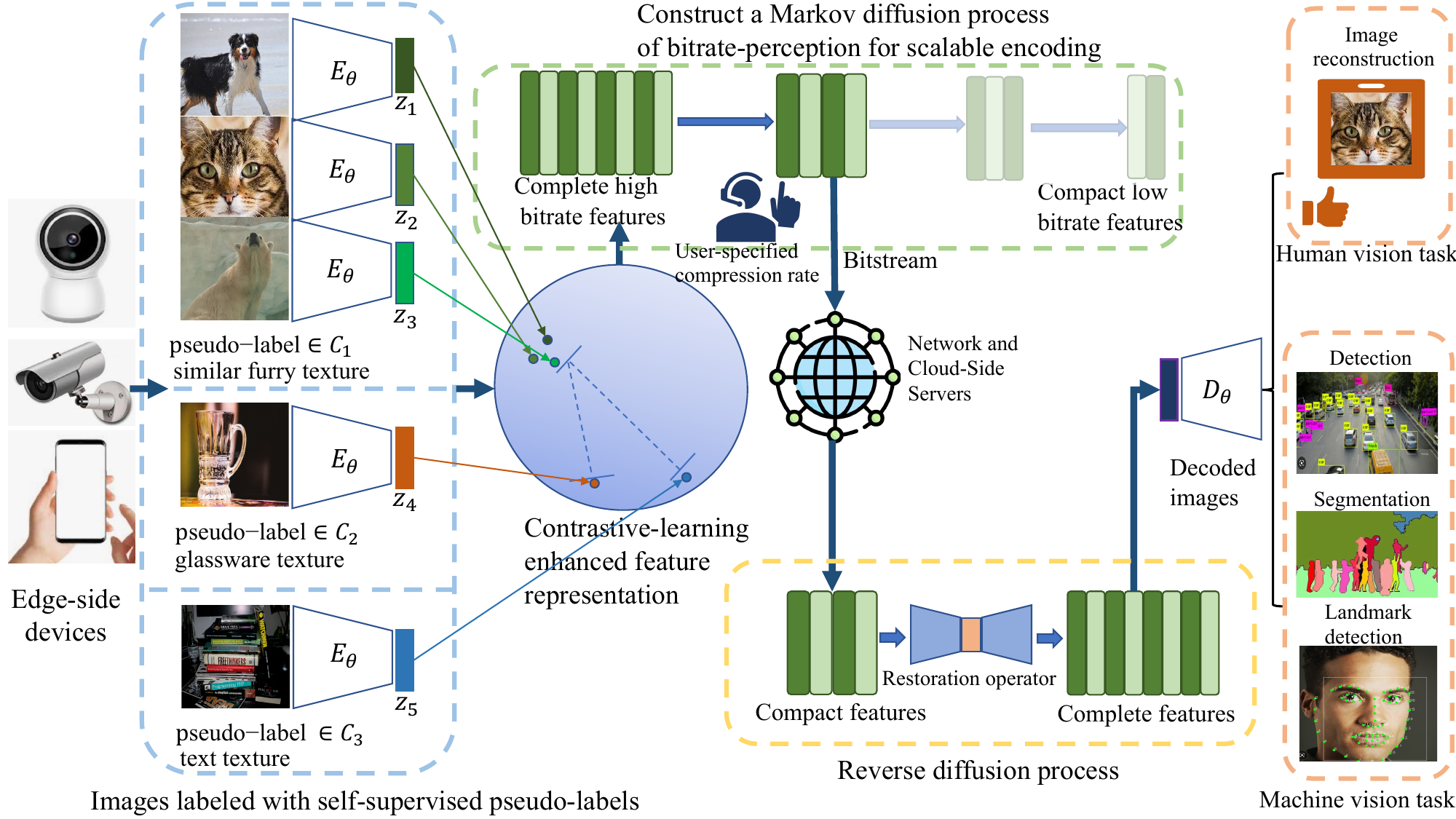}
  \caption{Our feature compression-transmission-decode-analysis paradigm: Features are extracted and compressed at front-end devices according to user-defined compression rates, with decompression and vision tasks carried out at the server side. }
\vspace{-0.4cm}
\end{figure}
In the era of big data, the vast amount of images and videos has posed significant challenges in terms of storage, transmission, and analysis. As the foundation of the compress-then-analyze paradigm \cite{duan2020video}, image and video compression techniques aim to balance bit-rate efficiency with perceptual quality for both human and machine vision.

Traditional compression methods, such as MPEG-4 AVC/H.264 \cite{wiegand2003overview}, High Efficiency Video Coding (HEVC) \cite{sullivan2012overview}, Versatile Video Coding (VVC) \cite{bross2021overview}, and Audio Video Coding Standards (AVS) \cite{ma2015avs2}, have significantly improved video coding efficiency by exploiting spatial-temporal pixel redundancy in video frames based on visual signal statistics and human perception priors. However, these methods, optimized for signal fidelity and low-level image characteristics (\textit{e.g.}, contours, edges, colors), often overlook semantic information, which limits their performance in machine vision tasks.

Recent advances in deep learning-based video coding \cite{minnen2017spatially, minnen2018joint, balle2018variational, xiao2023invertible, mentzer2020high, xie2021enhanced, he2022elic} have shown significant progress by leveraging deep feature representations and large-scale data priors. These approaches utilize hierarchical model architectures and deep-network-aided coding tools that can surpass traditional codecs. 

\begin{figure}  
\includegraphics[width=0.4\textwidth]{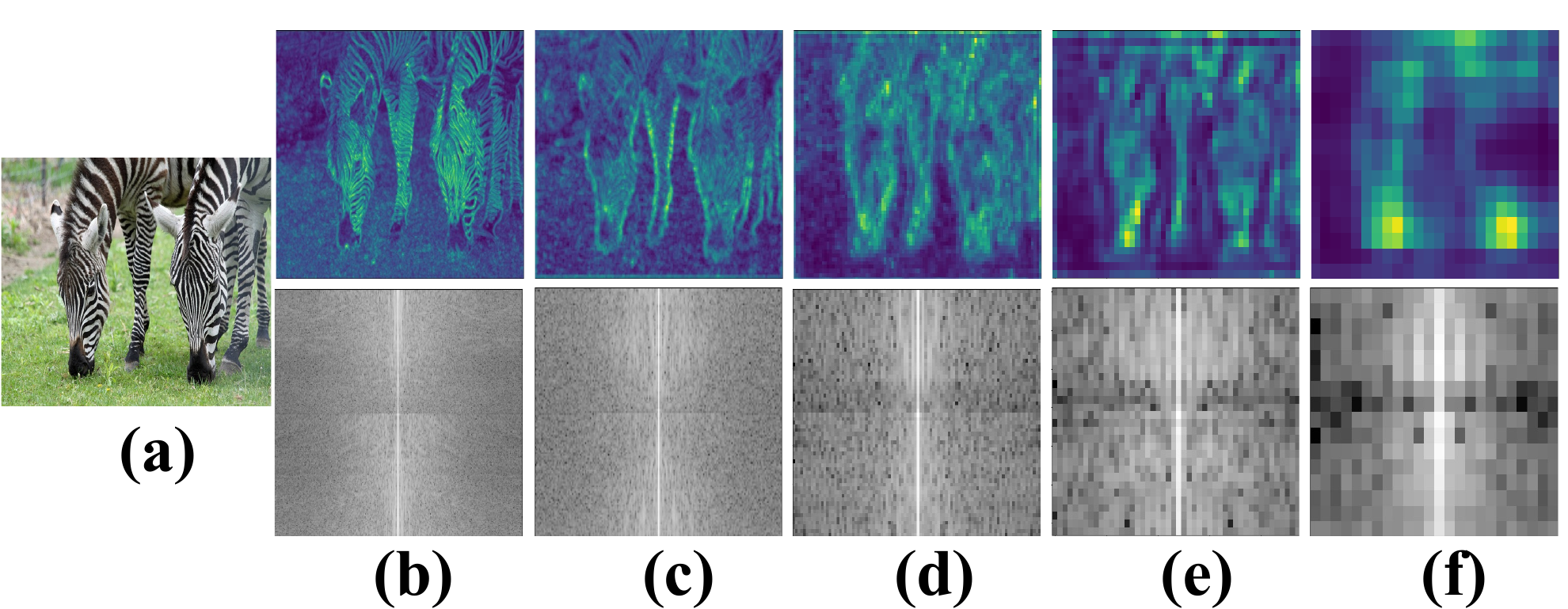}
\caption{The VGG \cite{simonyan2014very} decomposition of the "zebra" image: (a) Original image. (b)-(f) represent feature maps of \textbf{$conv1^{(2)}$}, \textbf{$conv2^{(2)}$}, \textbf{$conv3^{(3)}$}, \textbf{$conv4^{(3)}$}, \textbf{$conv5^{(3)}$} with their Fast Fourier Transform (FFT) \cite{bracewell1986fourier} analysis.}
  \label{fig:zebra}
\vspace{-0.5cm}
\end{figure}

However, as shown in Fig. \ref{fig:zebra} (f), deep learning representations tend to capture rich semantic information but often fail to preserve low-level details such as textures, edges, and contours, leading to visual artifacts in the decoded images. This limitation reduces their ability to accurately represent image patterns and compromises visual quality.

Moreover, CNN-based methods typically rely on an encoder-decoder architecture, where the encoder compresses the input into a lower-dimensional latent space, and the decoder reconstructs the image from this compressed representation in a single, deterministic step \cite{kingma2013auto}. In contrast, diffusion models introduce a stochastic process with a sequence of gradual transitions from noise to a fully reconstructed image, allowing the capture of more complex data distributions \cite{sohl2015deep, ho2020denoising, goose2023neural, bansal2022cold}.

Most deep-learning-based compression methods \cite{xiao2023invertible, mentzer2020high, minnen2018joint, xie2021enhanced, he2022elic} also face challenges with bitrate control. To support multiple trade-offs between bit-rate consumption and reconstruction quality, these methods often require training separate models for each bit rate, which limits their scalability and increases storage and computational demands. Choi et al. \cite{choi2019variable} introduced a conditional autoencoder framework that incorporates rate control parameters such as the Lagrange multiplier and quantization bin size, offering a more adaptive rate control mechanism.

In light of these limitations, this paper proposes a content-adaptive, diffusion-based compression framework that achieves strong performance for both human and machine vision tasks, with flexible operating points. The content-adaptive approach jointly analyzes coarse semantic information and fine-grained spectral texture details for self-supervised clustering, which generates pseudo-labels for image patches. Building on recent research \cite{deng2020disentangled, lee2021infomax, liu2021divco, park2020contrastive, wang2021dense, zhou2021cdl} that highlights the benefits of contrastive learning in generative vision tasks by aligning texture and semantic perceptual spaces, these pseudo-labels are used to train the feature extraction network with contrastive learning.

To enable efficient compression and reconstruction in the latent feature space, we design a diffusion-based scalable image feature compression method. During the compression process, content-adaptive hierarchical palettes form a Markov diffusion chain, allowing the compression ratio to increase while maintaining perceptual quality. In the reverse diffusion process, compact features are iteratively refined to reconstruct the full image.

The key contributions of this work include:

1) A Markov palette diffusion method for compressing image features in latent space, where a hierarchical $K$-means \cite{liu2020determine} clustering process enables gradual color distortion during compression and high-quality feature regeneration during reconstruction.

2) A feature extraction method that embeds discriminative semantic and texture information into the latent feature space, leveraging a coarse-to-fine collaboration and contrastive learning with clustered pseudo-labels in the frequency domain.

3) An efficient scalable coding mechanism that allows for compression at variable operating points without the need to train multiple deep learning models. Extensive experiments demonstrate the superiority of the proposed method in terms of both human visual perception and machine vision tasks.
\section{Related Work}

\textbf{Traditional Codecs}. Traditional codecs such as JPEG \cite{wallace1992jpeg} and WebP \cite{mukherjee2014webp} are widely used for image compression. JPEG employs the discrete cosine transform (DCT) to achieve compression ratios of up to 1:20 with minimal visual degradation. WebP, developed by Google, offers a 26\% reduction in bitrate compared to JPEG while maintaining image quality \cite{salomon2012webp}. In video compression, the H.265/HEVC and H.266/VVC standards have emerged, further improving compression efficiency. H.265 provides a 22\% improvement over H.264 \cite{sullivan2012overview, lainema2012intra}, while H.266 offers an additional 25\% improvement over H.265 \cite{bross2021overview, pfaff2021intra}. However, these traditional codecs focus primarily on pixel-level compression and often neglect higher-level semantic information, resulting in artifacts such as blockiness and blurring due to over-quantization and filtering. In contrast, our method considers both low-level textures and high-level semantic information during compression, enabling accurate texture reconstruction and preserving detailed contour shapes.

\noindent \textbf{Deep Learning Compression}. Previous approaches \cite{li2017convolutional, sun2020learned} proposed using downsampling blocks before applying normal intra coding, followed by upsampling to restore the original resolution. Image resampling techniques have been further refined through invertible flow-based encoding and generation \cite{xiao2023invertible, liang2021hierarchical}. More recently, significant progress has been made in end-to-end neural codecs for image and video compression. Several studies \cite{balle2018variational, mentzer2020high, he2022elic, xie2021enhanced} leverage variational autoencoders (VAE) \cite{kingma2013auto} or generative adversarial networks (GAN) \cite{goodfellow2014generative} to compress images into low-dimensional latent spaces. The compressed data is then quantized and encoded, with the compression controlled by a $\lambda_{rate}$ parameter corresponding to the desired bitrate. However, these methods are constrained by hyperparameters like $\lambda_{rate}$, making it difficult to share model parameters or switch between different bit rates flexibly.
Choi et al. \cite{choi2019variable} introduced a conditional autoencoder framework that incorporates two rate control parameters, demonstrating the potential for adaptive-rate training. Unlike existing methods with limited compression ratio options, our approach constructs a Markov chain model for scalable feature compression, effectively balancing perception and bitrate tradeoffs.
\section{Proposed Method}
\begin{figure*}
  \includegraphics[width=0.9\textwidth]{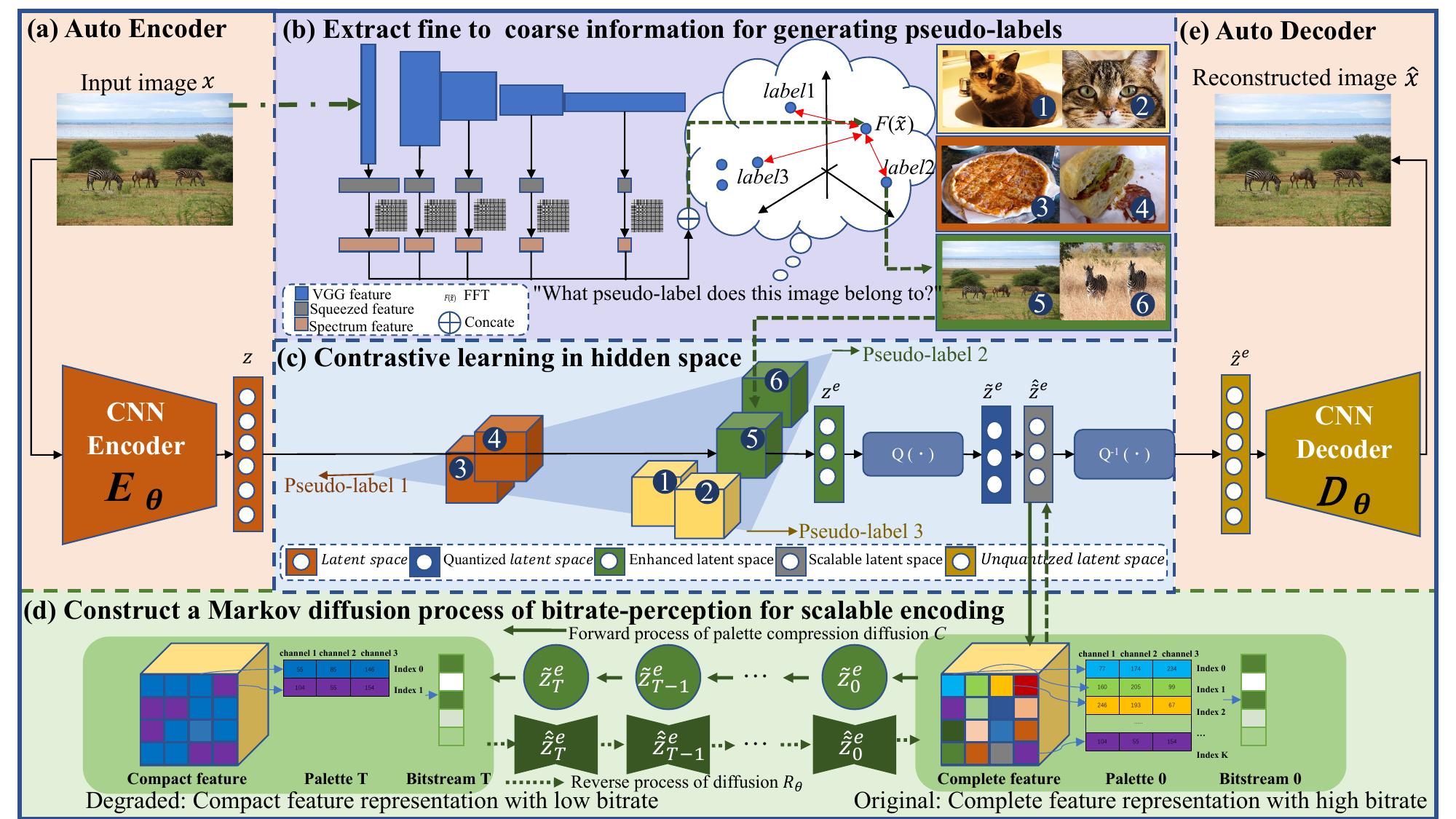} 
  \caption{Overview of our approach: (a) Compress the original image $x$ into a latent space $z$. (b) Extract fine-texture to coarse-semantic information of images and pseudo-labeling them in a self-supervised manner (Section \ref{subsubsection3-1-1} to \ref{subsubsection3-1-2}). (c) Enhance features via contrastive learning (Section \ref{subsubsection3-1-3}). (d) Constructing a Markov diffusion process of bitrate-perception for scalable encoding features(Section \ref{subsection3-2}). (e) Decode features and constructs an optimized estimation $\hat{x}$ of the original image $x$.}
  \vspace{-0.3cm}
  \label{fig:teaser}
\end{figure*}
In this section, we present our content-adaptive and diffusion-based scalable image feature compression approach in detail. Fig. \ref{fig:teaser} shows the flowchart for our framework. 
First, we extract the latent features from image patches by concatenating of a hierarchical encoder, and use clustering result of the extracted features to generate pseudo-labels for contrastive learning (Section \ref{subsection3-1}). Next, we design a Markov palette diffusion process for compression and regeneration of latent features (Section \ref{subsection3-2}). At last, in Section \ref{subsection3-3}, we define the overall training objective functions of the proposed method. 
\subsection{Texture-Semantic Pseudo-Label Extraction} \label{subsection3-1}
Instead of inefficient diffusion generation process in pixel-domain, many image regression methods \cite{rombach2022high, bansal2022cold} apply diffusion model in the feature domain extracted by a VAE \cite{kingma2013auto}. However, high-level features extracted by the encoder may discard most of the textural details, while shallow features cannot well describe image semantics. Therefore,  we integrate the low-level texture features and high-level semantic features to capture both the subtle texture and complex semantic concepts in the image data. In addition, in order to embed more discriminative texture and semantic information, contrastive learning is introduced to further improve the latent feature extraction. We thus propose a perceptual distance measurement in frequency domain to generate pseudo-labels for patches.
\subsubsection{Extraction of Texture-Semantic Representation}\label{subsubsection3-1-1}
Many previous studies \cite{ding2021locally, wang2023coarse} have shown that features extracted from pretrained CNNs (such as VGGNet and ResNet) can be used as a generic image representation and measure the perceptual distance. In this paper, we adopt VGGNet-16 \cite{simonyan2014very} pre-trained on the ImageNet database as the backbone of the image encoder.

Given a input image $x$, the convolution responses of five VGG \cite{simonyan2014very} layers are denoted as $conv1^{(2)}$, $conv2^{(2)}$, $conv3^{(3)}$, $conv4^{(3)}$, and $conv5^{(3)}$. A visualization of the feature maps of the five stages is shown in Fig. \ref{fig:zebra} to provide the interpretability. Spatial structures are preserved in all the stages, where shallow features (Fig. \ref{fig:zebra}(b)) emphasize high-frequency detailed information, the middle features focus on contours and deep features highlight low-frequency coarse-grained image semantic information (Fig. \ref{fig:zebra}(f)). By combining information from various dimensions, we can gain a more comprehensive understanding of the image. A transformation function $t: \Bbb R^n\mapsto \Bbb R^r$ maps the images $x$ from pixel domain to the texture-semantic collaborative representations $f_x$ using the equation:
\begin{equation}
f_x =t(x) = concat(\tilde{x}_j^{(i)}; i, \dots, m; j, \dots, n_i)\label{eq1},
\end{equation}
where $\tilde{x}$ denotes the feature map of image $x$, $m$ denotes the number of convolution layers and $n_i$ denotes the number of channel in the $i$-th convolution layer.

\subsubsection{Perceptual distance measurement} \label{subsubsection3-1-2}
For perceptual distance measurement, a desirable attribute is the flipping and translation invariance.
However, the extracted texture representation $f_x$ has a strong correlation with spatial coordinates, and thus is highly sensitive to translation, rotation, and flipping of image $x$. 
It is necessary to decouple the texture representation and the pixel coordinates. Therefore, we further perform Fast Fourier Transform (FFT) \cite{bracewell1986fourier} spectrum analysis for each $\tilde{x}$.
\begin{equation}
F(k,l)=\sum_{p=0}^{N-1}\sum_{q=0}^{N-1}\tilde{x}(p,q)e^{-i\pi/2(\cfrac{kp}{N}+\cfrac{lq}{N})} \label{eq2}, 
\end{equation}
where $\tilde{x}$ denotes the fine-detail to coarse-semantic information extractor described in Eqn. \ref{eq1}, $F$ is the coefficient value in the frequency domain, $(p,q)$ is the spatial coordinates of the feature space, $(k,l)$ is the basis in the frequency domain. 

As the magnitude of FFT spectrum \cite{bracewell1986fourier} contains all the information required to represent the geometric structure of the image, we only consider the magnitude $|F(k,l)|$.  We then use the Frobenius norm to measure the perceptual distance between two FFT \cite{bracewell1986fourier} spectrum matrices, as follows:
\begin{equation}
d(x, y)= \cfrac{\sum_{i=1}^m F(\tilde{x})^{(i)}) \times F(\tilde{y})^{(i)}}{ \sqrt{\sum_{i=1}^m (F(\tilde{x})^{(i)})^2}x\sqrt{\sum_{i=1}^m (F(\tilde{y})^{(i)})^2}} ,\label{eq4}
\end{equation}
where, given images $x$ and $y$, $\tilde{x}$ and $\tilde{y}$ are their VGG16 features according to Eqn.(\ref{eq1}), $m$ denotes the number of convolution layers , $F(\tilde{x})$ and $F(\tilde{y})$ denote the FFT spectrum \cite{bracewell1986fourier} response of the feature maps according to Eqn.(\ref{eq2}), and $d(x, y)$ denotes the perceptual distance between the two images.

\subsubsection{Pseudo-label Generation and Contrastive Learning} \label{subsubsection3-1-3}
\begin{figure}   
\includegraphics[width=0.4\textwidth]{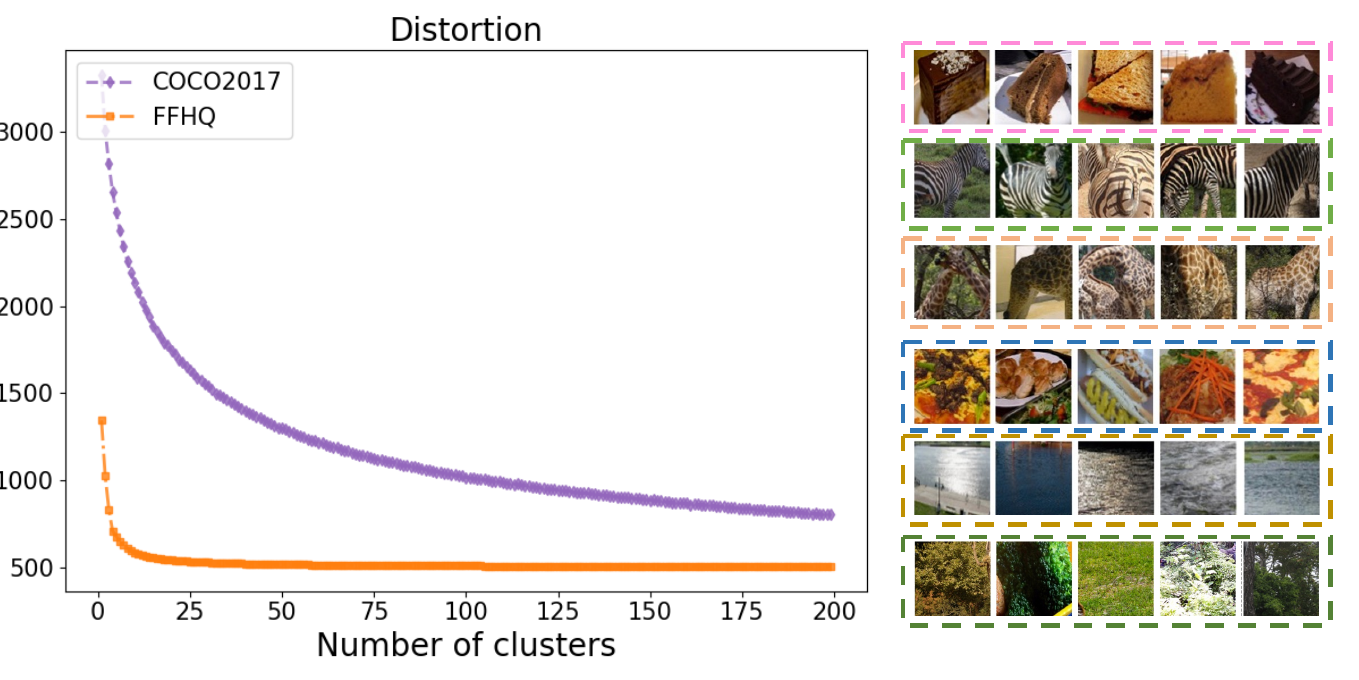}
  \caption{Clustering number and distortion curves on COCO 2017 \cite{lin2014microsoft} and FFHQ \cite{karras2019style}, and clustering results visualization.}
  \label{fig:four}
\vspace{-0.5cm}
\end{figure}

To enhance the compactness of features with similar texture-semantic information and maximize the margin between dissimilar  features, We use K-means \cite{hartigan1979algorithm} to cluster image patches, where the distance of samples is measured by Eqn. (\ref{eq4}), to generate pseudo-labels in a self-supervised manner. The optimal value of K is determined using the Elbow Method \cite{liu2020determine}. Figure \ref{fig:four} portrays the average clustering variation as a function of the number of clusters, with the "elbow" point occurring at where the average clustering variation starts to level off. 

The pseudo-label generation task allows for the determination of which patches are similar (having the same pseudo-label) and hence provides supervisory signals for comparative learning training \cite{he2020momentum}, through which we embed the collaborative fine-detail and coarse semantic knowlege into latent space. We thus enhance the discrimination of latent space $z$ and obtain $z^e$ for scalable encoding in Section \ref{subsection3-2}. The training loss is discussed in detail in Section \ref{subsection3-3}.
\subsection{Diffusion-Based Image Feature Compression} \label{subsection3-2}
Besides enhancing the feature representation to be discriminative, we also proposed a novel diffusion based scalable image feature compression method which tries to address the following challenges. Firstly, the feature representation can switch between fully complete and sparsely compact, depending on preserving pixel details or semantic information. Secondly, the bitrate-perceptual trade-off adjustment mechanism needs to be scalable while ensuring that the same model parameters can satisfy variable bitrate-perception quality requirements during the inference process. Lastly, the decoded images should be diverse and realistic. 

Different from the random noise degradation used in the original Stable Diffusion, we propose a novel degradation operator: the Palette Compression Algorithm based on Hierarchical Clustering. During the training phase, a Markov degradation is created which compresses the high-birate and complete feature map to compact and low-bitrate feature map. The diffusion model is then trained to learn how to restore the compressed and damaged features to their original high-bitrate state during the reverse process. When a compression ratio is specified by the codec user, our method performs palette compression as per the forward process during the training phase. Then restoration model can maximize the posterior probability of the complete feature. This ensures the reconstruction of a realistic image, even with extremely low bit-rate.
\vspace{-0.2cm}
\subsubsection{Forward Process of Diffusion}\textbf{Quantization.} As shown in Fig.\ref{fig:teaser}, in the compression process, the quantization module $Q(\cdot)$ first quantized the latent feature $z^e$ from floating-point values to an $8$-bit unsigned integer representation, indicated as $\tilde{z}^e$. To optimize our model during training, when calculating the gradients, we apply the straight-through estimator method (as referenced in Invertible Rescaling Network (IRN) \cite{xiao2023invertible}) to the quantization module. 

\noindent \textbf{Palette compression.} Palette compression is a the critical degradation operator in our diffusion forward process. This technique generates a Color LookUp Table (CLUT) of $K$ entries in an image-content adaptive manner, which can be used to map the input feature maps $\tilde{z}$ with $\Bbb R^{W\times H \times C}$, to the color table including $W\times H$ index values. The range of the index values is $0$ to $K-1$, significantly reducing the number of required bits to represent information and favoring further compression algorithms like Huffman coding. 

\begin{figure}[h]
  \centering  \includegraphics[width=0.4\textwidth]{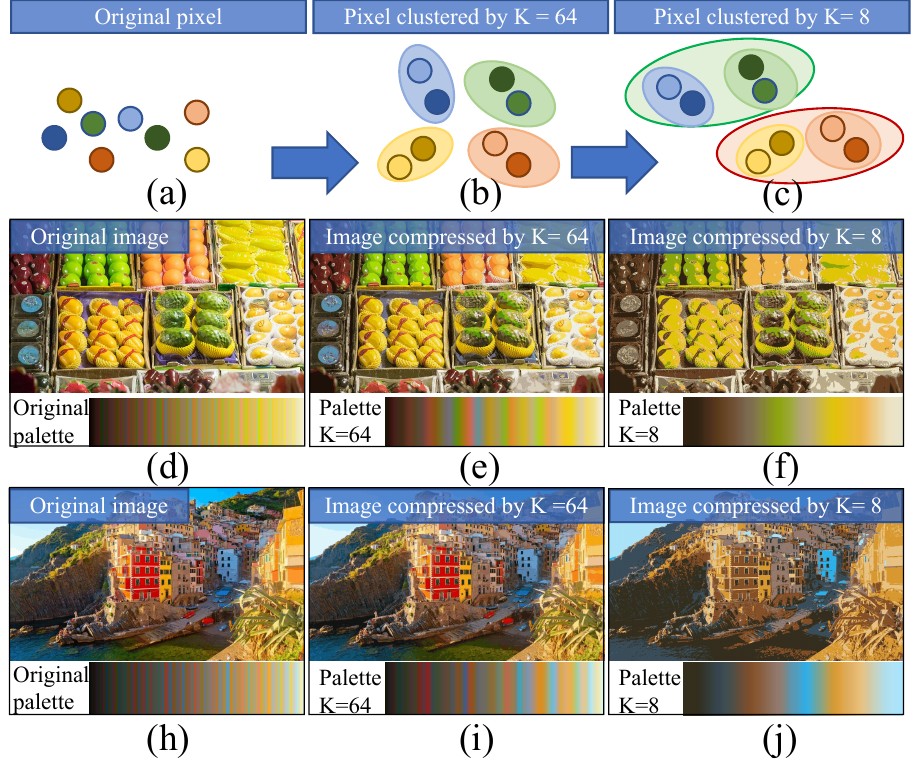}
  \vspace{-0.4cm}
  \caption{An example of using the hierarchical clustering method to construct a palette compression. }
  \Description{A woman and a girl in white dresses sit in an open car.}
  \label{fig4}
  \vspace{-0.5cm}
\end{figure}

Algorithm \ref{alg1} describes the construction of the adaptive palette utilized in the forward diffusion process. Additionally, Fig. \ref{fig4}(a)-(b) illustrates how the original data are aggregated into K$=$64. (d)-(e) and (h)-(i) show the qualitative degradation of the resultant compressed image. The second and third rows of Fig. \ref{fig4} show that different images build their color palette, illustrating that this approach is self-supervised and content-adaptive. Hence we do not need to compress the entire color palette codebook for all images in the dataset, effectively reducing data to be transmitted.

\noindent \textbf{Markov state transition equation.} To enable users to select the cost-effectiveness of compression during the testing phase, inspired by Cold Diffusion \cite{bansal2022cold}, we use hierarchical bottom-up clustering method to build a Markov chain of palettes with increasingly sparse CLUT containing K entries. The process is shown in Fig. \ref{fig4} (a-c). Initially, a palette of K colors is constructed, and the data points are then gradually merged into $K-1,...,1$ clusters, and the quality of the image reconstructed by the decoder is gradually degraded. Given a latent space $\tilde{z}^e \in \Bbb R^{W\times H \times C}$, consider the degradation of $\tilde{z}^e$ by the operator hierarchical clustering $C$ with severity $t$, denoted by $\tilde{z}^e_t = C(\tilde{z}^e_0, t)$. Define the number of color entries at step $t$ is K, and operator $C_t $ means perform Algo. \ref{alg1} in  $\tilde{z}^e_{t-1}$ with specified K= K-1, The output distribution $C(\tilde{z}_0,t)$ of the degradation should vary continuously in t, and the operator should satisfy:
\begin{equation}
\tilde{z}^e_t = C_t * \tilde{z}^e_{t-1} =C_t *...*C_1 * \tilde{z}^e_0 = C(\tilde{z}^e_0,t)\label{eq6},
\end{equation}
In the standard diffusion framework, operator $C_t $ adds Gaussian noise with variance proportional to $t$. In our formulation, operator $C$ denotes hierarchical palette compression, the degree of which depends on $t$. The forward process of diffusion is shown in Fig.\ref{fig:teaser} (d) to illustrate how it works in our scalable feature compression framework.
 
\subsubsection{Reverse Process of Diffusion} 
In addition, we require a restoration operator $R$ that approximately inverts $C$. This operator has the following target:
\begin{equation}
R(\tilde{z}^e_t, t)\approx \tilde{z}^e_0  \label{eq7}.
\end{equation}
The reverse diffusion process maximizing the posterior probability, and the state transition can be formulated as:
\begin{equation}
\hat{\tilde{z}}^e_0 = R(\hat{\tilde{z}}^e_t, t),
\end{equation}
\begin{equation}
\tilde{z}^e_{t-1}=\tilde{z}^e_{t} - C(\hat{\tilde{z}}^e_0, t) + C(\hat{\tilde{z}}^e_0, t-1).
\end{equation}
In practice, this recovery operator $R$ is implemented via a neural network parameterized by $\theta$, similar to the approach in DDPM\cite{ho2022cascaded, ho2022imagen}. The structure of this network is a U-Net consisting encoder part and a decoder part both comprised of ResNet blocks. To prevent the U-Net from losing important information while downsampling, short-cut connections are added between the downsampling ResNets of the encoder to the upsampling ResNets of the decoder. Additionally, U-Net is able to condition its output on timestep $t$ embeddings via cross-attention layers. The cross-attention layers are added to both the encoder and decoder part of the U-Net usually between ResNet blocks. The restoration network is trained via the following minimization problem:
\begin{equation}
\min_{\theta} \Bbb E_{\tilde{z}^e\sim \chi} \lVert  R_{\theta}(C(\tilde{z}^e,t),t)-\tilde{z}^e \rVert \label{eq8},
\end{equation}
where $\tilde{z}$ denotes a random image sampled from distribution $\chi$ and $\lVert \cdot \rVert$ denotes a norm, taken as $l_1$ in our experiments. Thus far, we have used $  R_{\theta}$ to emphasize the dependence of $  R$ on $\theta$ during training.

During the test stage, the user specifies the compression quality $t$. The complete features extracted by the auto-encoder are compressed into a compact representation $\tilde{z}^e_t$ according to the user's instructions. $\tilde{z}^e_t$  and $t$ are used as the inputs of the reverse diffusion $R$. The feature is repaired by $ R_{\theta}$ to the completeness of its representation $\tilde{z}^e_0 $, as shown in Fig. \ref{fig:teaser} (d).
\begin{algorithm}
\small
  \renewcommand{\algorithmicrequire}{\textbf{Input:}}  
    \renewcommand{\algorithmicensure}{\textbf{Output:}}  
    \caption{Build content-adaptive palette index }  
    \label{alg1}
    \begin{algorithmic}[1] 
        \REQUIRE A quantized latent space $\tilde{z}$ of $\Bbb R^{ W \times H \times C}$, clustering number K, and randomly chosen K points ${u_1,...u_K}$ as the initialized centroids.
        \ENSURE A sequence of the palette index to represent $\tilde{z}$        
        \STATE Initialize $C_i,i=1,2,...K \Leftarrow \emptyset$
        \FOR {$i,j = (1,1),...,(H,W)$} 
            \STATE $d_{ij1} \Leftarrow {\Vert \tilde{z}_{ij}-u_1 \Vert}^2,..., d_{ijK} \Leftarrow {\Vert \tilde{z}_{ij}-u_K \Vert}^2$
            \IF {$d_{ijk} \leq d_{ij1},..., d_{ij1K}$}
                \STATE $C_k \Leftarrow C_k \cup \{\tilde{z}_{ij}\}$
            \ENDIF
        \ENDFOR
        \STATE $\tilde{u_1} \Leftarrow \frac{1}{\vert C_1 \vert}\sum_{\tilde{z} \in C_1} \tilde{z},..., \tilde{u_K} \Leftarrow \frac{1}{\vert C_K \vert}\sum_{x \in C_K} \tilde{z}$  
        \STATE $u_1 \Leftarrow \tilde{u_1},..., u_K \Leftarrow \tilde{u_K}$  
        \RETURN index of $\tilde{z}_{ij} $ where $\tilde{z}_{ij} \in \{C_1,...,C_K\}$
    \end{algorithmic}
\end{algorithm}
\subsection{Training Objectives} \label{subsection3-3}
\textbf{Contrastive Learning Loss for Compact Texture Representation.}
As described in Section \ref{subsection3-1}, we designed a proxy task to generate the pseudo-labels of images, thus providing  supervisory signals for comparative learning training models. As referenced in MOCO \cite{he2020momentum}, consider an encoded query $z_q$ and a set of encoded samples ${z_{k0}, z_{k1}, z_{k2},...}$ that are the keys of a dictionary. Assume that there is a single key (denoted as $z_{k+}$) in the dictionary that $z_q$ matches. A contrastive loss is a function whose value is low when $z_q$ is similar to its positive key $z_{k+}$ and dissimilar to all the other keys (considered negative keys for $z_q$). With the similarity measured using the dot product, a type of contrastive loss function called InfoNCE is considered in this study:
\begin{equation}
\mathcal{L}_{contrast}=-\log\frac{\exp(z_q\cdot z_{k+}/\tau)}{\sum_{i=0}^K\exp(z_q\cdot z_{ki}/\tau)} \label{9},
\end{equation}
where $\tau$ is a hyper-parameter for temperature. The sum is over a positive sample and $K$ negative samples. Intuitively, this loss is the log loss of a $(K+1)$-way softmax-based classifier that attempts to classify $z_q$ with the positive sample $z_k{+}$.  

\noindent \textbf{Diffusion Loss for Compact Semantic Feature Compression.}
As described in Section \ref{subsection3-2}, the training objective can be defined in the context of variational inference by approximating the posterior distribution of the latent variables $p_{\theta}(\tilde{z}^e_{1:T}|\tilde{z}^e_{0})$ using the forward process $q (\tilde{z}^e_{1:T}|\tilde{z}^e_{0})$. Further, by using Bayes' rule to obtain $q (\tilde{z}^e_{t-1}|\tilde{z}^e_{t}, \tilde{z}^e_{0})$, maximizing the evidence lower bound (ELBO) on $p_{\theta}(\tilde{z}^e_{0})$ is equivalent to minimizing the sum of T Kullback--Leibler (KL) divergences. This objective function can then be expressed as a simple minimization between true data and a denoising prediction:
\begin{equation}
\mathcal{L}_{DDPM}=\min_{\theta} \Bbb E_{\tilde{z}^e\sim \chi} \lVert  R_{\theta}(C(\tilde{z}^e,t),t)-\tilde{z}^e \rVert \label{eq10},
\end{equation}
where $R_{\theta}(C(\tilde{z}^e,t),t)$ denotes a model that predicts $\tilde{z}^e_0$ from $ \tilde{z}^e_t$ . The aforementioned equation integrates $t$ in the expectation. Although the complete loss should sum over all $t$, it is a common practice to sample $t$ and perform Monte Carlo integration over time instead.

\noindent \textbf{Image Reconstruction Loss.}
In this study, we use a decoder $D_{\theta}^d$ to reconstruct $ x$ , which allows it to learn to improve the perceptual quality, as outlined below:
\begin{equation}
\mathcal{L}_{percep}=\min_{\theta} \Bbb E_{\tilde{z}^e\sim \chi} \lVert  D_{\theta}(R_{\theta}(C(\tilde{z}^e,t),t))-x \rVert \label{eq11}.
\end{equation}

\noindent \textbf{Total Loss.}
Our training stage minimizes the following total objective:
\begin{equation}
\mathcal{L}_{total}:=\lambda_1 \mathcal{L}_{contrast} + \lambda_2 \mathcal{L}_{DDPM} + \lambda_3 \mathcal{L}_{percep} \label{eq12},
\end{equation}
where $\lambda_1$, $\lambda_2$, $\lambda_3$ are coefficients for balancing different loss terms.

\section{Experiments}
Our experiments consist of two parts: 1) evaluating human vision through image compression and reconstruction; and 2) assessing machine vision tasks, including image object detection, segmentation, and facial landmark detection, using the reconstructed images. In Section \ref{subsection4-1}, we present the dataset and experimental settings. In Section \ref{subsection4-2}, we provide both quantitative and qualitative evaluations of our method for human vision perception. Section \ref{section4-3} discusses the performance of our method on three machine vision tasks: object detection, instance segmentation, and facial landmark detection. Finally, in Section \ref{section4-4}, we discuss our scalable mechanism.
\subsection{Datasets and Settings} \label{subsection4-1}
\textbf{Dataset.} We train our models for image coding using widely-used COCO 2017 \cite{lin2014microsoft} and  FFHQ \cite{karras2019style}, and accordingly evaluate the model on their validation set. For evaluation of machine vision tasks, we use COCO 2017 \cite{lin2014microsoft} for object detection and segmentation, and WIDER FACE \cite{yang2016wider} for the facial landmark detection task. 

\noindent \textbf{Evaluation Metrics.} In Setion \ref{subsection4-2}, to evaluate the human perceptual quality, we utilized Learned Perceptual Image Patch Similarity (LPIPS) \cite{zhang2018unreasonable}, and Fréchet Inception Distance (FID) \cite{heusel2017gans} as metrics. We plotted the curves for different bitrates and the peceptual metrics. Additionally, Peak Signal to Noise Ratio (PSNR) and Structural Similarity Index (SSIM) indices are also  provided as references for signal-level fidelity. In Section \ref{section4-3}, we compressed images at an extremely low bit-rate (0.15 bpp) and used mean Average Precision (mAP) and Average Recall (AR) as machine vision task metrics.
 
\noindent \textbf{Training Settings.} Our network is trained using the Adam optimizer with $\beta_1$=0.9 and $\beta_2$=0.999, while the mini-batch size is set to 16. Before training, the input image is randomly cropped into N × N and further augmented by random horizontal and vertical flips. We initialize the learning rate as $2\times10^{-4}$ and train the model for a total of 500,000 iterations, with the learning rate halving at $[100k, 200k, 300k, 400k]$ mini-batch updates. The hyper-parameters specified in Eqn.(\ref{eq12}) are set as $\lambda_1=0.1$, $\lambda_2=1$, and $\lambda_3=1$. 
\begin{figure*}[h]
  \centering
  \begin{subfigure}{.24\linewidth}
  \centering
  \includegraphics[width=1.0\linewidth]{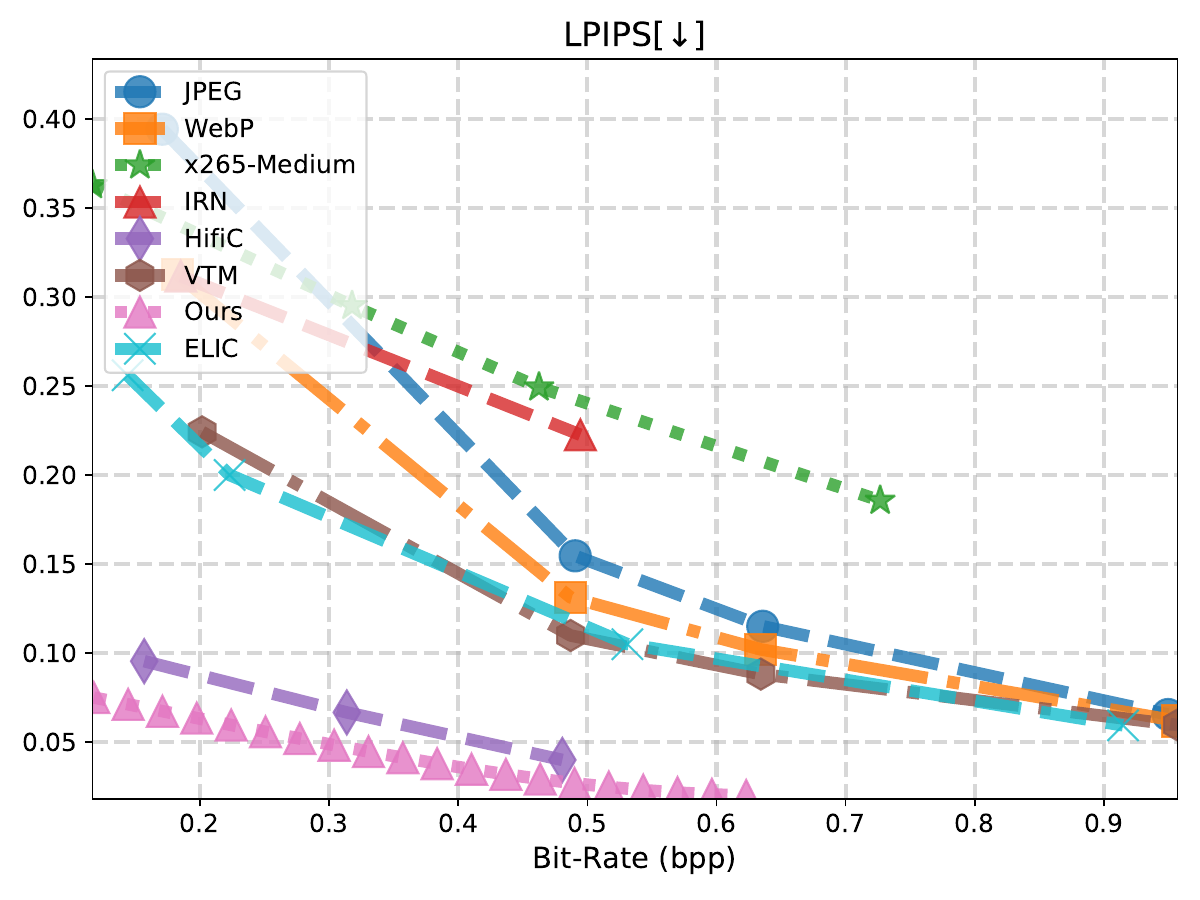}
  \caption{}
  \label{fig5_a}
  \vspace{-0.5cm}
  \end{subfigure}
  \begin{subfigure}{.24\linewidth}
  \centering
  \includegraphics[width=1.0\linewidth]{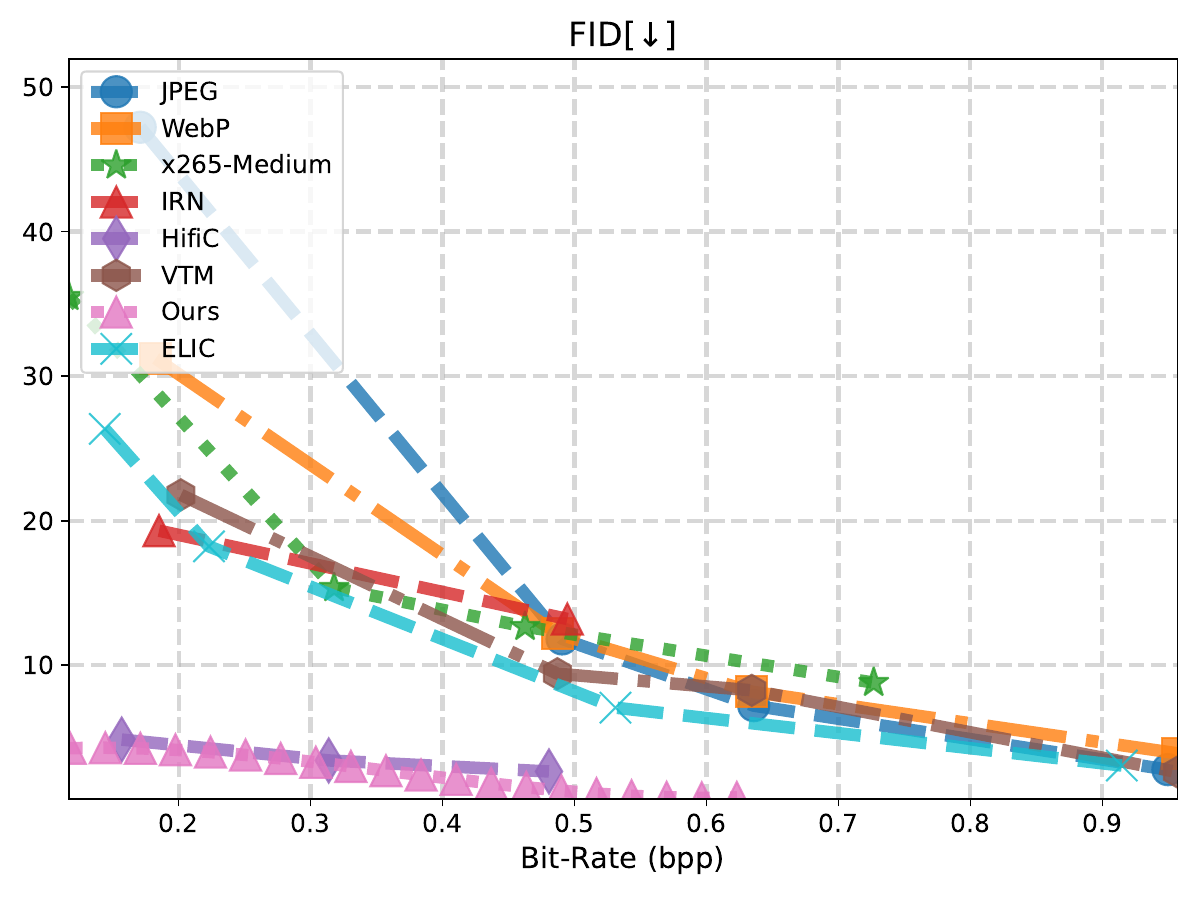}
  \caption{}
   \label{fig5_b}
   \vspace{-0.5cm}
  \end{subfigure}
    \begin{subfigure}{.24\linewidth}
  \centering
  \includegraphics[width=1.0\linewidth]{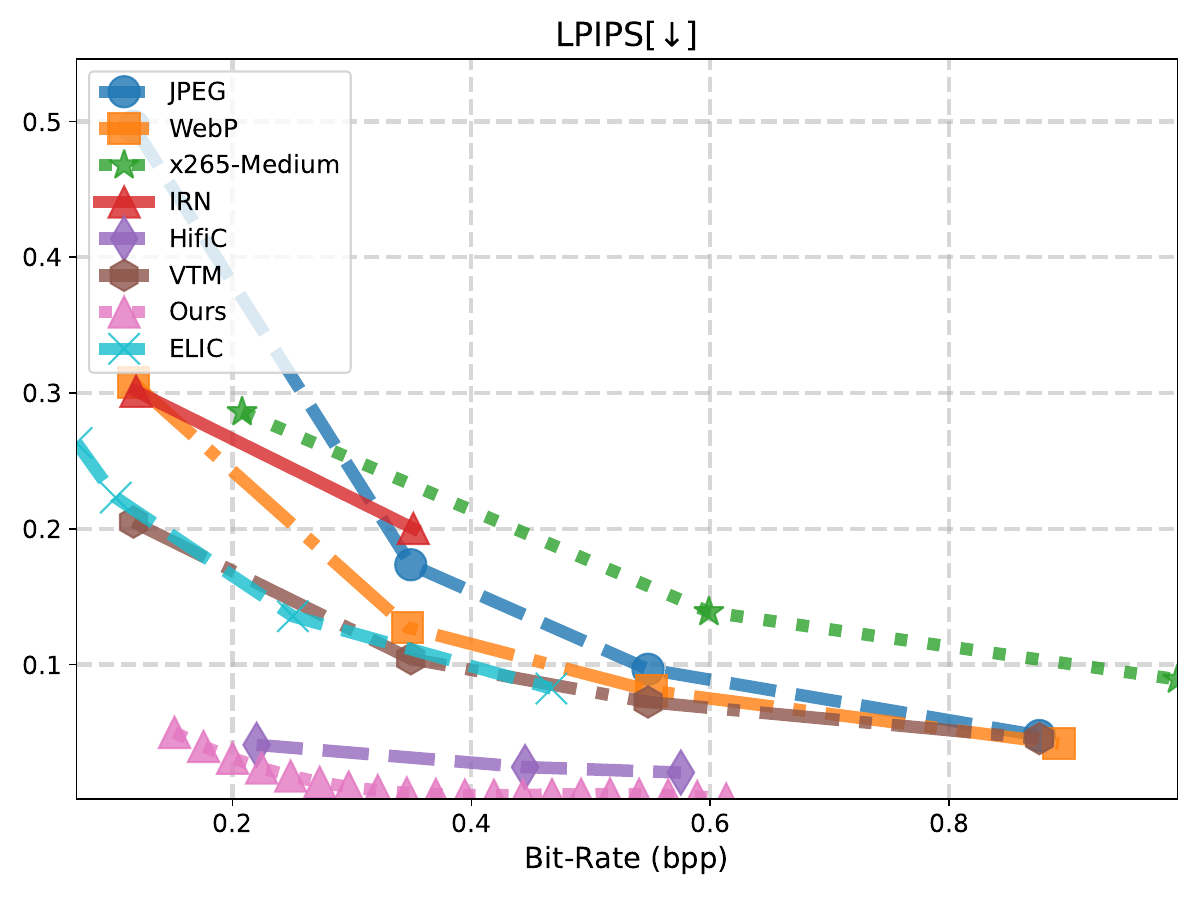}
  \caption{}
  \label{fig5_c}
  \vspace{-0.5cm}
  \end{subfigure}
  \begin{subfigure}{.24\linewidth}
  \centering
  \includegraphics[width=1.0\linewidth]{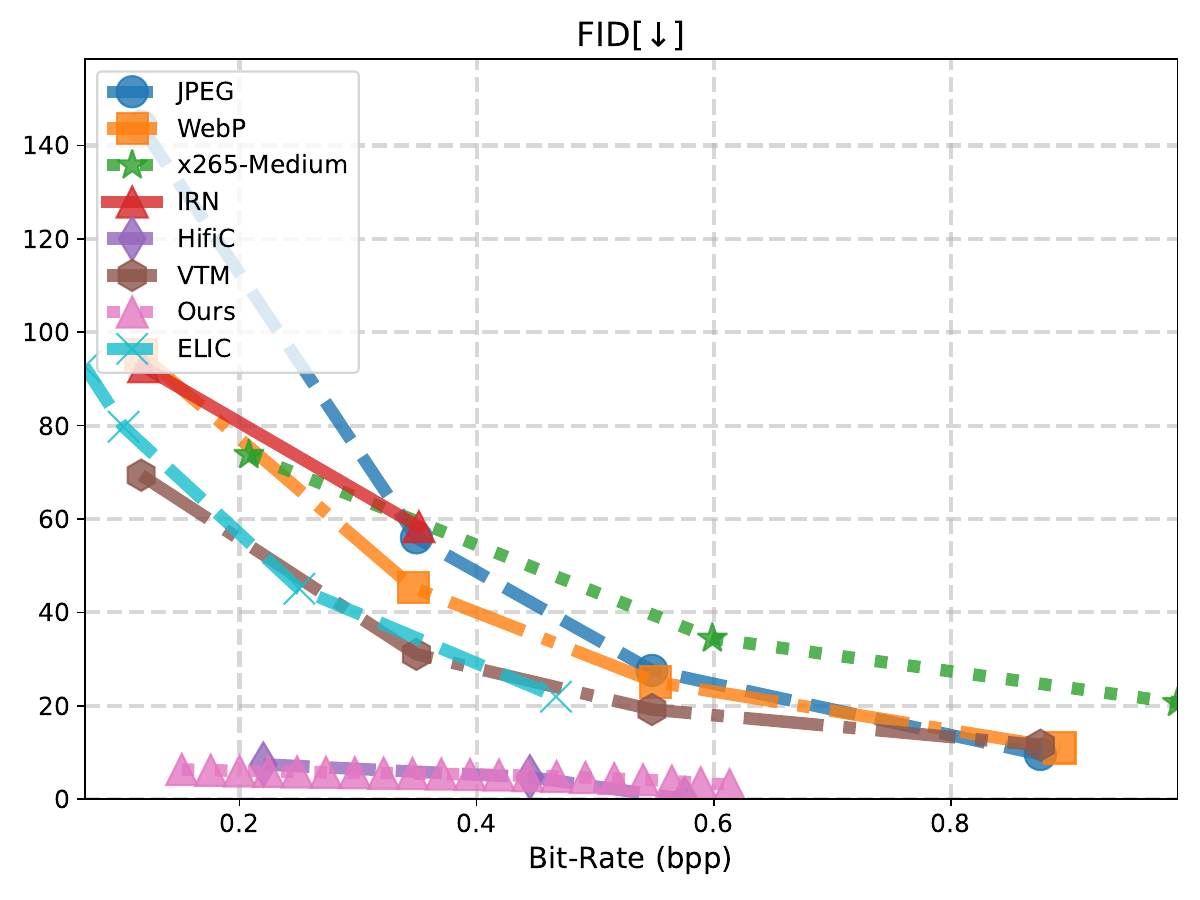}
  \caption{}
    \label{fig5_d}
    \vspace{-0.5cm}
  \end{subfigure}
  \caption{ Rate-perception curves: (a) and (b) are the results on COCO 2017 \cite{lin2014microsoft}, (c) and (d) are the results on FFHQ \cite{karras2019style}.}
  \label{fig5}
  \vspace{-0.2cm}
\end{figure*}
\vspace{-0.2cm}
\subsection{Evaluation for Human Vision} \label{subsection4-2}
In this section, we present a comprehensive comparison of the quantitative and qualitative performance of images reconstructed using our method, against several image compression techniques, such as JPEG \cite{wallace1992jpeg}, WebP \cite{mukherjee2014webp}, x265 \cite{ramachandran2013x265}, and VTM \cite{bross2021overview}, as well as state-of-the-art deep learning-based downscaling and upscaling methods, notably the Invertible Rescaling Network (IRN) \cite{xiao2023invertible}, and end-to-end compression method High Fidelity Generative Compression (HiFiC) \cite{mentzer2020high}, Efficient learned image compession (ELIC) \cite{he2022elic}. Additionally, we investigate the influence of different patch sizes $N$ and the effectiveness of pseudo-label based contrastive learning. 

\noindent \textbf{Quantitative Results.} Figure \ref{fig5} presents the rate-perception tradeoffs of our method in comparison to JPEG \cite{wallace1992jpeg}, WebP \cite{mukherjee2014webp}, x265 \cite{ramachandran2013x265}, VTM \cite{bross2021overview}, IRN \cite{xiao2023invertible} , HiFiC \cite{mentzer2020high} and ELIC \cite{he2022elic}. To guarantee sufficient sample size for all metrics to be reliable (especially FID \cite{heusel2017gans}), we selected COCO 2017 validation dataset and the HHFQ validation dataset for testing. We observed that our proposed method performs best in LPIPS \cite{zhang2018unreasonable} and competes with the state-of-the-art HiFiC \cite{mentzer2020high} method in FID \cite{heusel2017gans}, despite the fundamental differences in design motivation and algorithmic details. HifiC \cite{mentzer2020high}  has only three bit-rate points because it offers three models of high, medium and low bitrate. In contrast, our method allow users to switch flexibly between high and low bitrates based on different perception quality requirements. As demonstrated in Figure \ref{fig5}, the adjustable range of our method covered the range of HiFiC \cite{mentzer2020high}. Note that traditional encoding algorithms can reconstruct images without loss of quality at peak bitrates. However, feature compression algorithms have limitations in achieving lossless reconstruction. Nonetheless, our proposed method can achieve superior encoding performance at extremely low bitrates (below 0.2 bits per pixel) and thus satisfy data-intensive contexts, such as IoT and monitoring.

We also provided PSNR and SSIM of all methods at a compression ratio of 0.15 bpp in Table \ref{table:111} and found that our method did not perform well in these fidelity metrics. This is mainly due to the fact that our method is based on a  generative model that captures high-level features that do not perfectly match the original image at a pixel level, resulting in low values for PSNR/SSIM indices. In addition, the results of the generative model include vivid details that satisfy subjective quality, but have a disadvantage in terms of the PSNR/SSIM metric. Previous research \cite{blau2018perception} shows that PSNR/SSIM can run counter to subjective quality in terms of restoring realistic textures. The qualitative result in Section \ref{subsection4-2} and machine vision task performance in Section \ref{section4-3} also demonstrate that lower PSNR/SSIM did not harm the perception of the human eye and machine.
\begin{table}[h!]
\small
\caption{PSNR/SSIM results on mixed COCO 2017 \cite{lin2014microsoft} and FFHQ validation \cite{karras2019style} show that our method enhances fine textures to improve perceptual visual quality but results in a significant decrease in PSNR indicators.}
\centering
\begin{tabularx}{\linewidth}{*{1}{>{\centering\arraybackslash}X} | p{6em}        |*{1}{>{\centering\arraybackslash}X}|*{1}{>{\centering\arraybackslash}X}} 
 \hline
Method                                                    & Bit-Rate (bpp) & PSNR $\uparrow $                    & SSIM $\uparrow$\\[0.5ex] 
 \hline\hline
 \multirow{1}{*}{JPEG  \cite{wallace1992jpeg}}   & \centering 0.153   & 24.045  & 0.641   \\ 
 \multirow{1}{*}{WebP \cite{mukherjee2014webp} }& \centering 0.154 & 26.342  &  0.709  \\  
\multirow{1}{*}{x265 \cite{ramachandran2013x265}}  & \centering 0.168   & 24.790  & 0.651    \\ 
 \multirow{1}{*}{ IRN \cite{xiao2023invertible}} & \centering 0.154    & 25.443  & 0.688   \\  
 \multirow{1}{*}{ VTM \cite{bross2021overview}} & \centering  0.161    & 28.736 & 0.790    \\  
 \multirow{1}{*}{ HifiC \cite{mentzer2020high}} & \centering 0.156    & 26.689  & 0.761   \\ 
  \multirow{1}{*}{ ELIC \cite{he2022elic}} & \centering 0.144    & \textbf{28.749}  & \textbf{0.796}    \\ 
 \multirow{1}{*}{ Ours} & \centering 0.153    & 23.434  & 0.716    \\  
 \hline
\end{tabularx}
\label{table:111}
\vspace{-0.3cm}
\end{table}

\begin{table}[h!]
\small
\caption{Quantitative evaluation of various loss function combinations and training patch size $N$ on mixed COCO 2017 \cite{lin2014microsoft} and FFHQ validation datasets \cite{karras2019style}.}
\centering
\begin{tabularx}{\linewidth}{p{3em} |p{3em}|p{3em}|p{5em}|*{1}{>{\centering\arraybackslash}X}|*{1}{>{\centering\arraybackslash}X}} 
 \hline
 {\fontsize{5pt}{6pt}\selectfont $\mathcal{L}_{contrast}$} &{\fontsize{5pt}{6pt}\selectfont $\mathcal{L}_{DDPM} $}&{\fontsize{5pt}{6pt} \selectfont $\mathcal{L}_{percep}$ }& \fontsize{7pt}{6pt} \selectfont LPIPS \cite{zhang2018unreasonable} $\downarrow$ & \fontsize{7pt}{6pt}\selectfont PI \cite{blau2018perception} $\downarrow$ & \fontsize{7pt}{6pt} \selectfont FID \cite{heusel2017gans} $\downarrow$\\ [0.5ex] 
 \hline\hline
 \multirow{1}{*}{$\checkmark$} & \multirow{1}{*}{$\checkmark$}& \multirow{1}{*}{$\checkmark$} & \centering  \textbf{0.069} & \textbf{14.196} & \textbf{4.943}  \\ 
 \multirow{1}{*}{ } & \multirow{1}{*}{$\checkmark$} & \multirow{1}{*}{$\checkmark$} & \centering 0.085  & 14.674  & 14.253 \\  
 \hline
 \hline
 \multicolumn{3}{c|}{\fontsize{7pt}{6pt}\selectfont Patch Size N} & \fontsize{7pt}{6pt}\selectfont LPIPS \cite{zhang2018unreasonable} $\downarrow$ & \fontsize{7pt}{6pt}\selectfont PI \cite{blau2018perception} $\downarrow$ & \fontsize{7pt}{6pt}\selectfont FID \cite{heusel2017gans} $\downarrow$ \\ [0.5ex] 
 \hline\hline
\multicolumn{3}{c|}{128}   & \centering 0.089   & 14.239  & 16.652   \\ 
 \multicolumn{3}{c|}{256 } & \centering 0.063 & 13.964  &  4.875  \\ 
  \multicolumn{3}{c|}{320}  & \centering \textbf{0.061}   & \textbf{13.873}  & \textbf{4.682}    \\ 
 \multicolumn{3}{c|}{ 512} & \centering 0.078    & 14.829  & 7.101 \\ 
 \hline
\end{tabularx}
\label{table:2}
\end{table}

\begin{figure*}[ht]
\label{fig:subjective1}
\centering
\begin{subfigure}{.09\linewidth}
  \centering
  \includegraphics[width=1.0\linewidth]{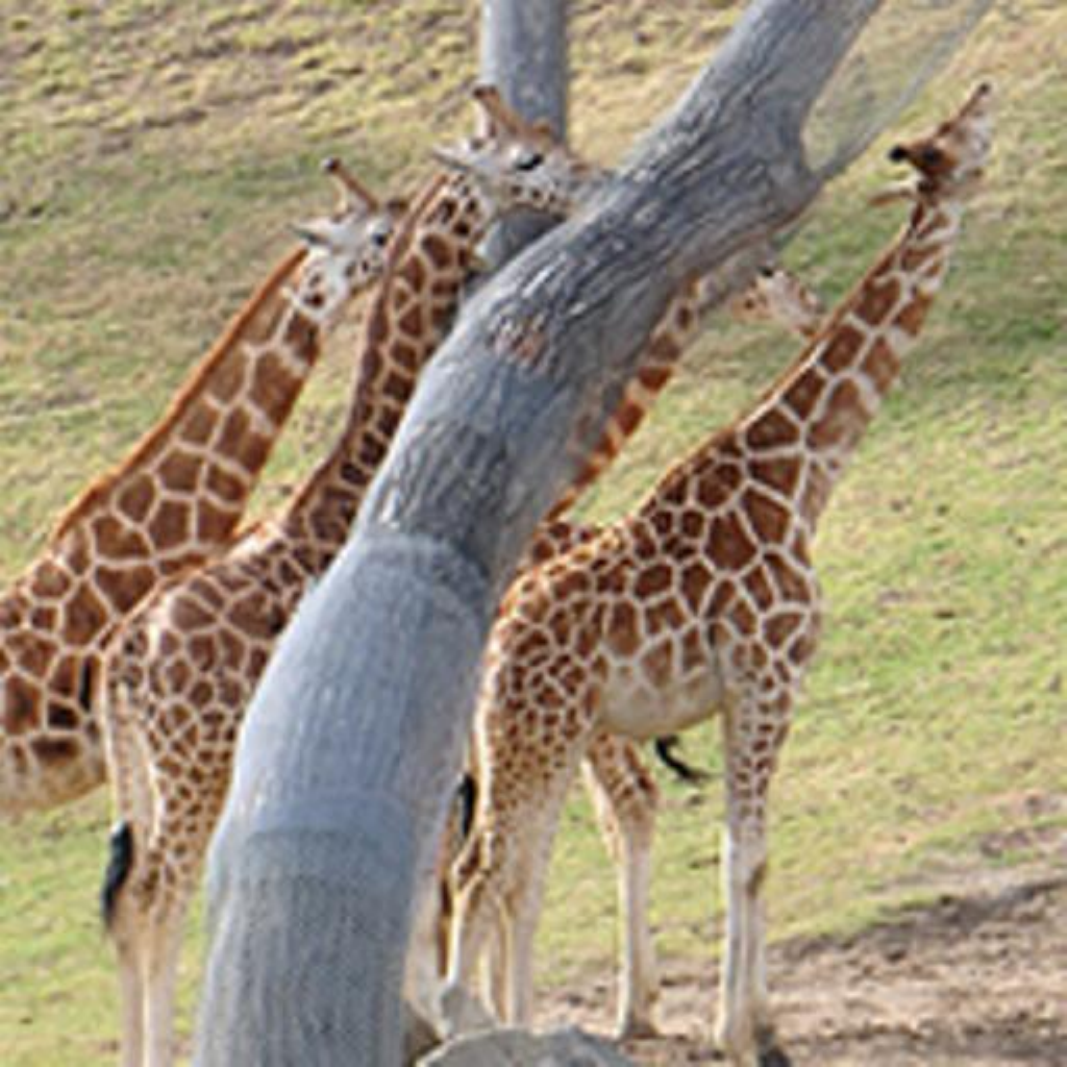}
  \includegraphics[width=1.0\linewidth]{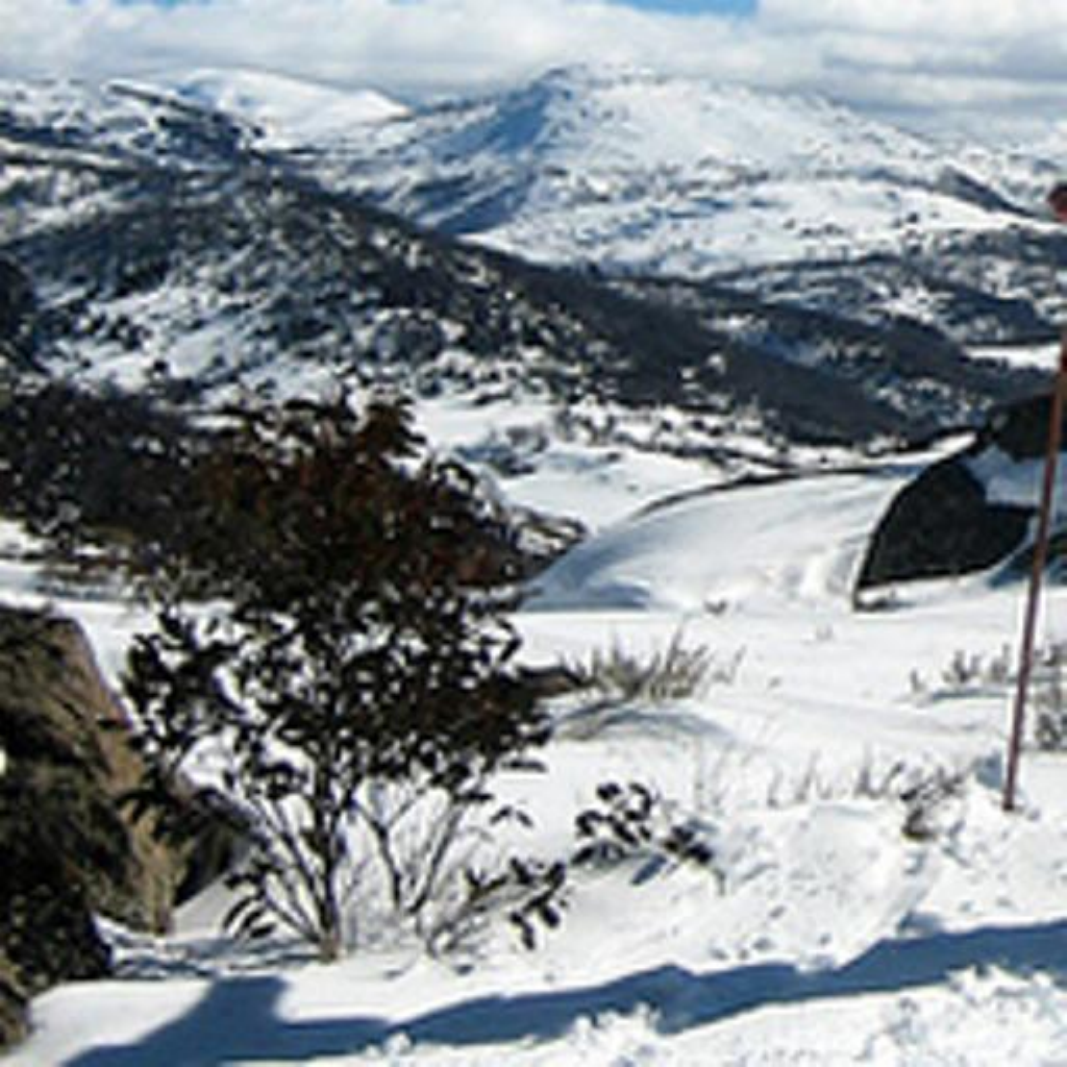}
  \includegraphics[width=1.0\linewidth]{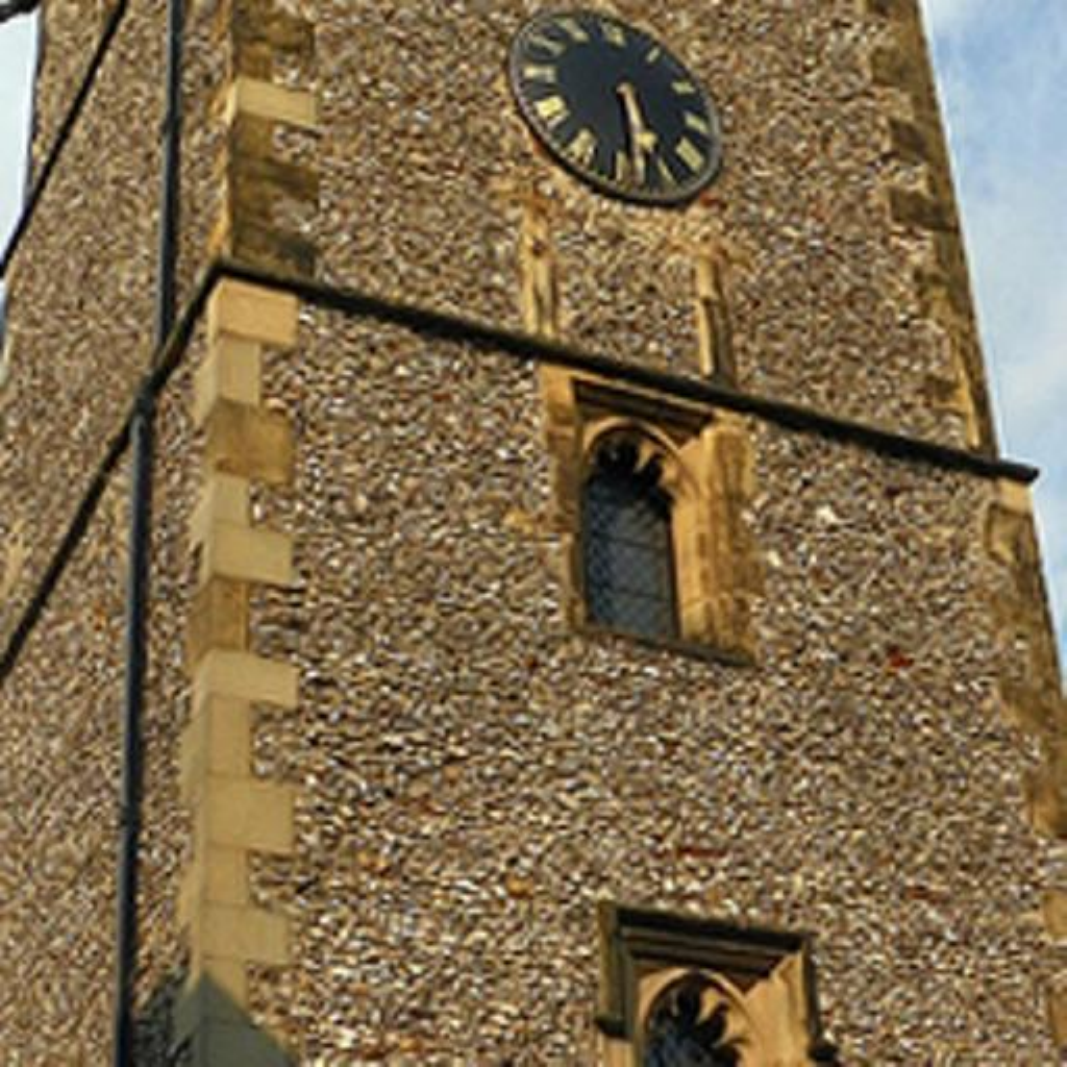}
  \includegraphics[width=1.0\linewidth]{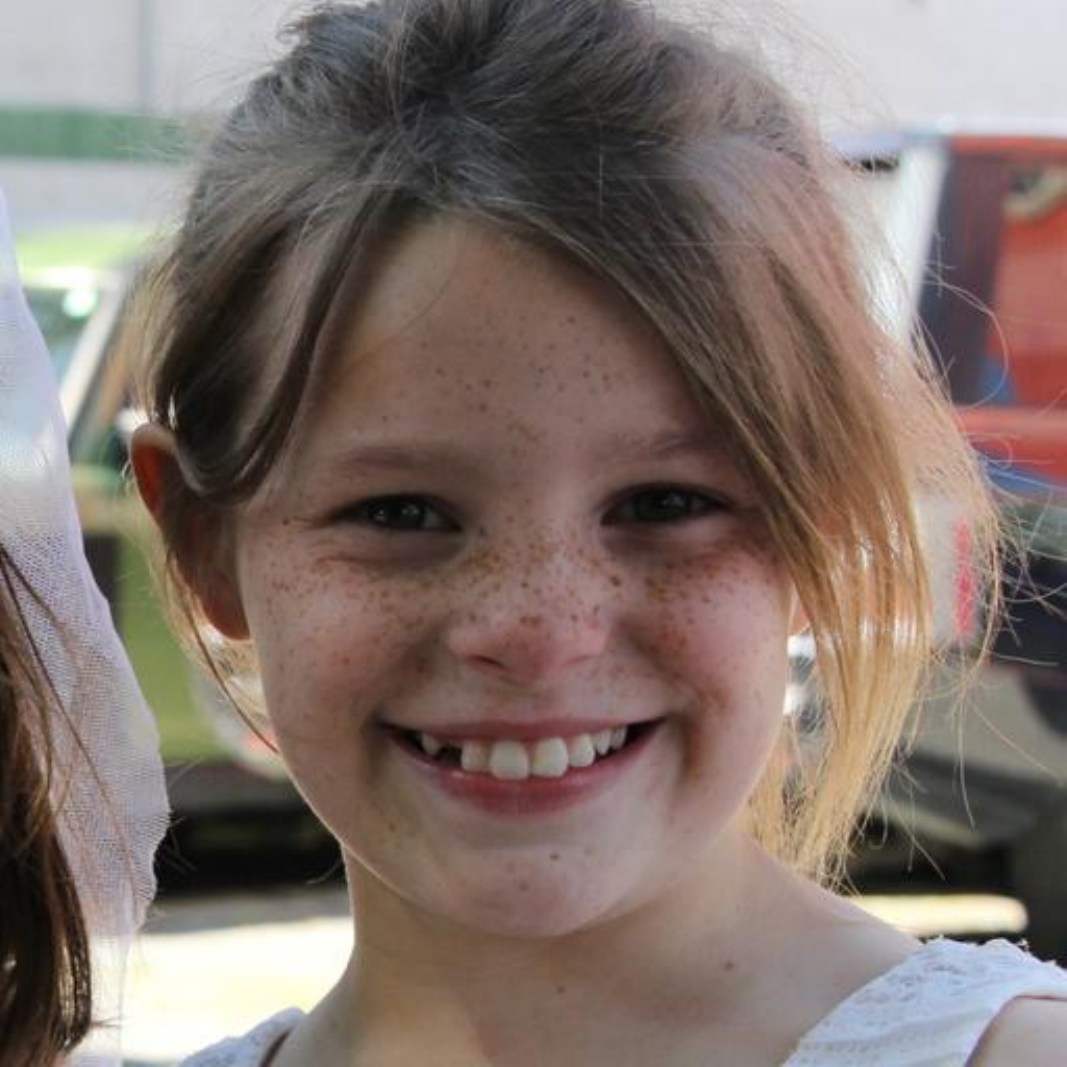}
    \includegraphics[width=1.0\linewidth]{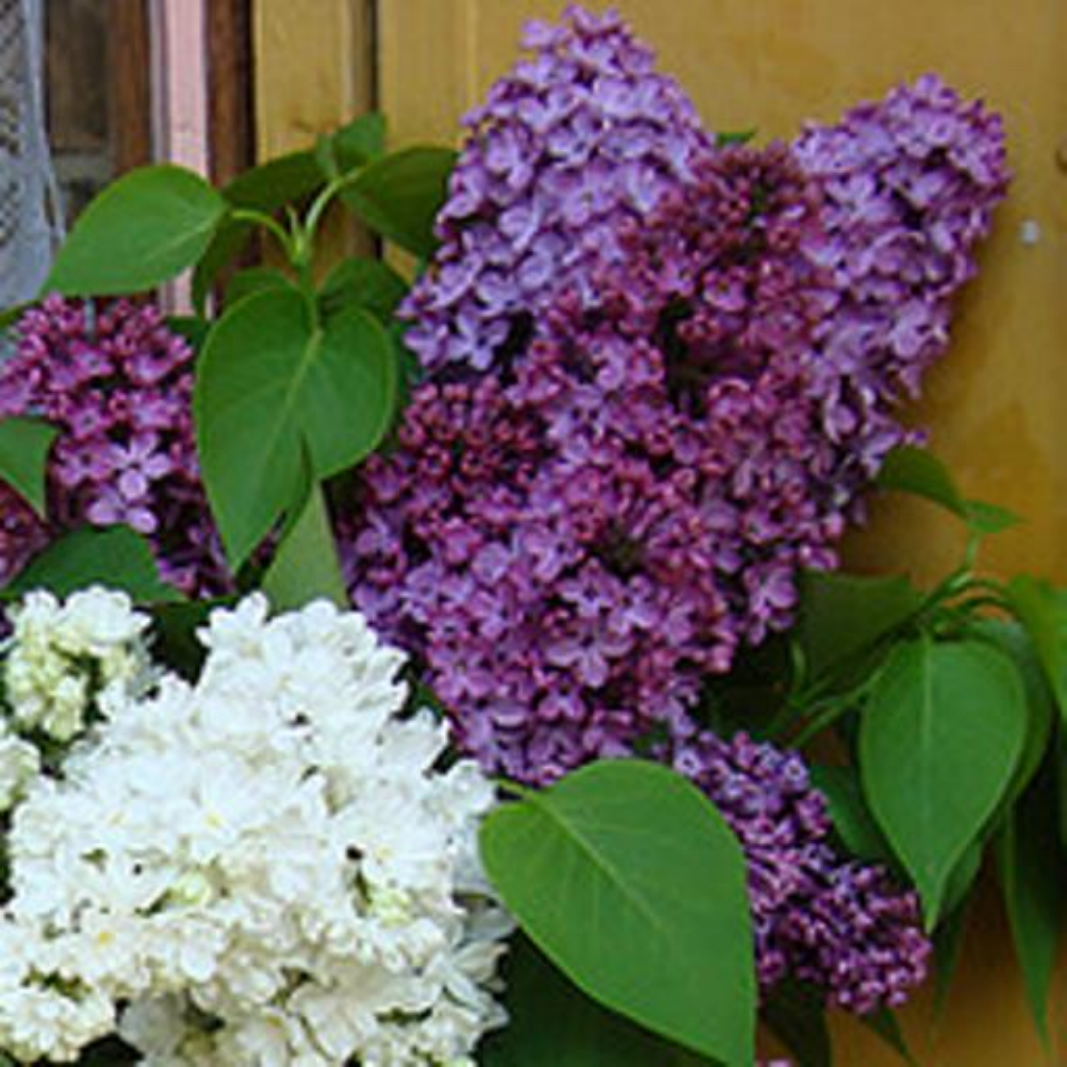}
  \caption{GT}
  \label{dataset_a}
\end{subfigure}
\begin{subfigure}{.09\linewidth}
  \centering
  \includegraphics[width=1.0\linewidth]{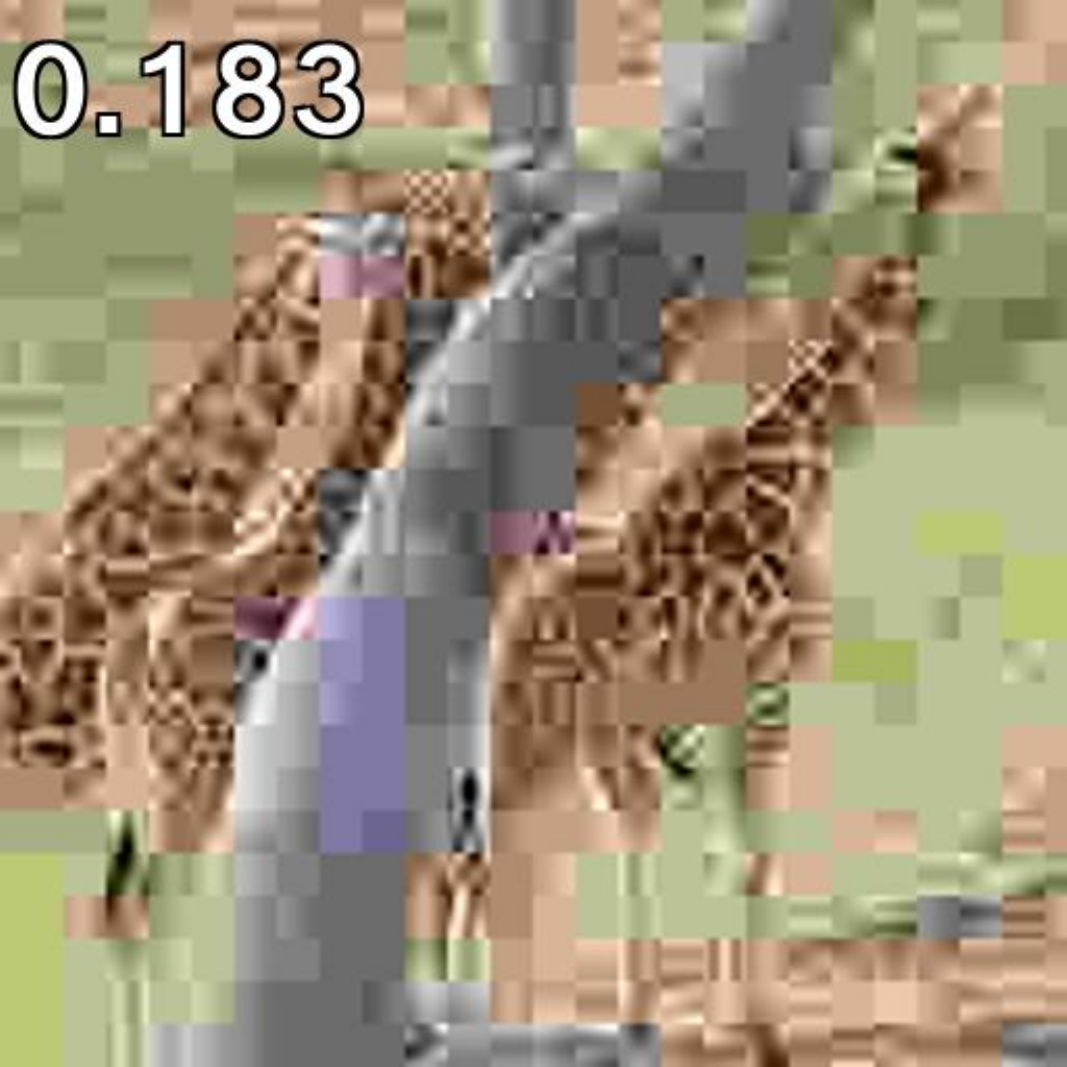}
  \includegraphics[width=1.0\linewidth]{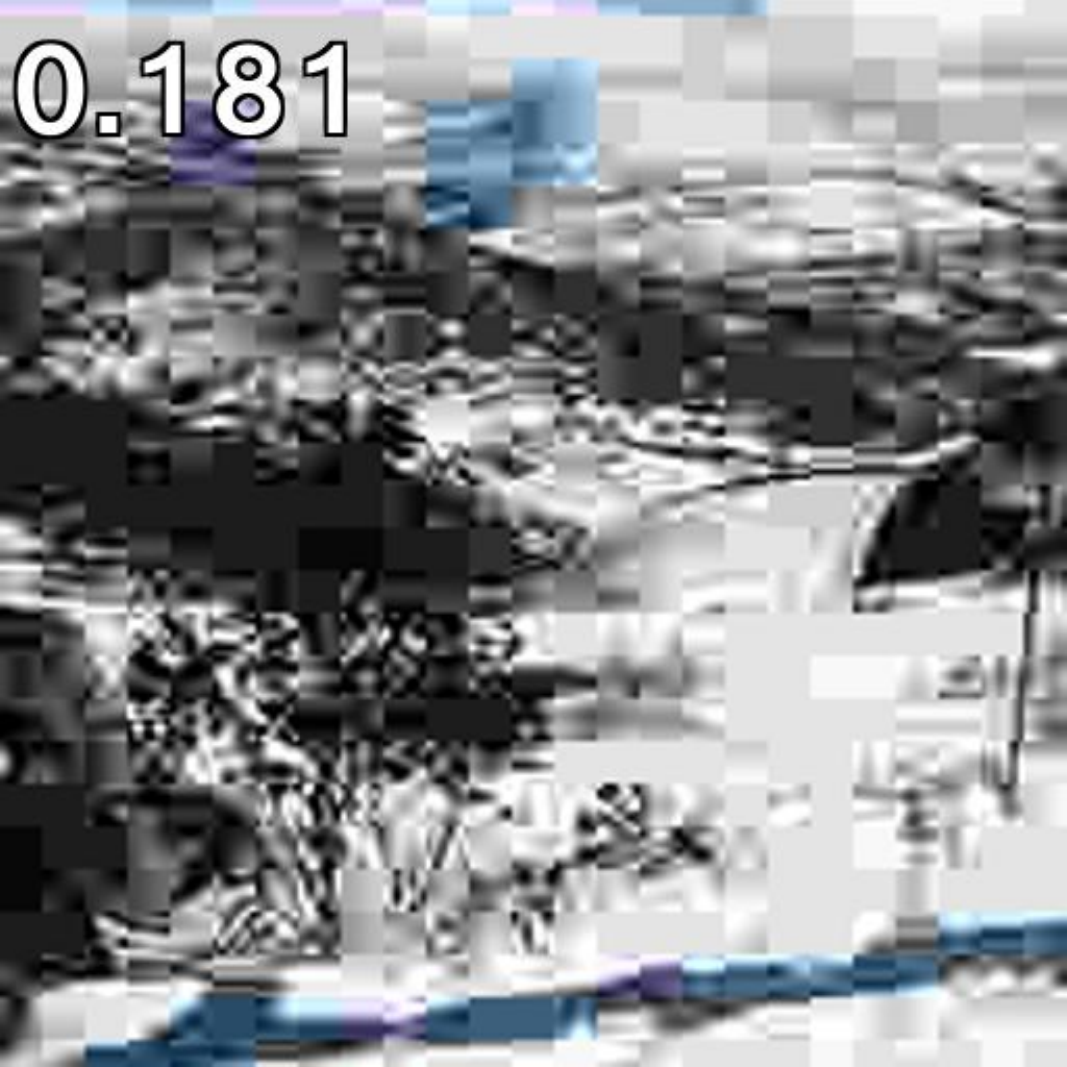}
  \includegraphics[width=1.0\linewidth]{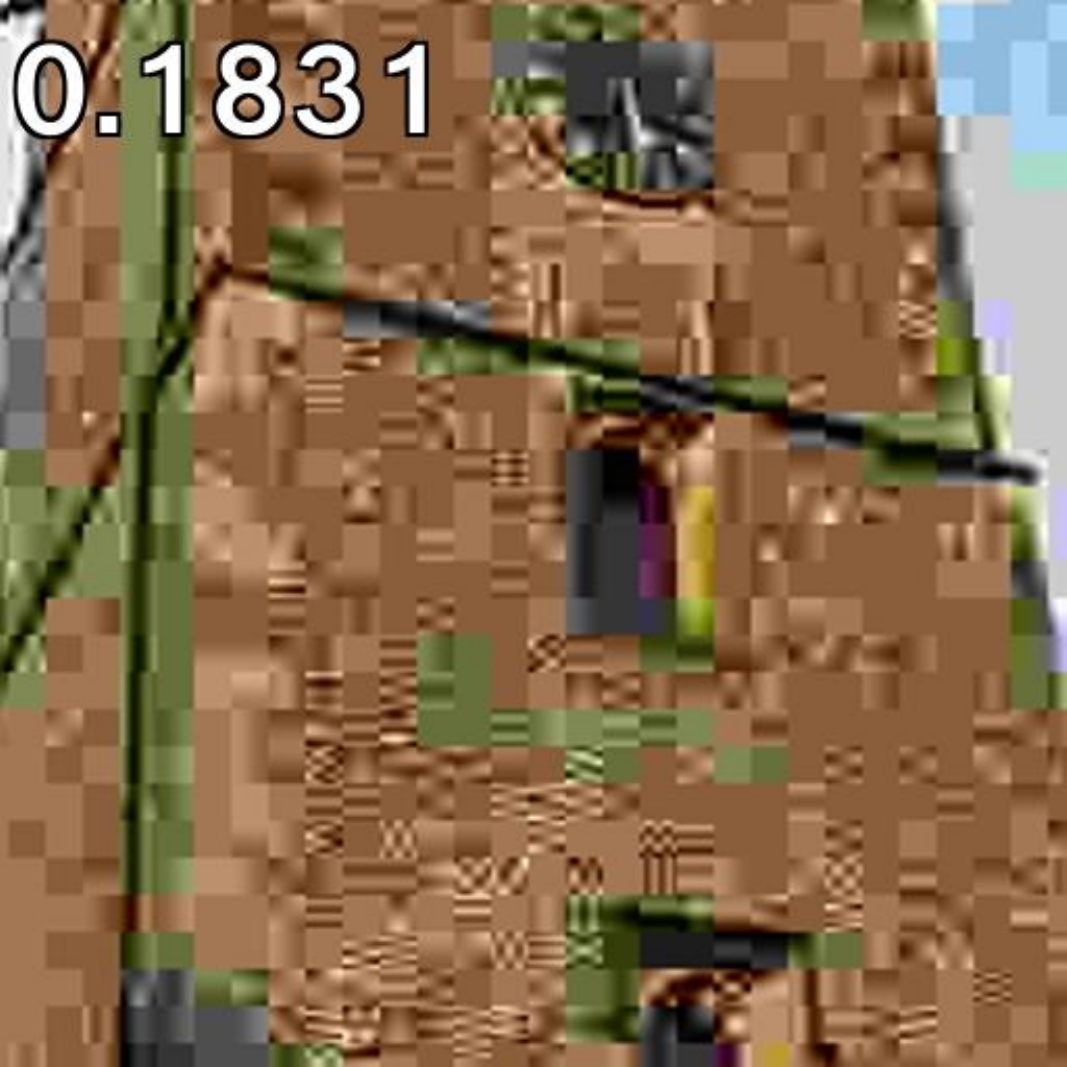}
  \includegraphics[width=1.0\linewidth]{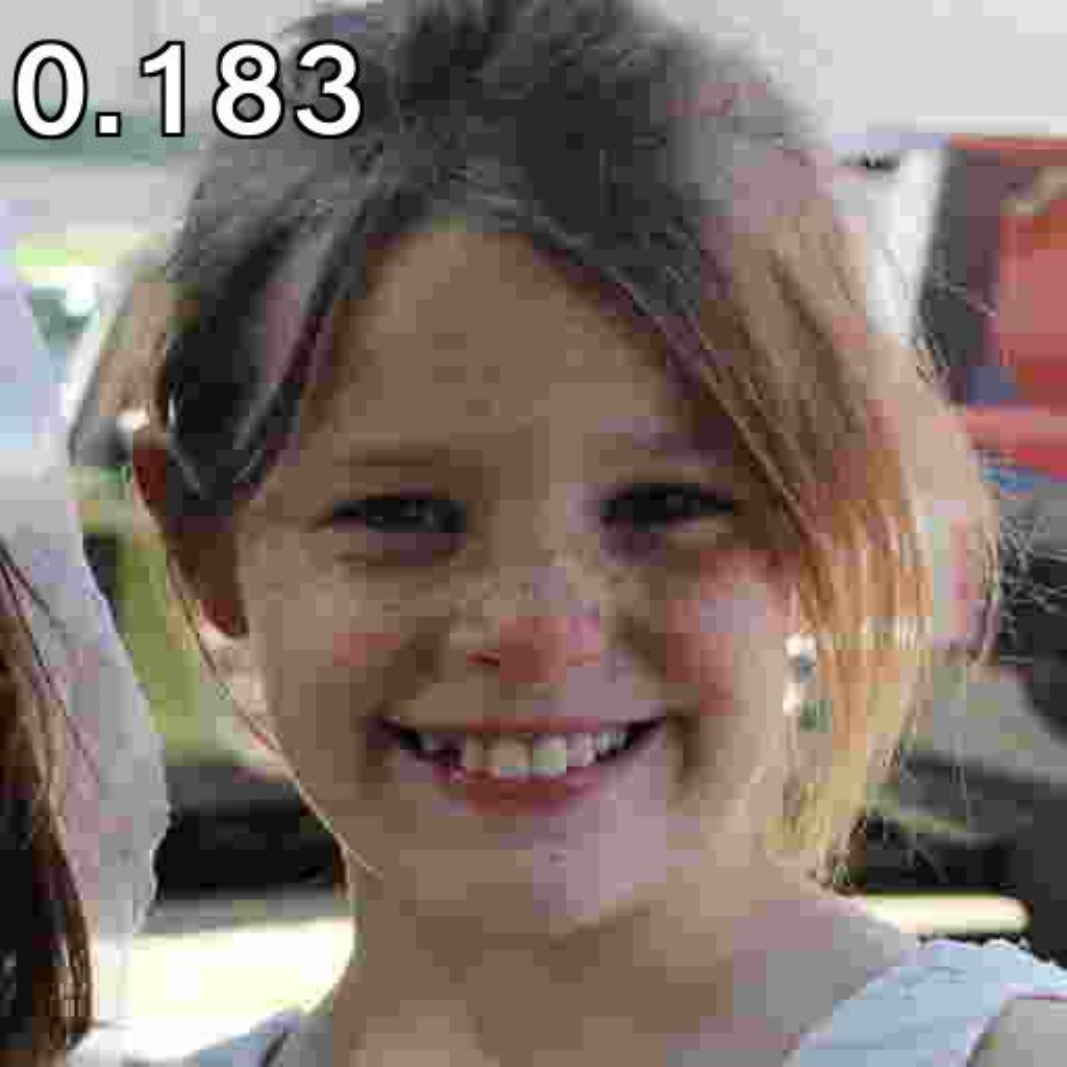}
         \includegraphics[width=1.0\linewidth]{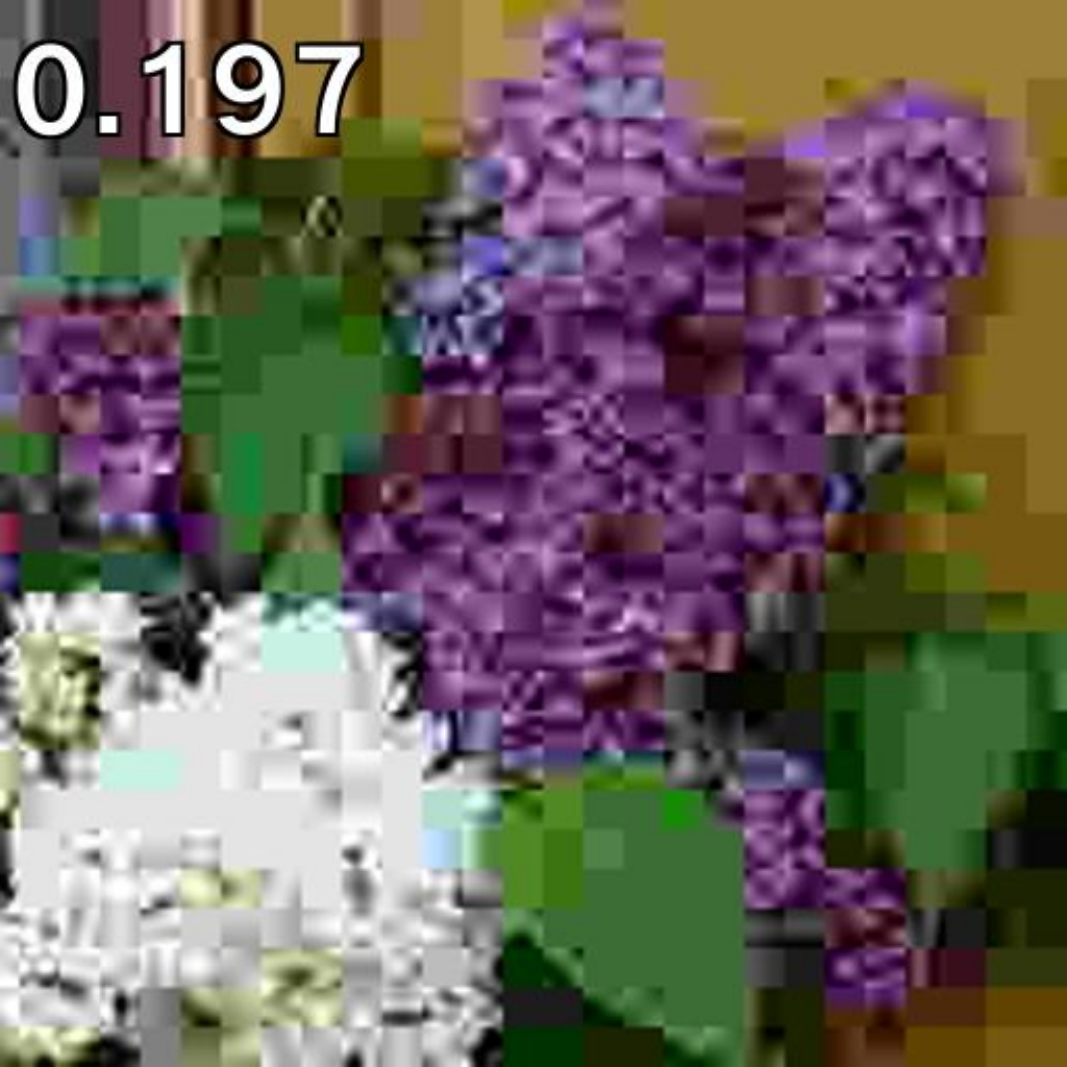}
  \caption{JPEG \cite{wallace1992jpeg}}
  \label{dataset_b}
\end{subfigure}
\begin{subfigure}{.09\linewidth}
  \centering
  \includegraphics[width=1.0\linewidth]{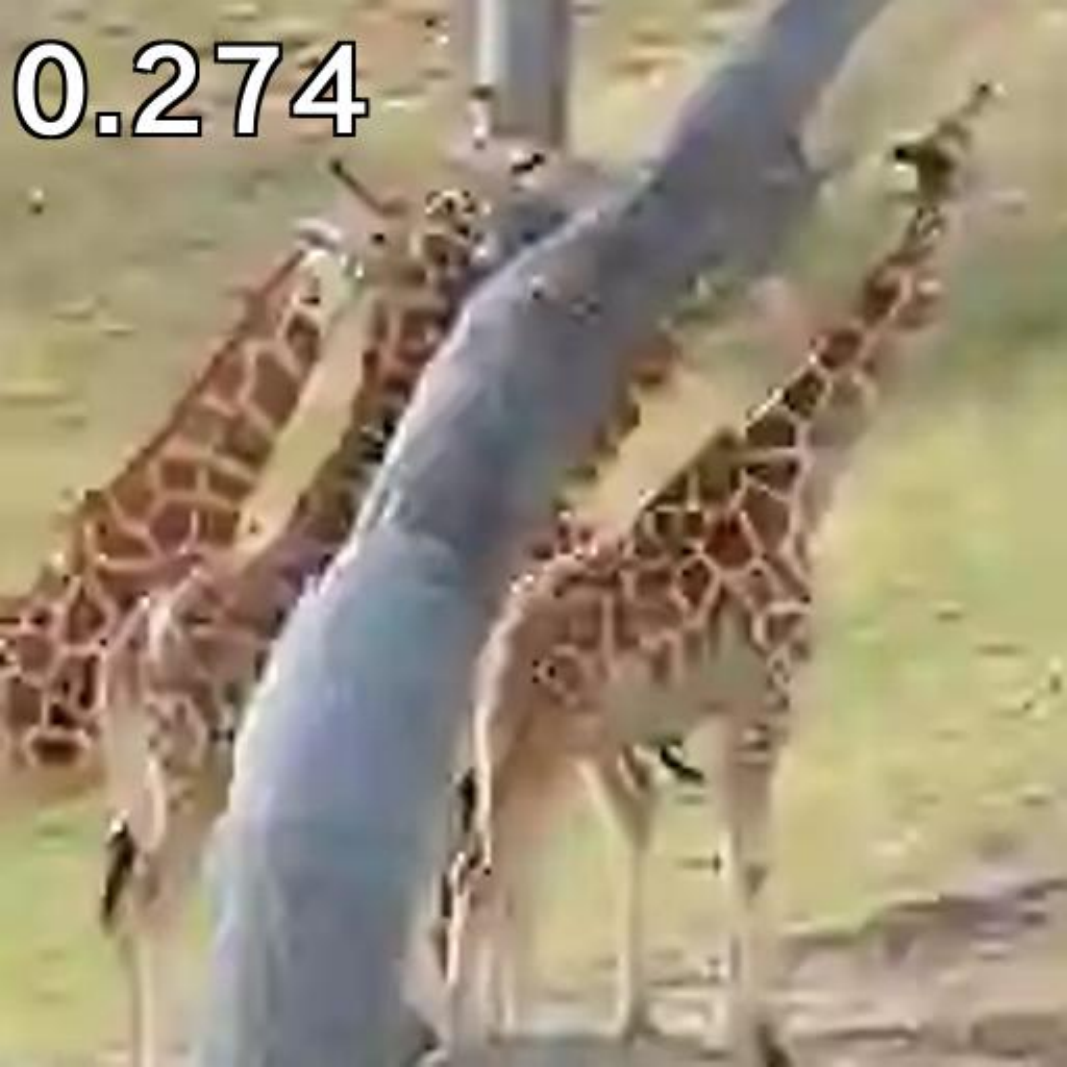}
  \includegraphics[width=1.0\linewidth]{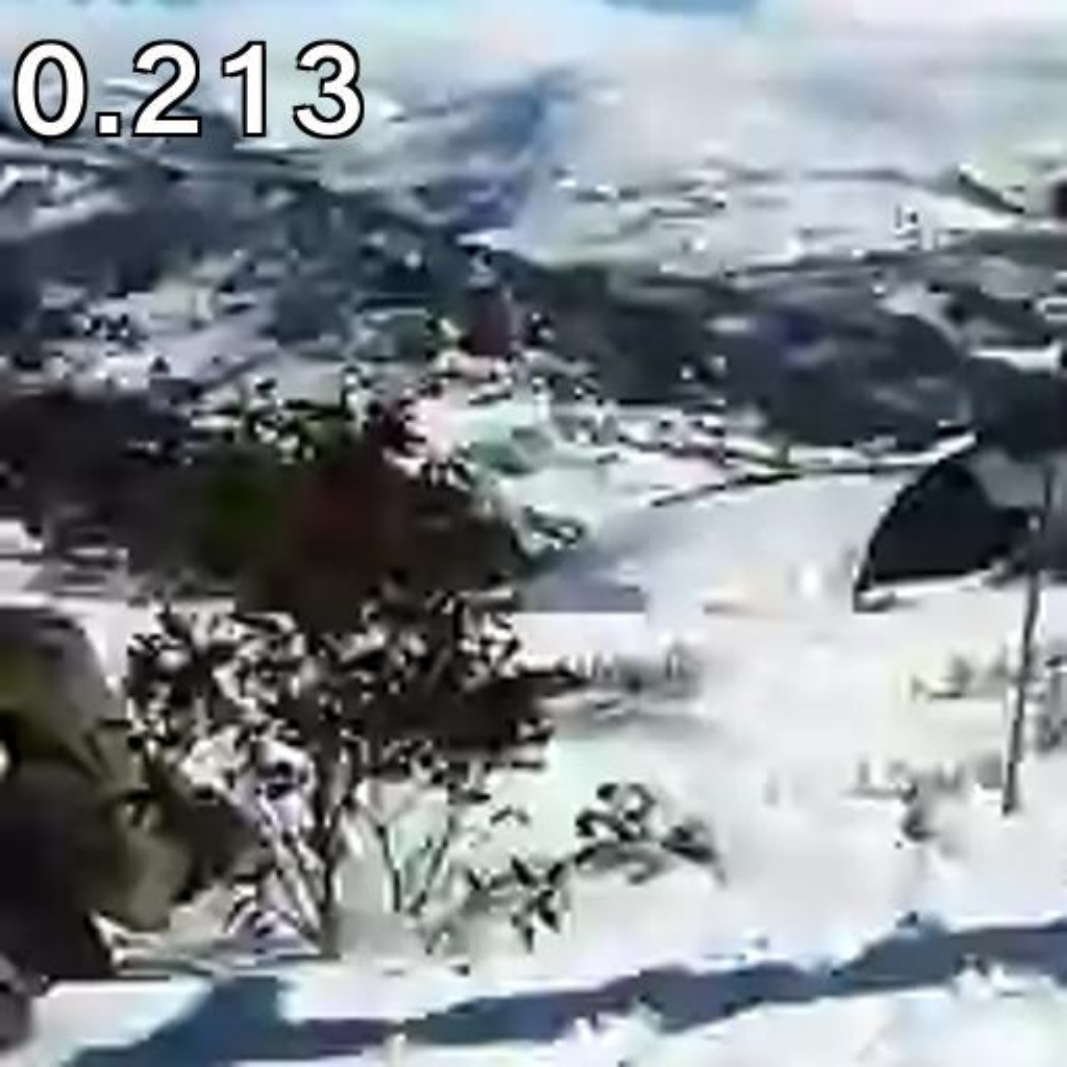}
  \includegraphics[width=1.0\linewidth]{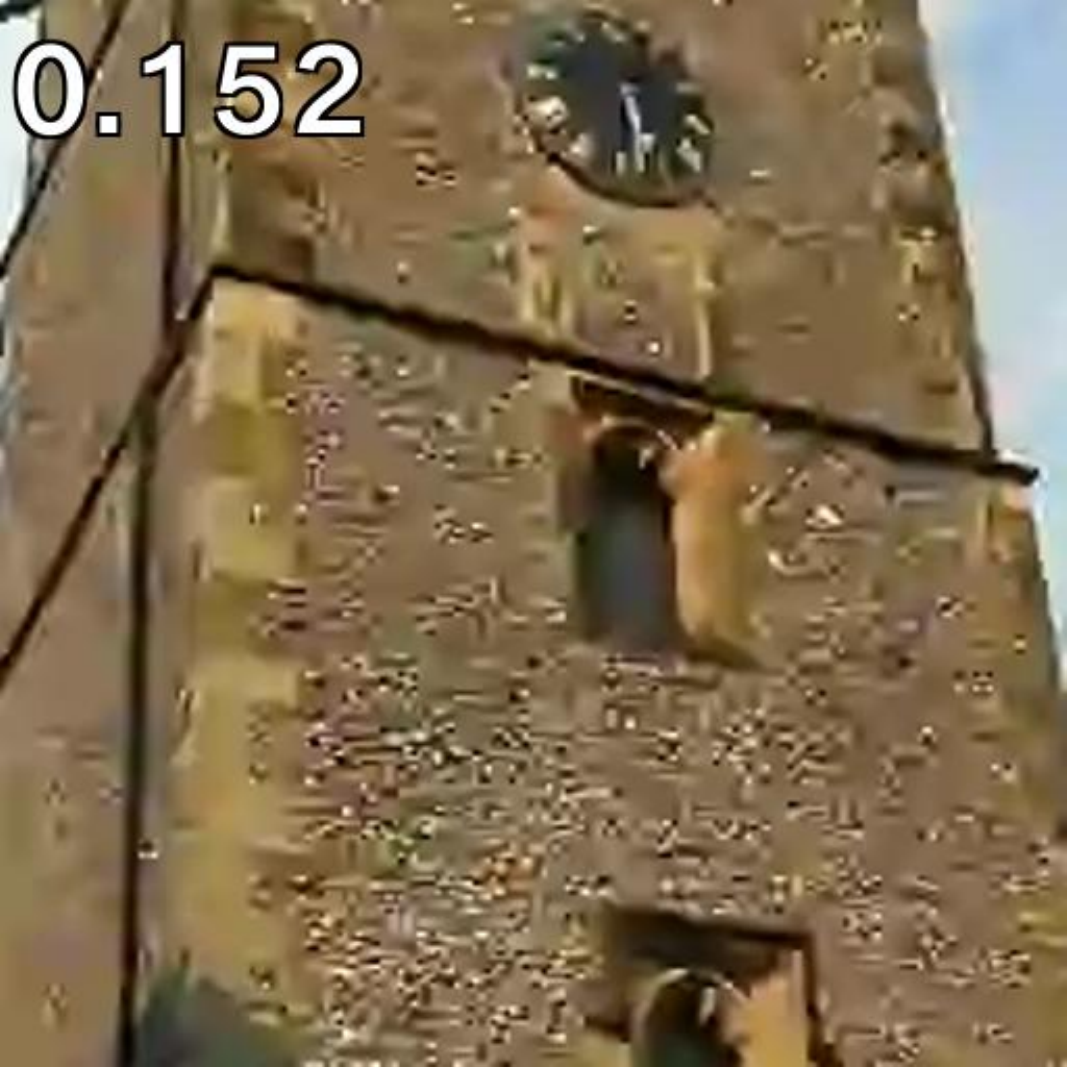}
  \includegraphics[width=1.0\linewidth]{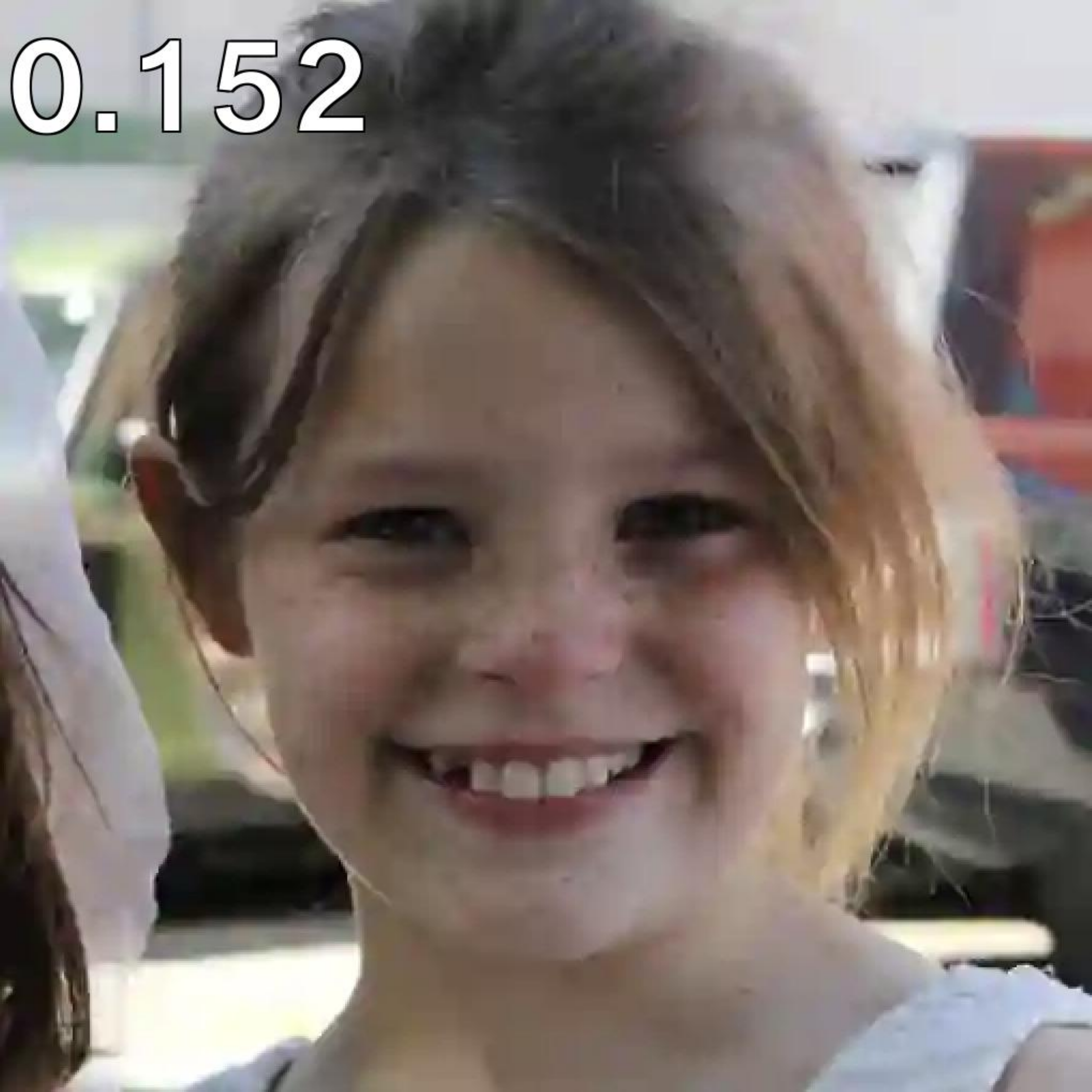}
   \includegraphics[width=1.0\linewidth]{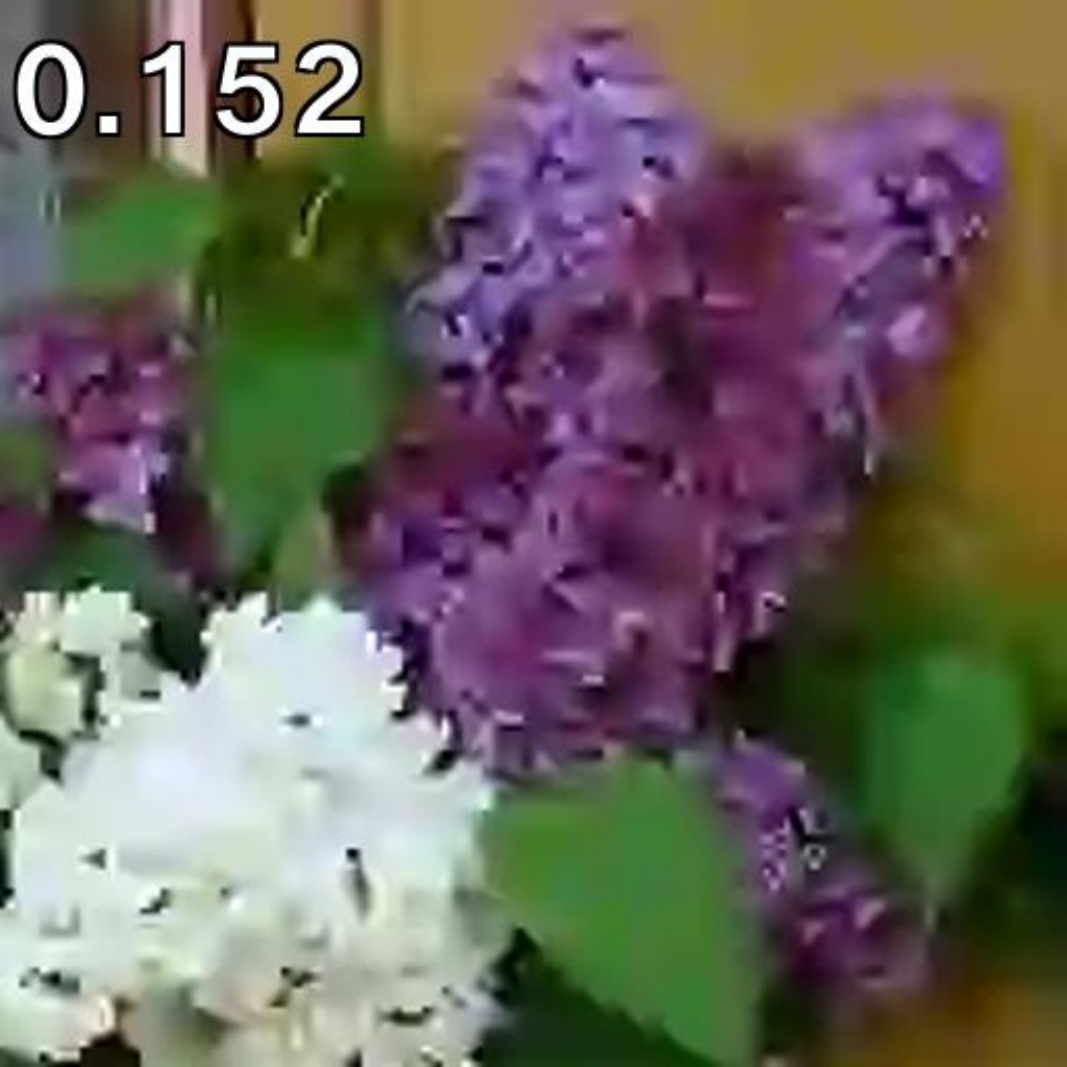}
  \caption{WebP\cite{mukherjee2014webp}}
  \label{dataset_c}
\end{subfigure}
\begin{subfigure}{.09\linewidth}
  \centering
  \includegraphics[width=1.0\linewidth]{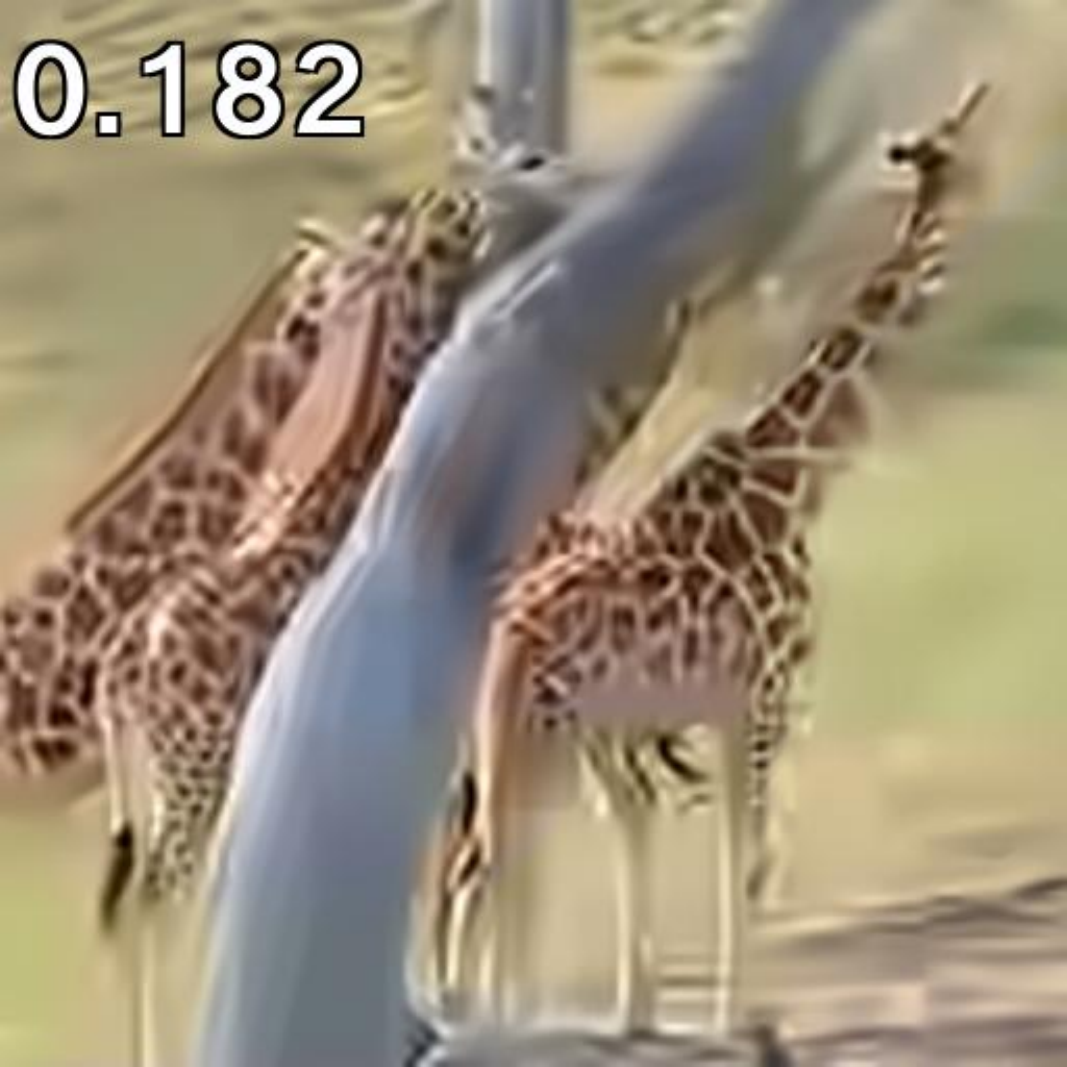} 
  \includegraphics[width=1.0\linewidth]{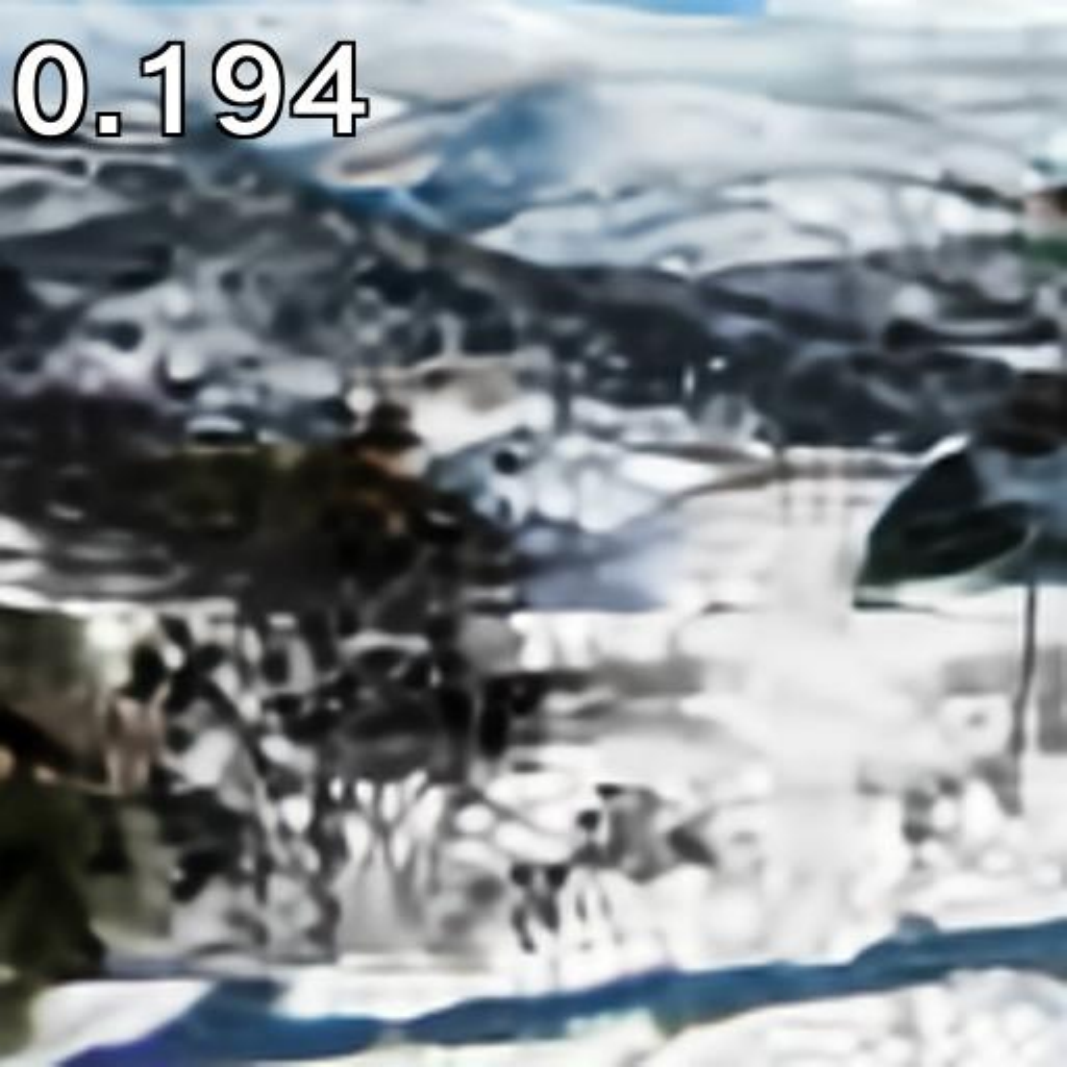} 
  \includegraphics[width=1.0\linewidth]{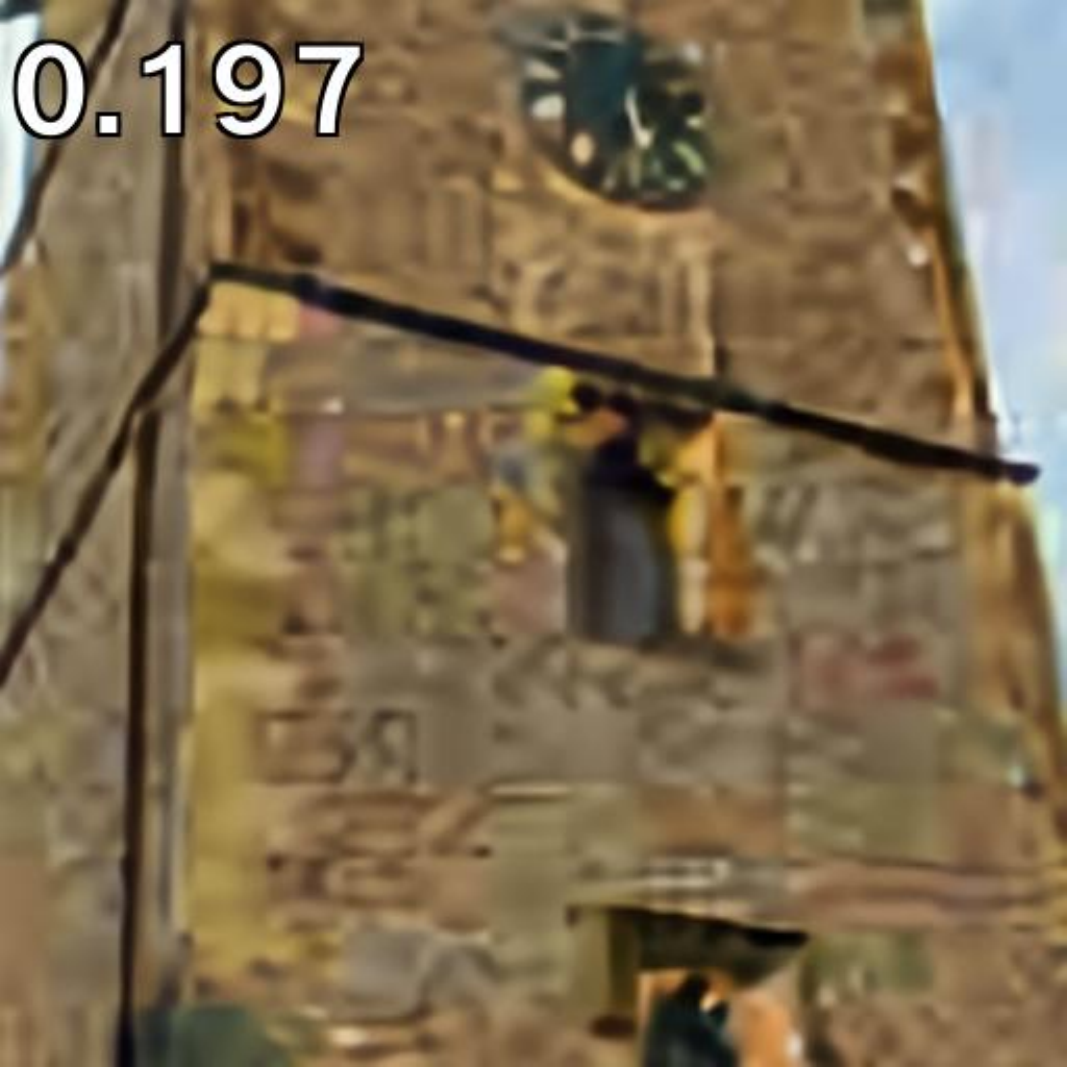}
  \includegraphics[width=1.0\linewidth]{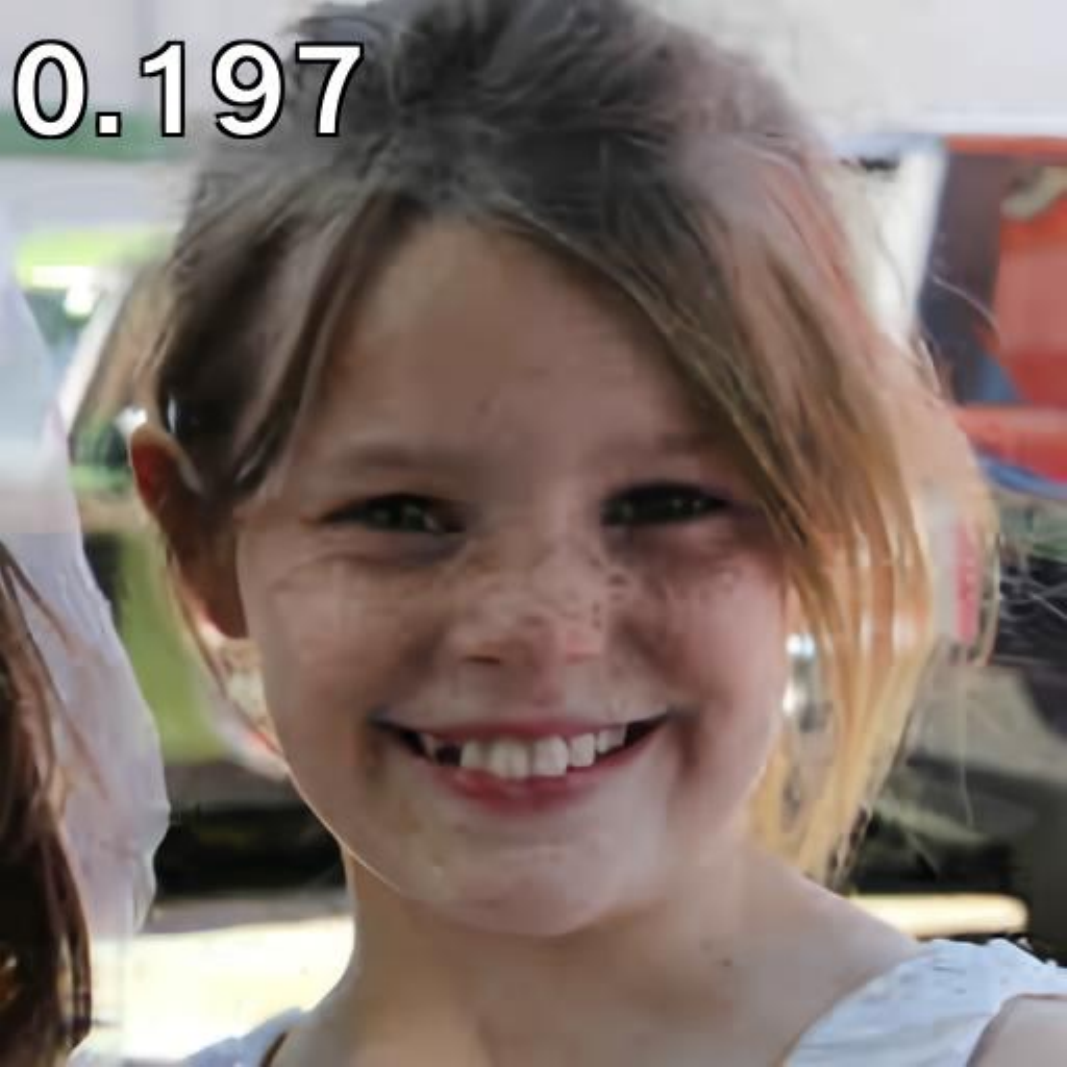}
  \includegraphics[width=1.0\linewidth]{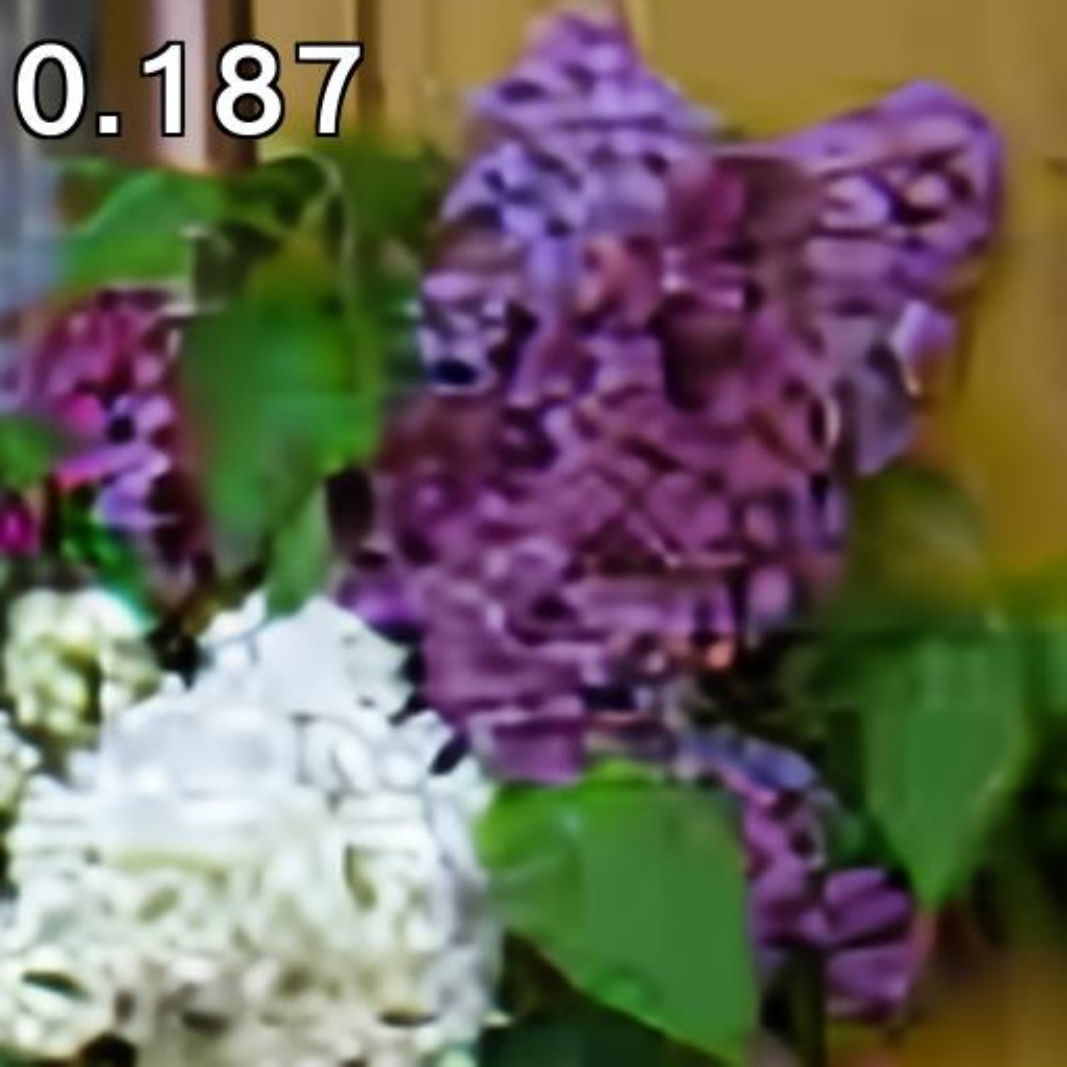}
  \caption{IRN \cite{xiao2023invertible}}
  \label{dataset_d}
\end{subfigure}
\begin{subfigure}{.09\linewidth}
  \centering
 \includegraphics[width=1.0\linewidth]{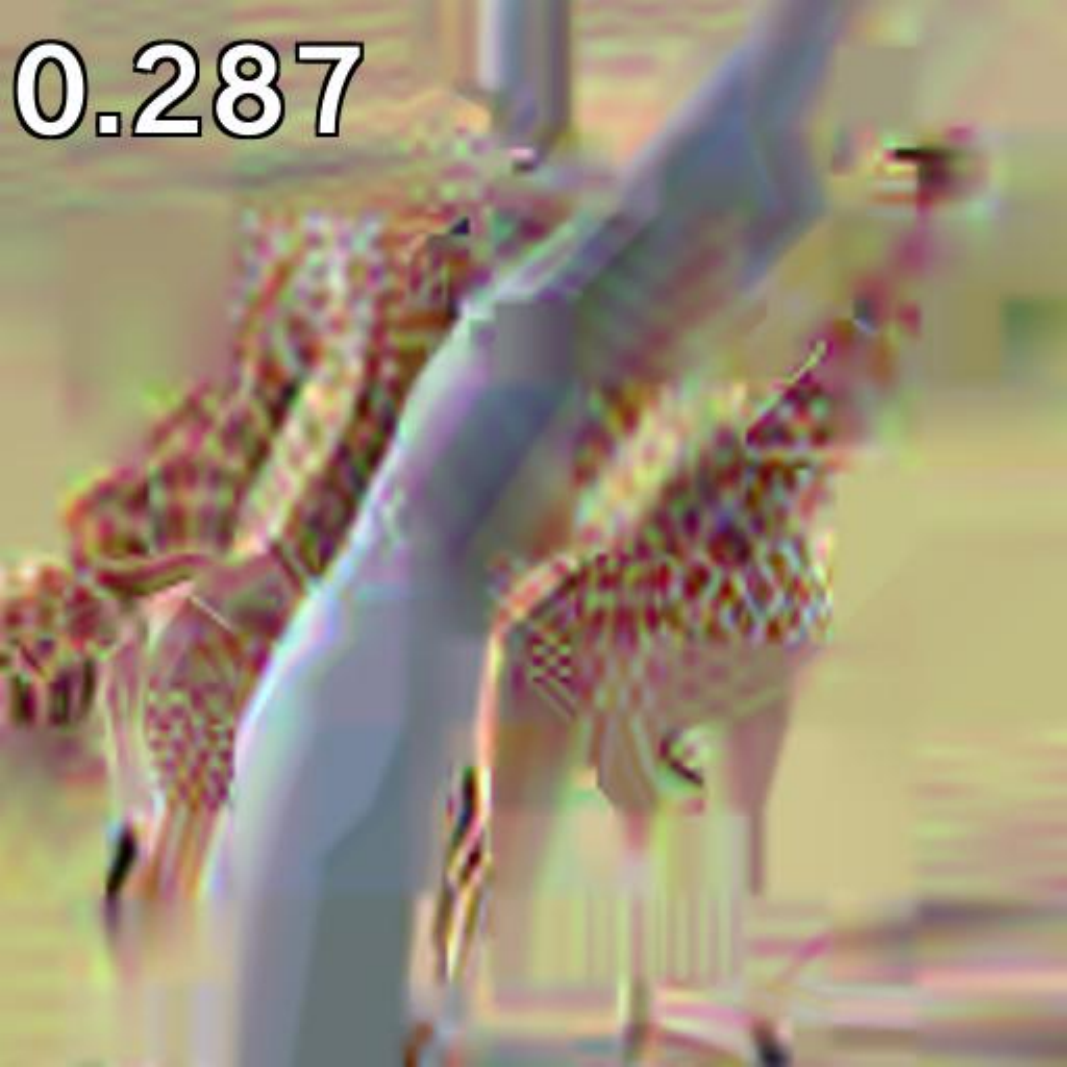}
 \includegraphics[width=1.0\linewidth]{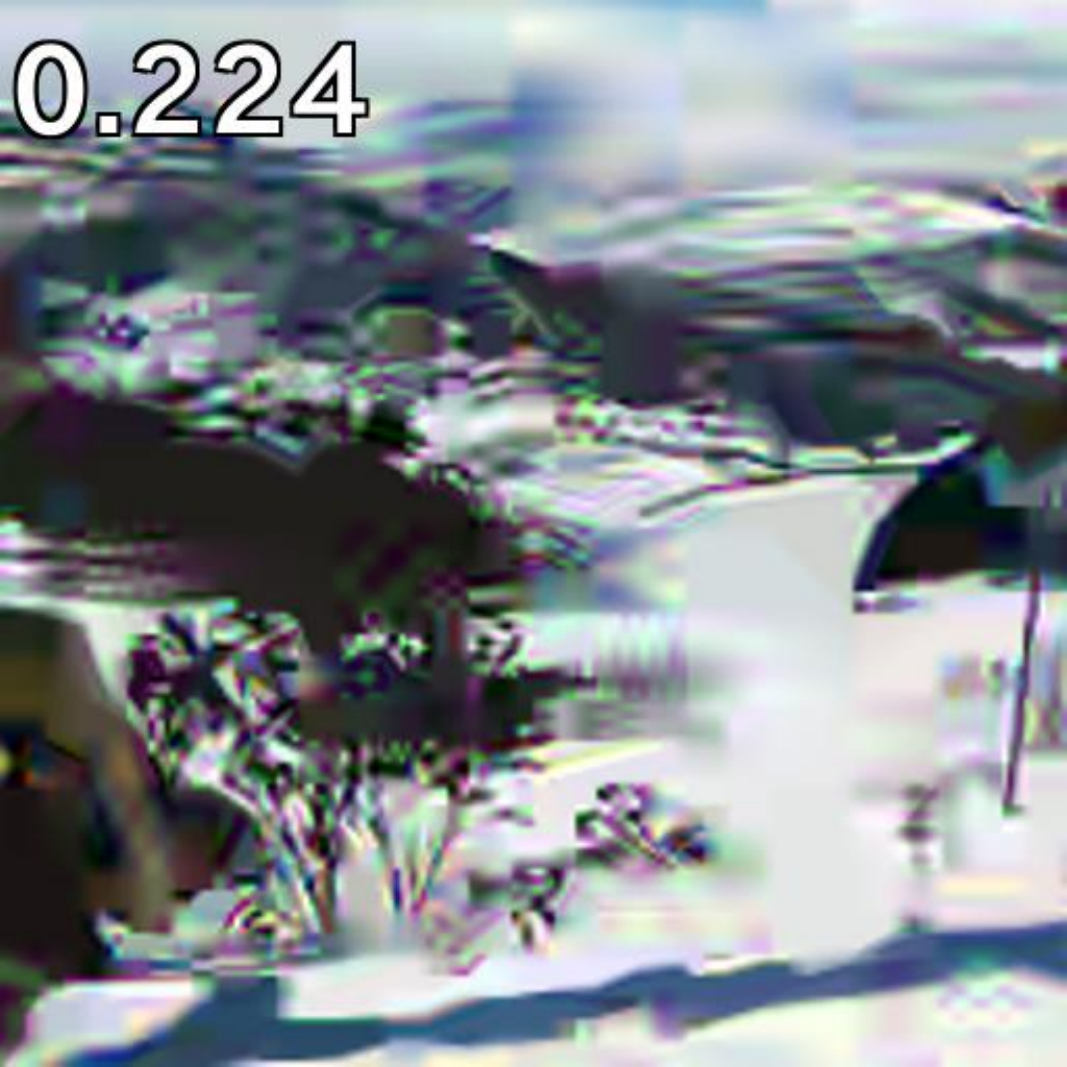} 
   \includegraphics[width=1.0\linewidth]{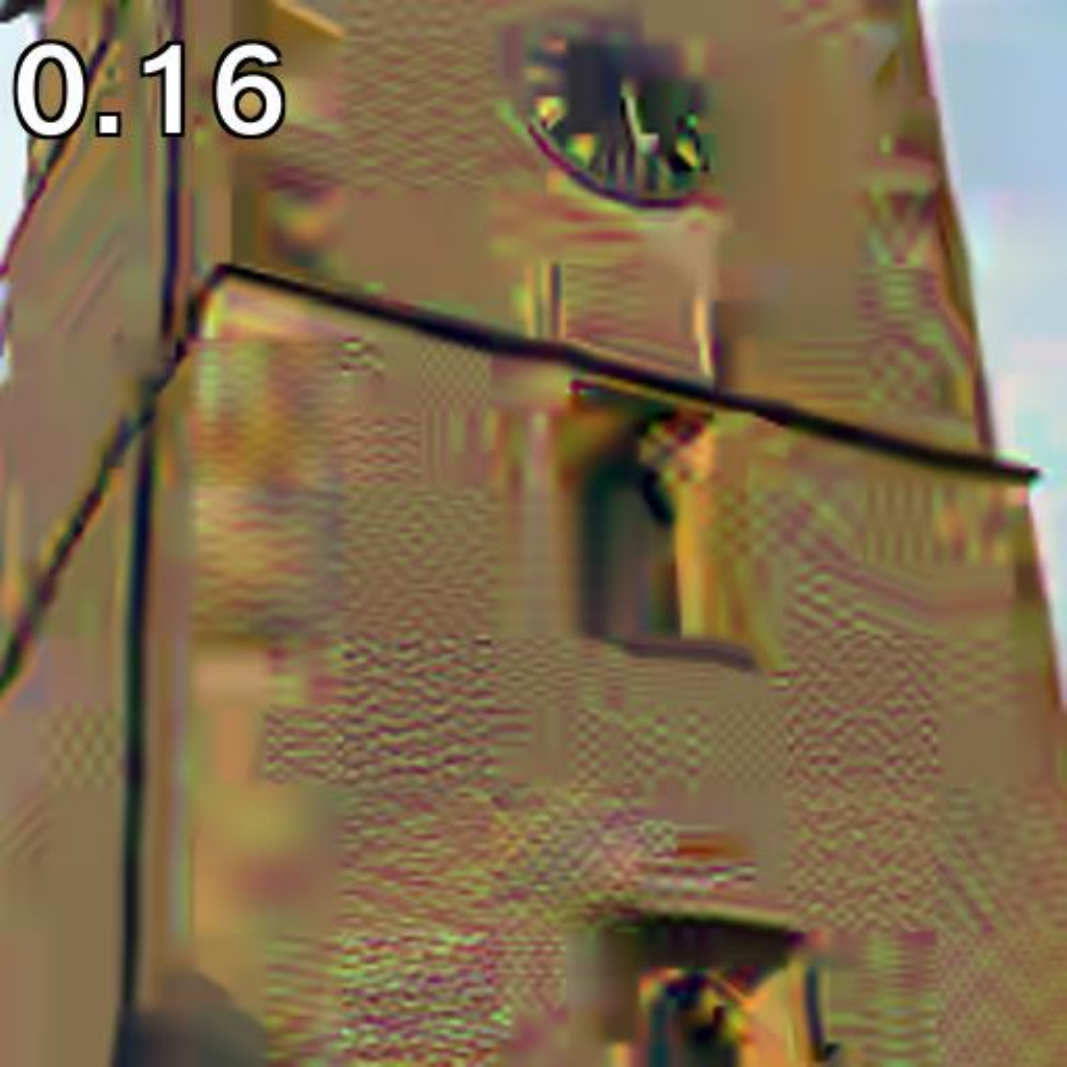}
   \includegraphics[width=1.0\linewidth]{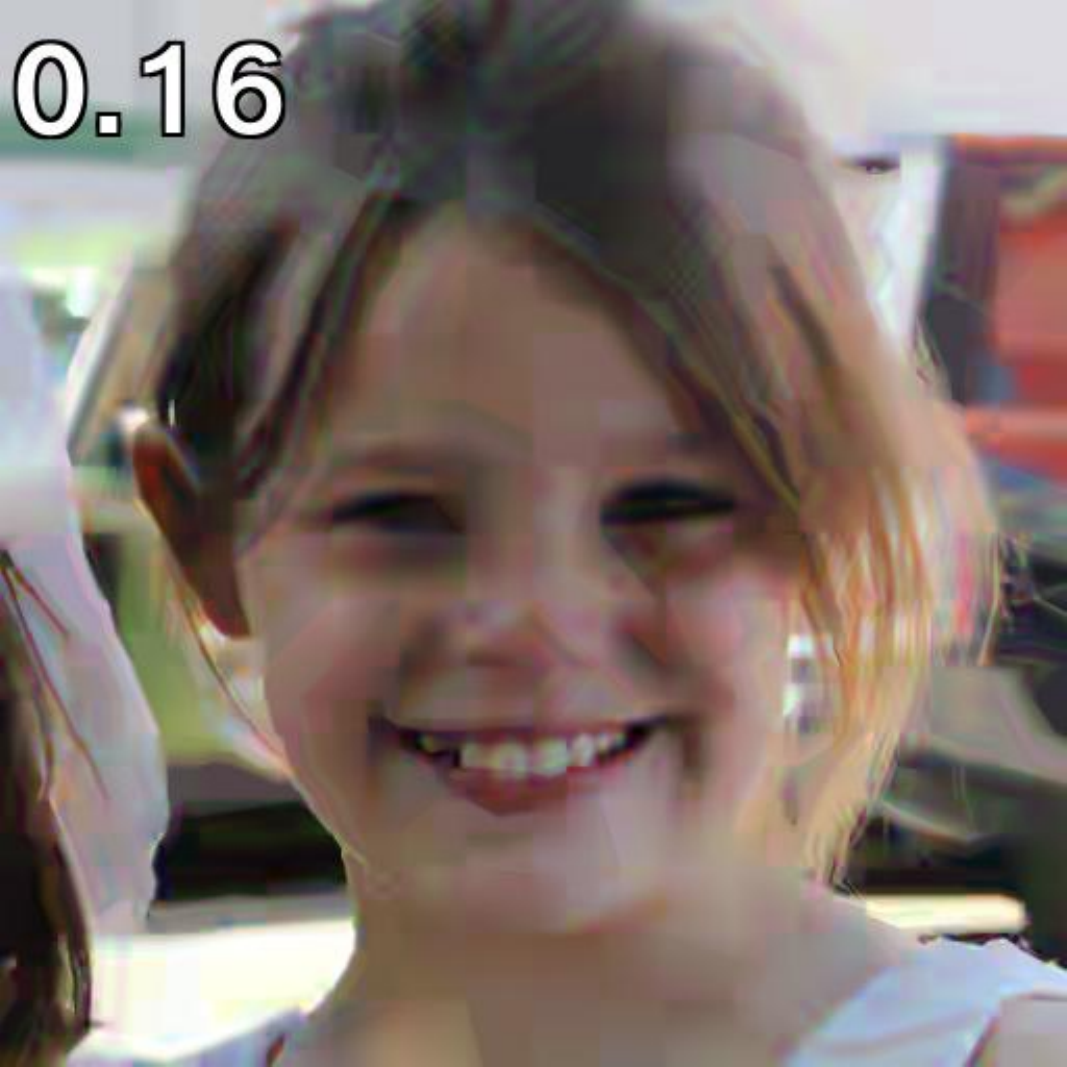}
       \includegraphics[width=1.0\linewidth]{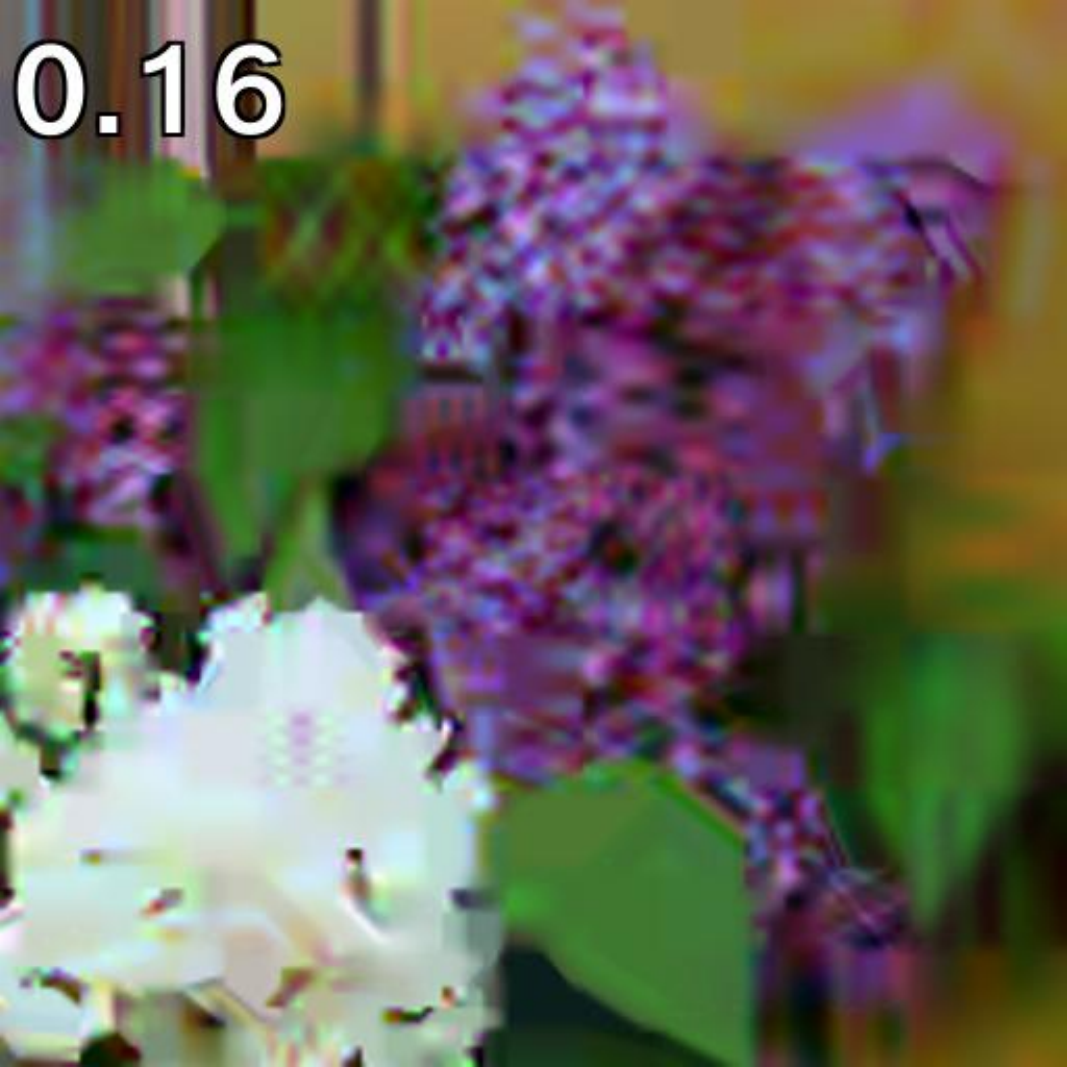}
 \caption{x265 \cite{ramachandran2013x265}}
  \label{dataset_e}
\end{subfigure}
\begin{subfigure}{.09\linewidth}
  \centering
  \includegraphics[width=1.0\linewidth]{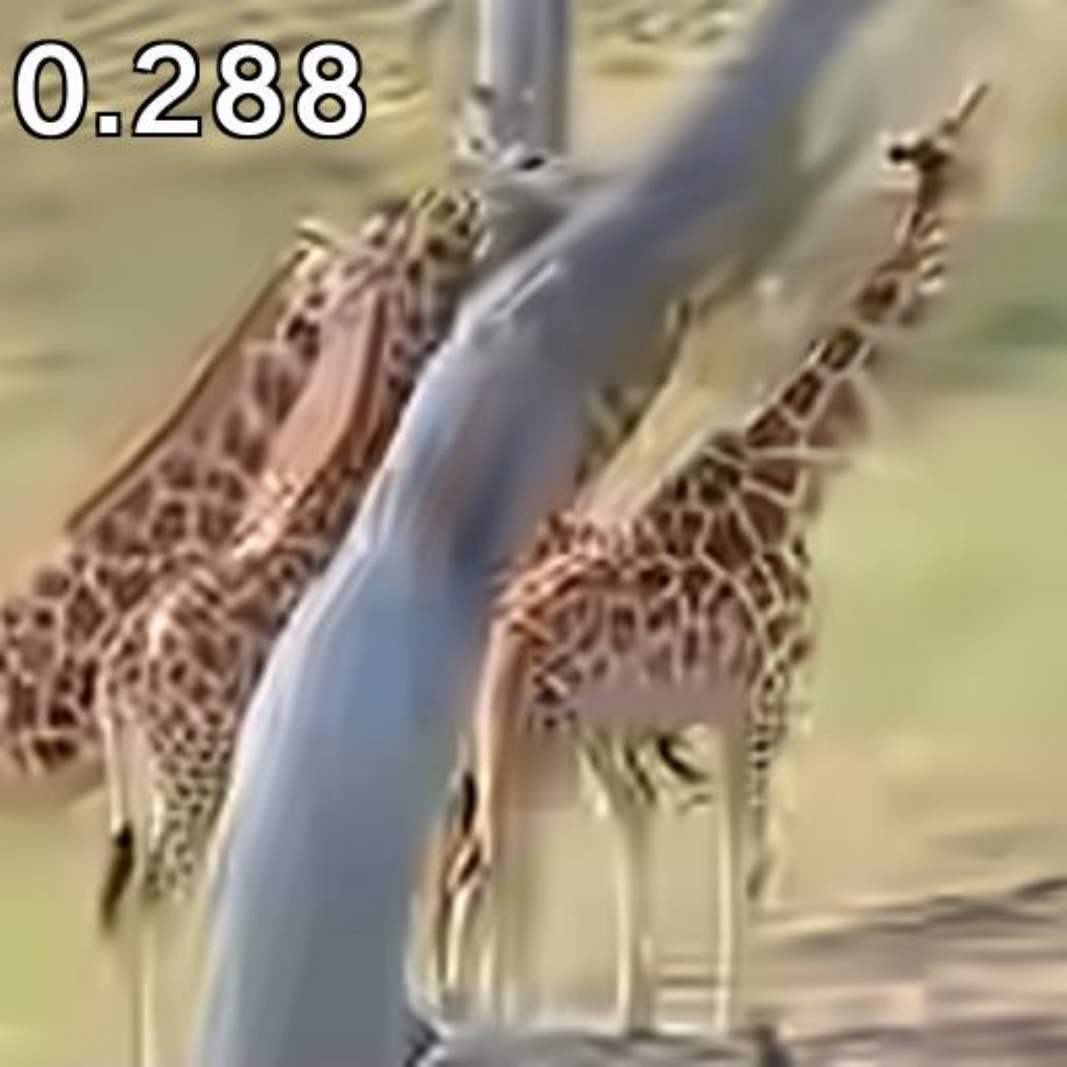}
  \includegraphics[width=1.0\linewidth]{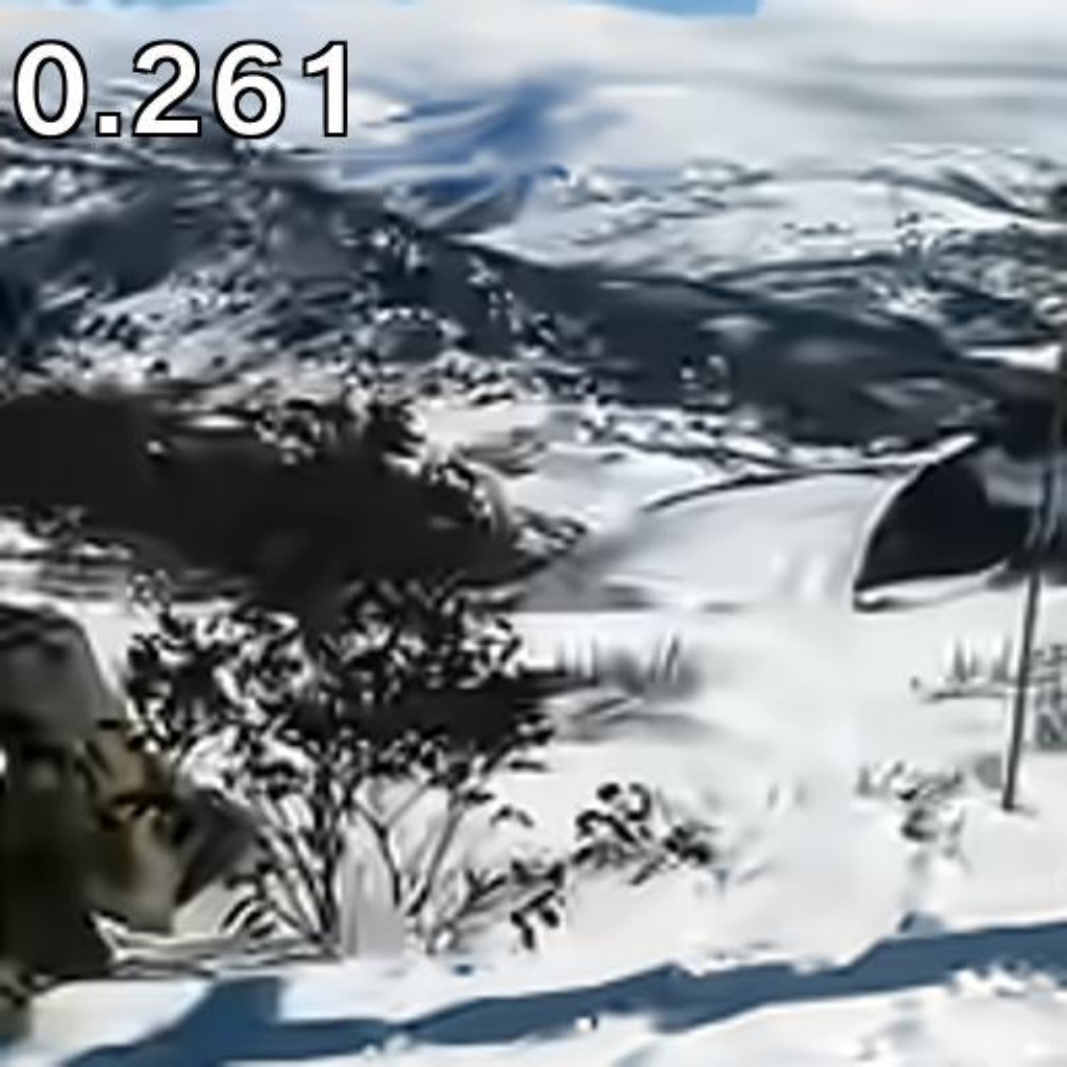}
    \includegraphics[width=1.0\linewidth]{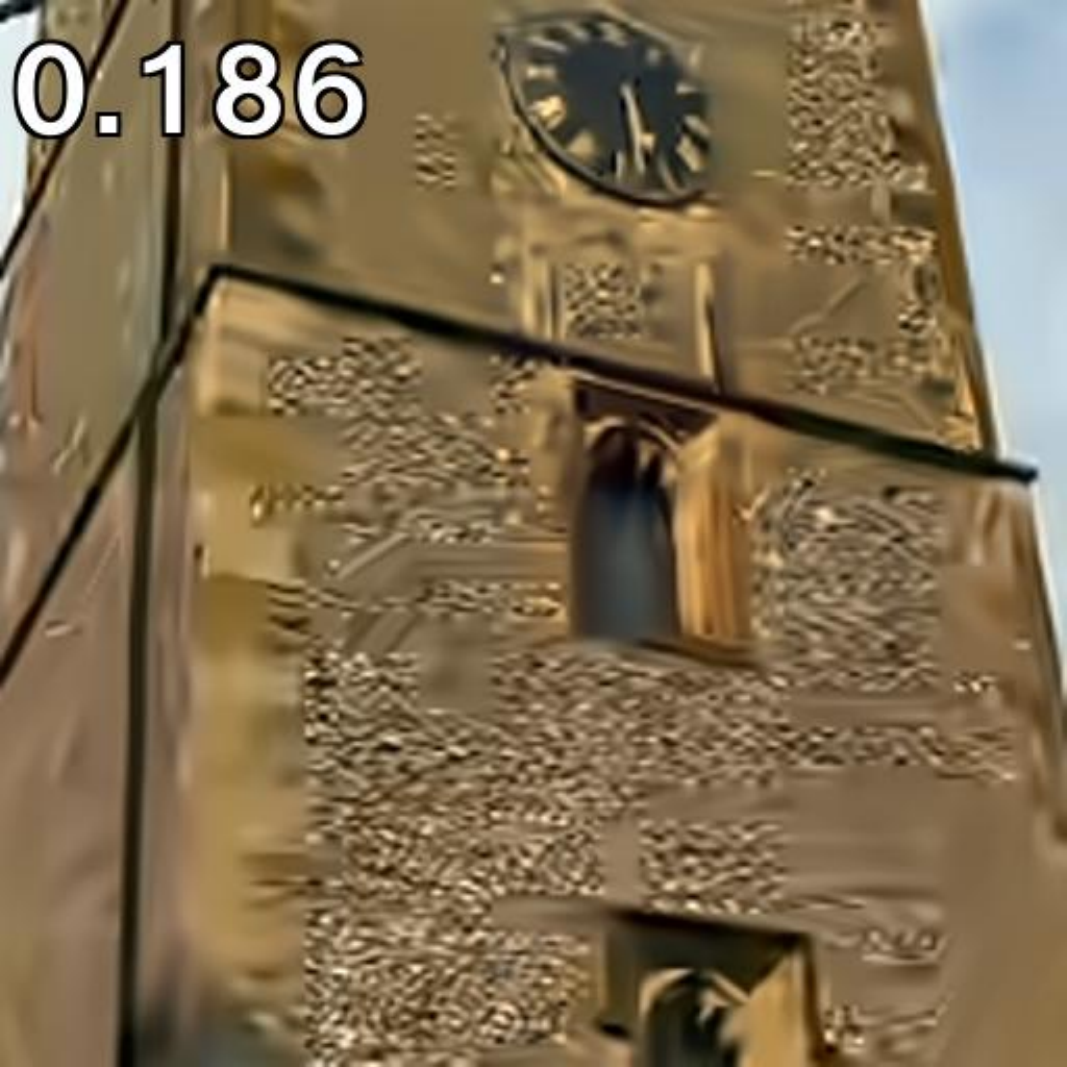}
    \includegraphics[width=1.0\linewidth]{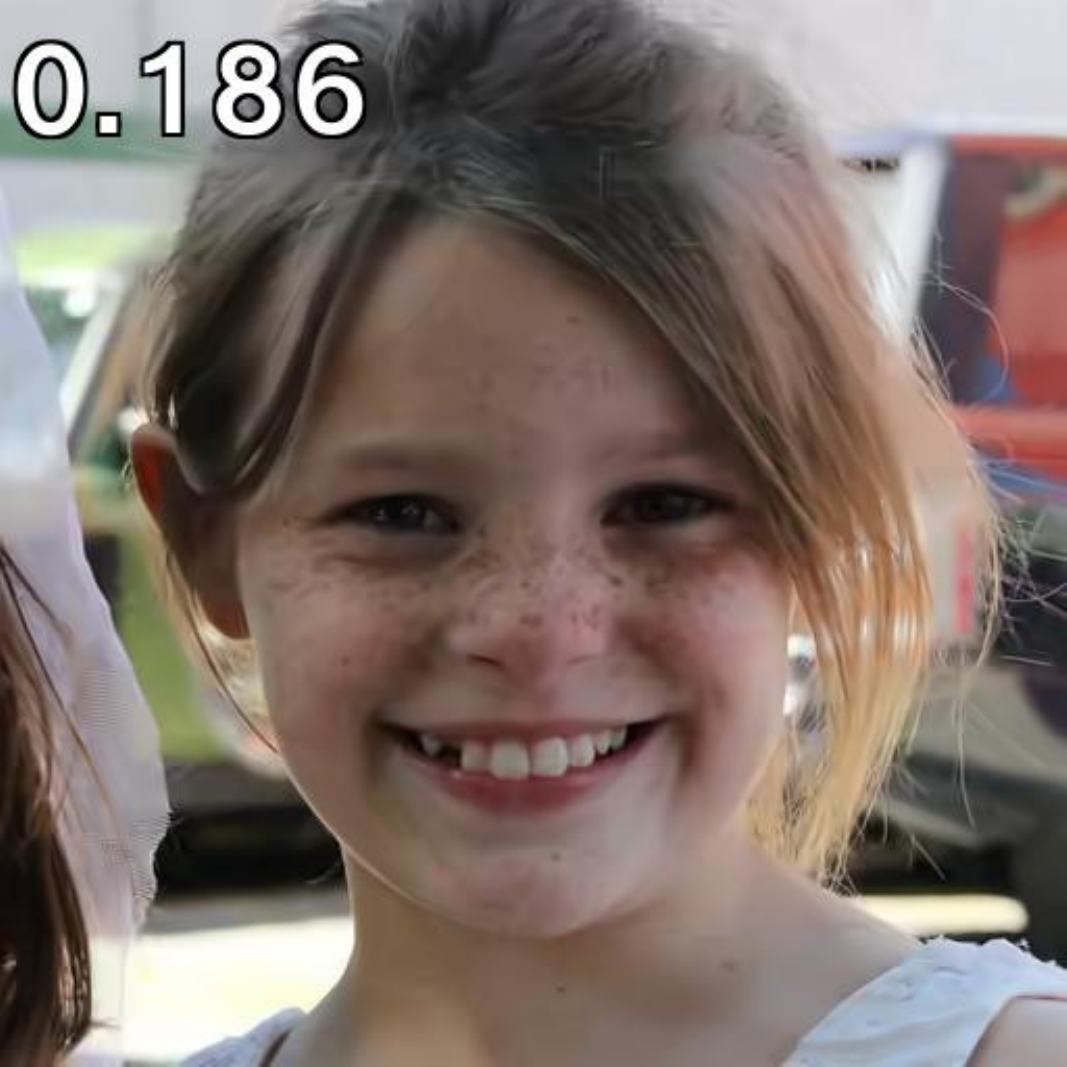}
          \includegraphics[width=1.0\linewidth]{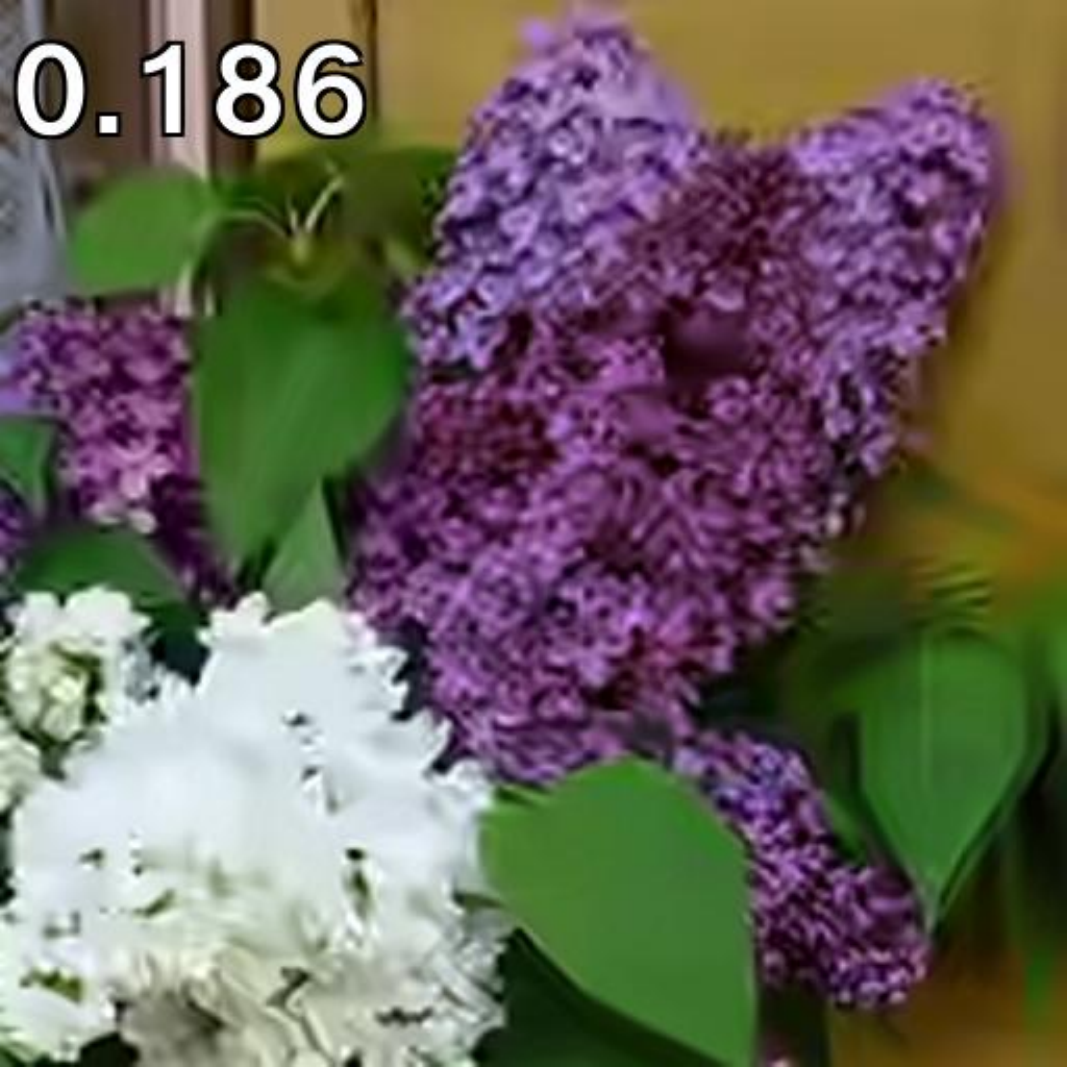}
  \caption{VTM \cite{bross2021overview}}
  \label{dataset_f}
\end{subfigure}
\begin{subfigure}{.09\linewidth}
  \centering
  \includegraphics[width=1.0\linewidth]{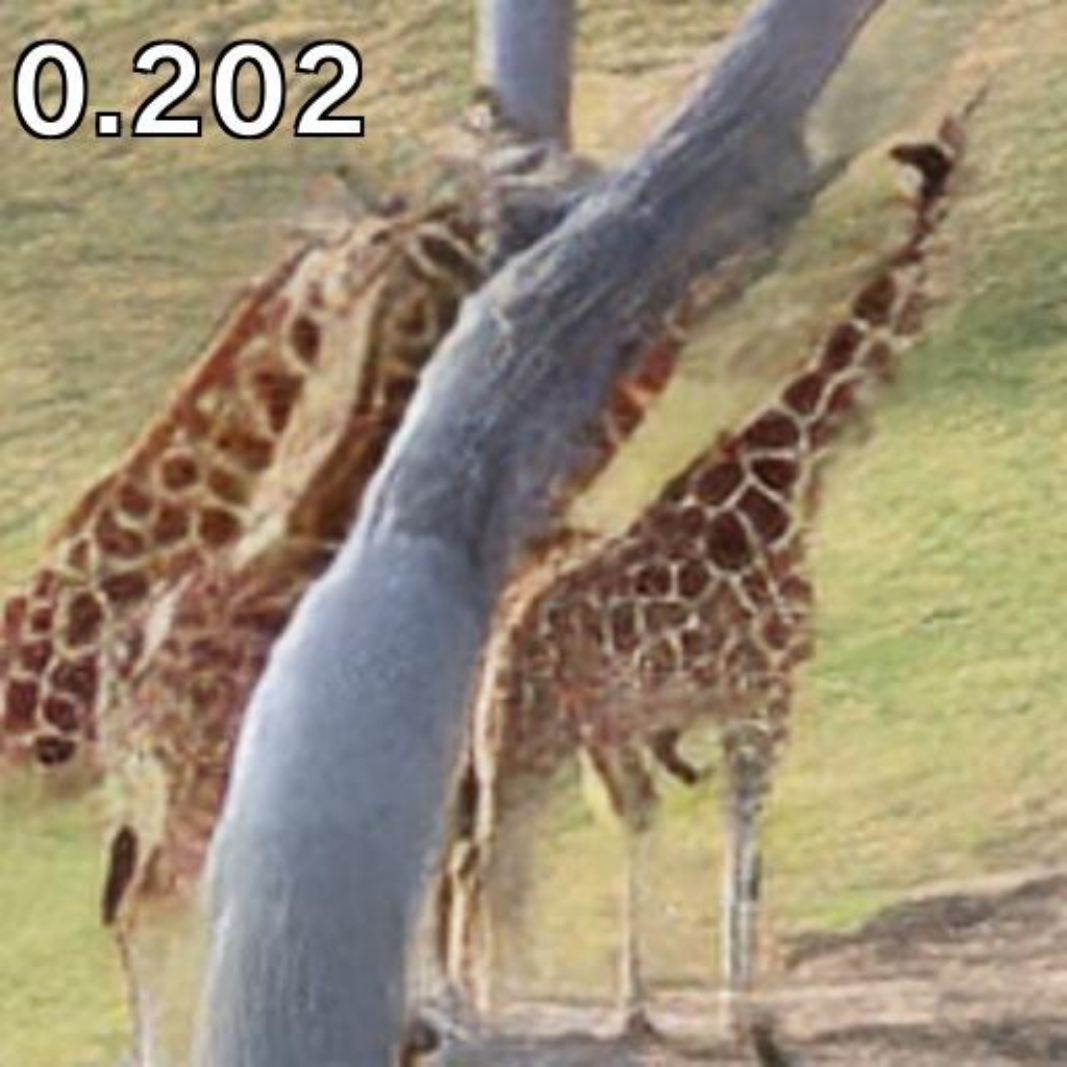}
  \includegraphics[width=1.0\linewidth]{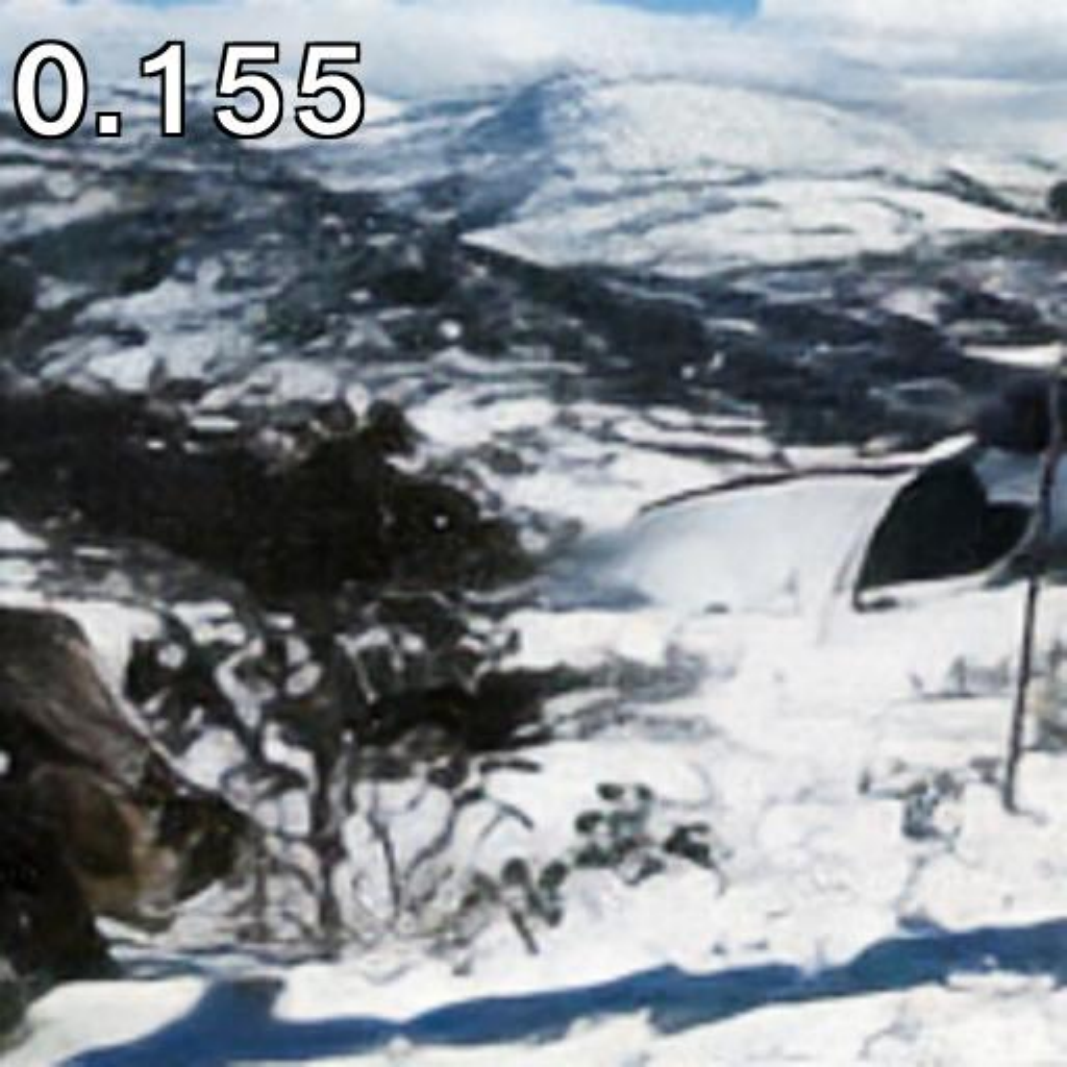}
    \includegraphics[width=1.0\linewidth]{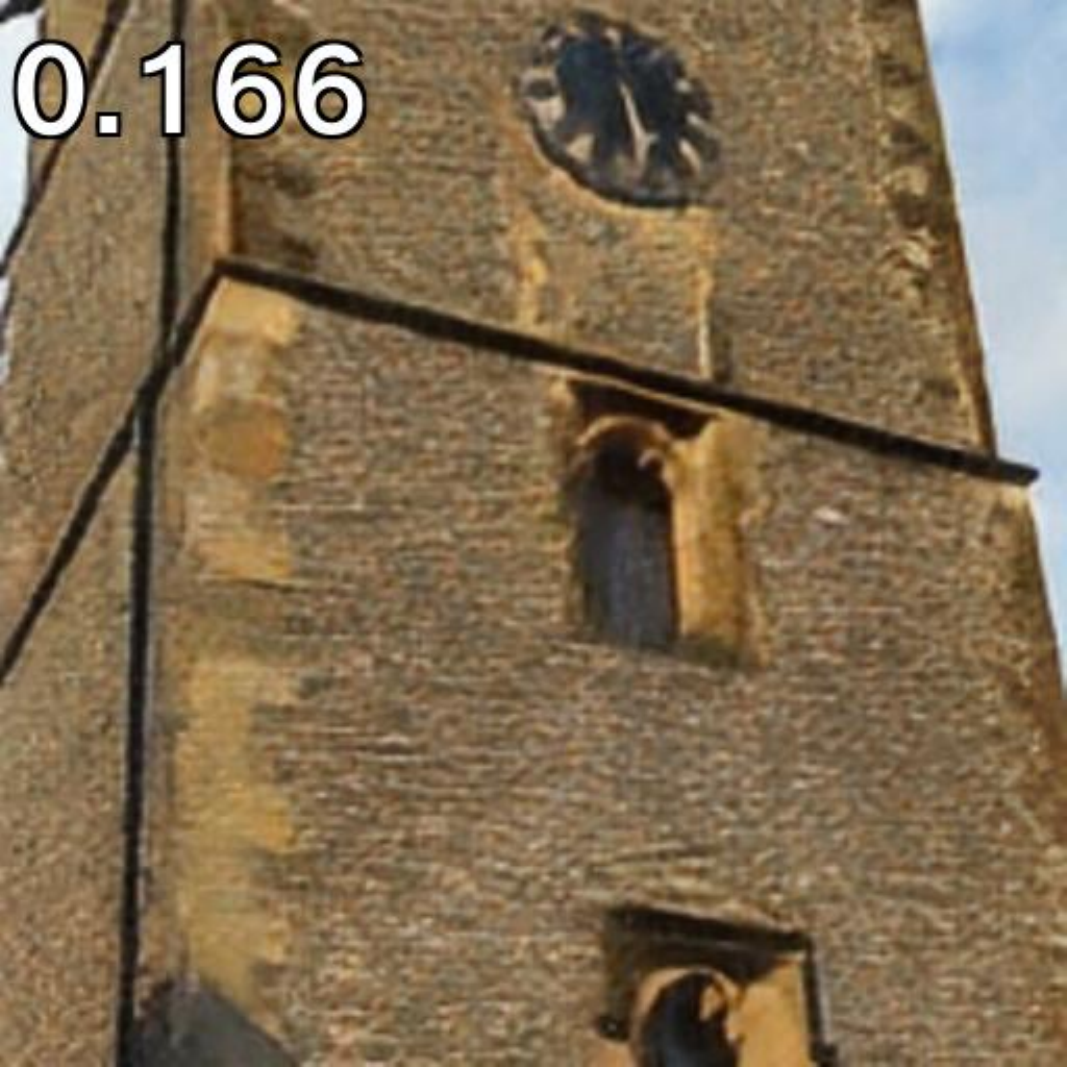}
     \includegraphics[width=1.0\linewidth]{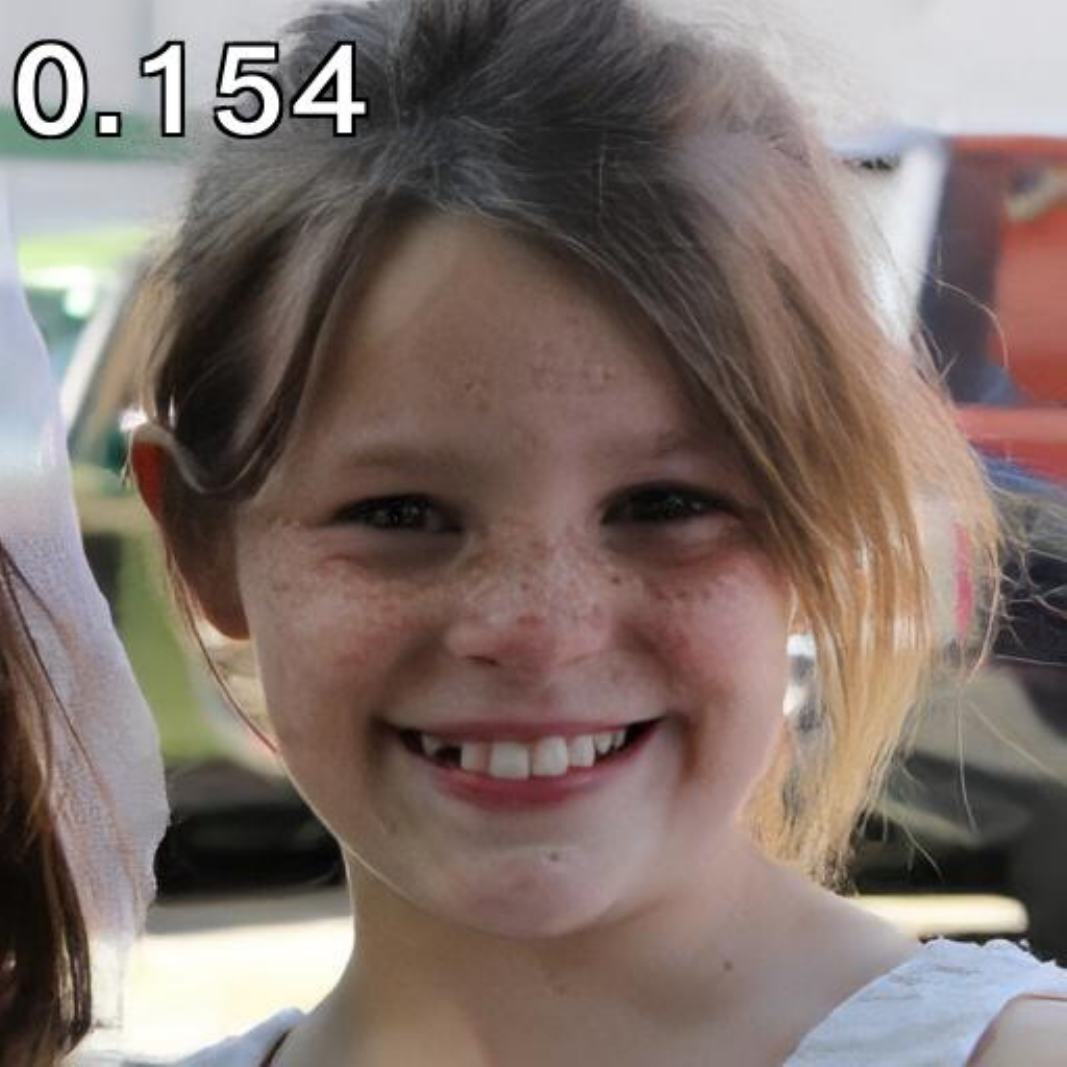}
           \includegraphics[width=1.0\linewidth]{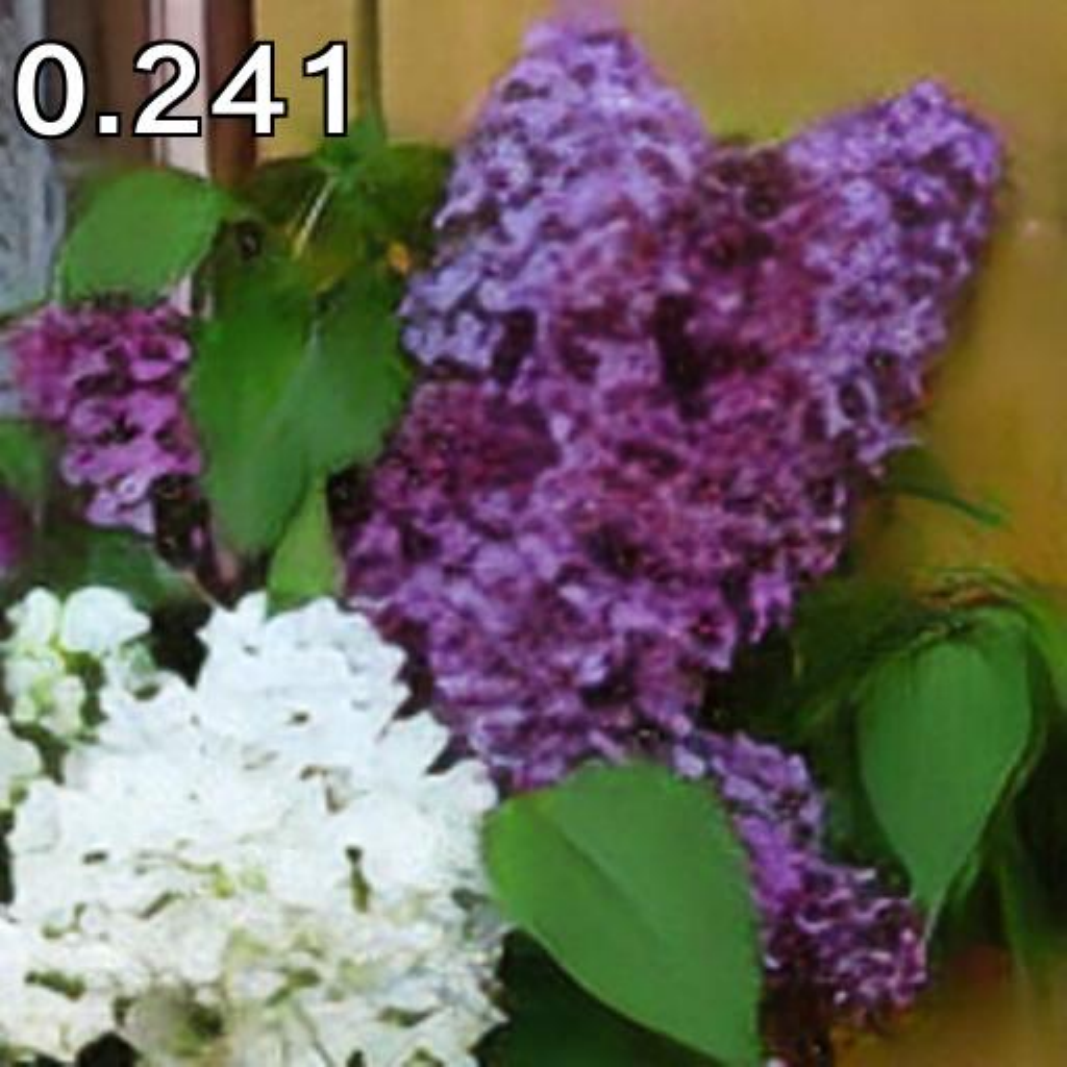}
  \caption{HifiC\cite{mentzer2020high}}
  \label{dataset_g}
\end{subfigure}
\begin{subfigure}{.09\linewidth}
  \centering
  \includegraphics[width=1.0\linewidth]{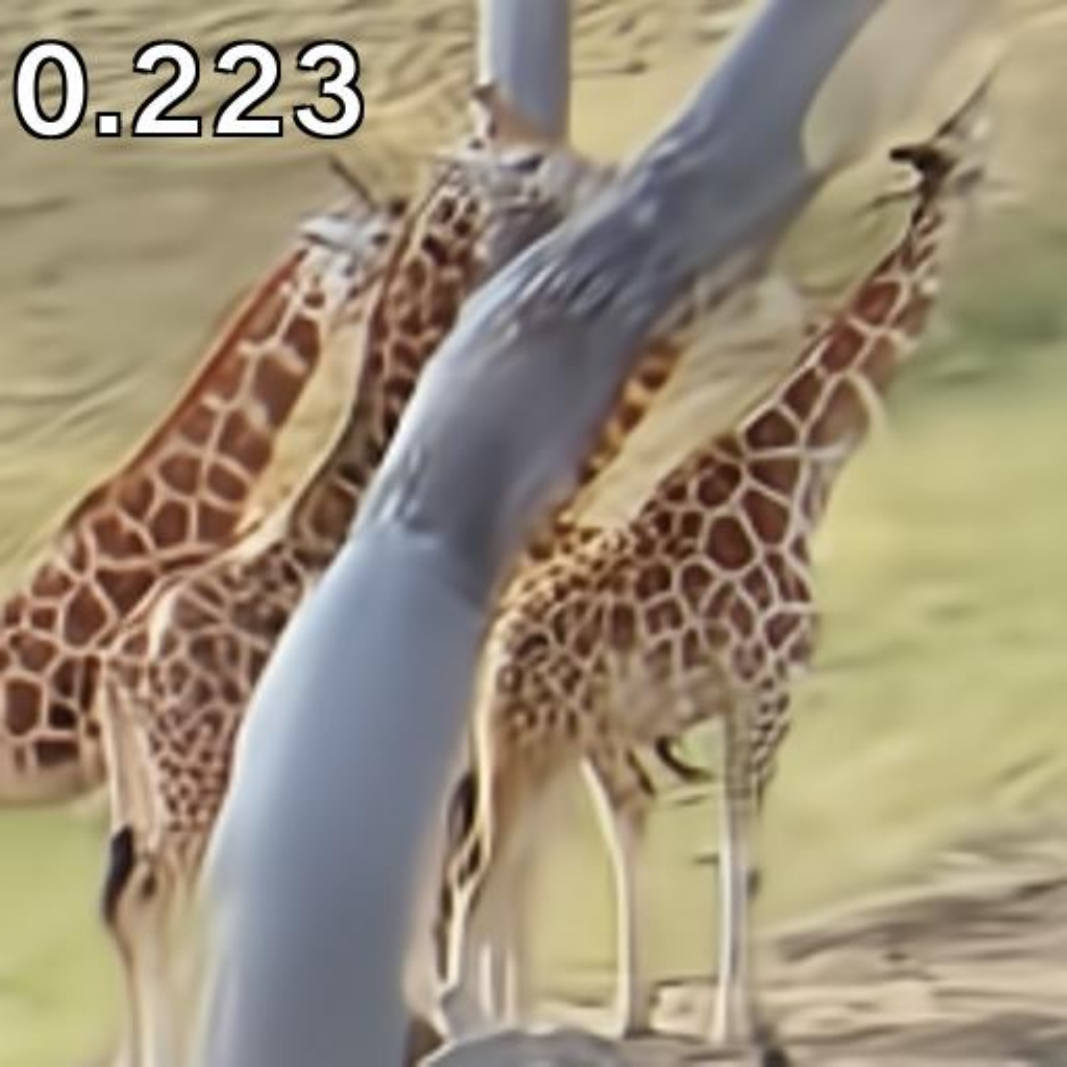}
  \includegraphics[width=1.0\linewidth]{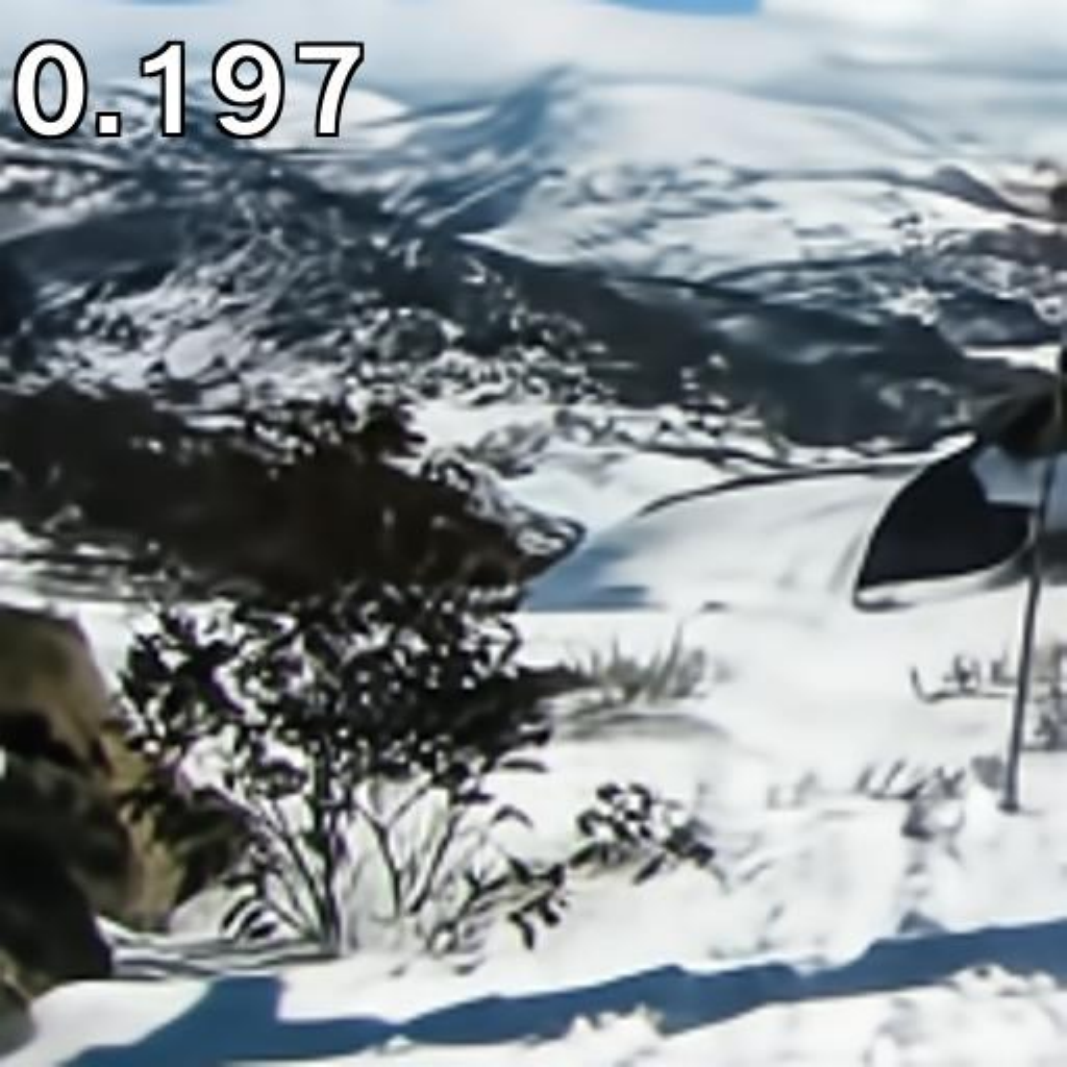}
    \includegraphics[width=1.0\linewidth]{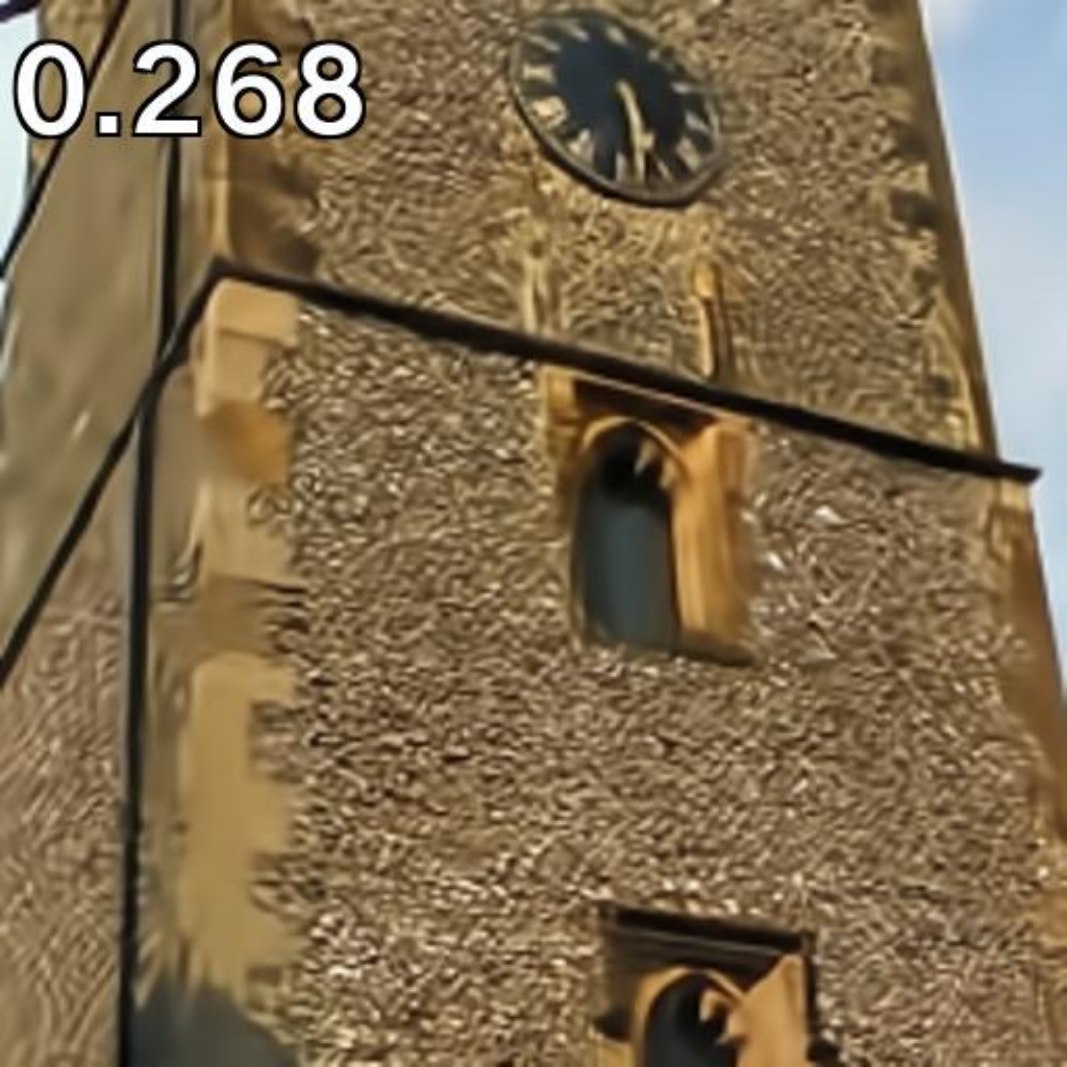}
     \includegraphics[width=1.0\linewidth]{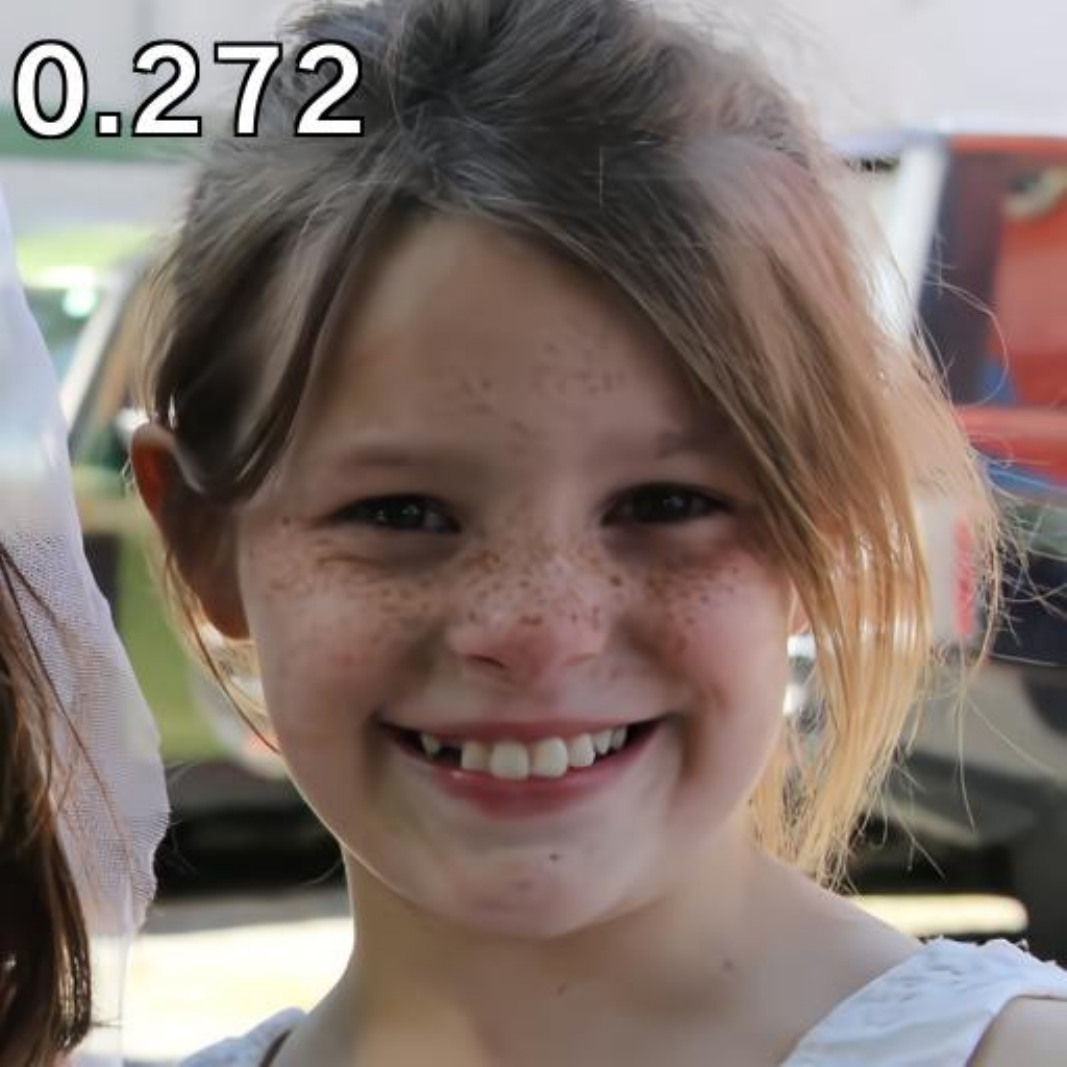}
           \includegraphics[width=1.0\linewidth]{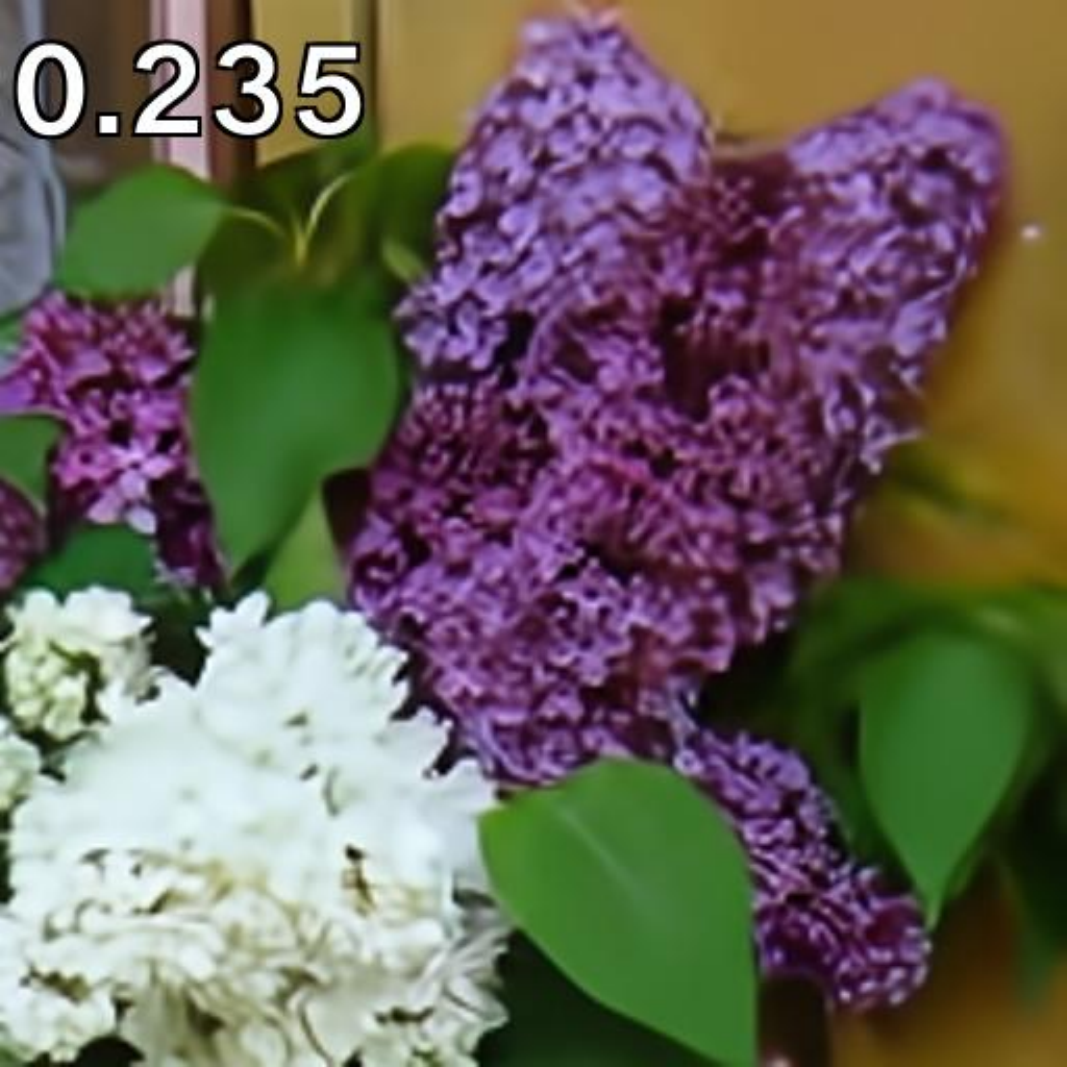}
  \caption{ELIC \cite{he2022elic}}
  \label{dataset_h}
\end{subfigure}
\begin{subfigure}{.09\linewidth}
  \centering
  \includegraphics[width=1.0\linewidth]{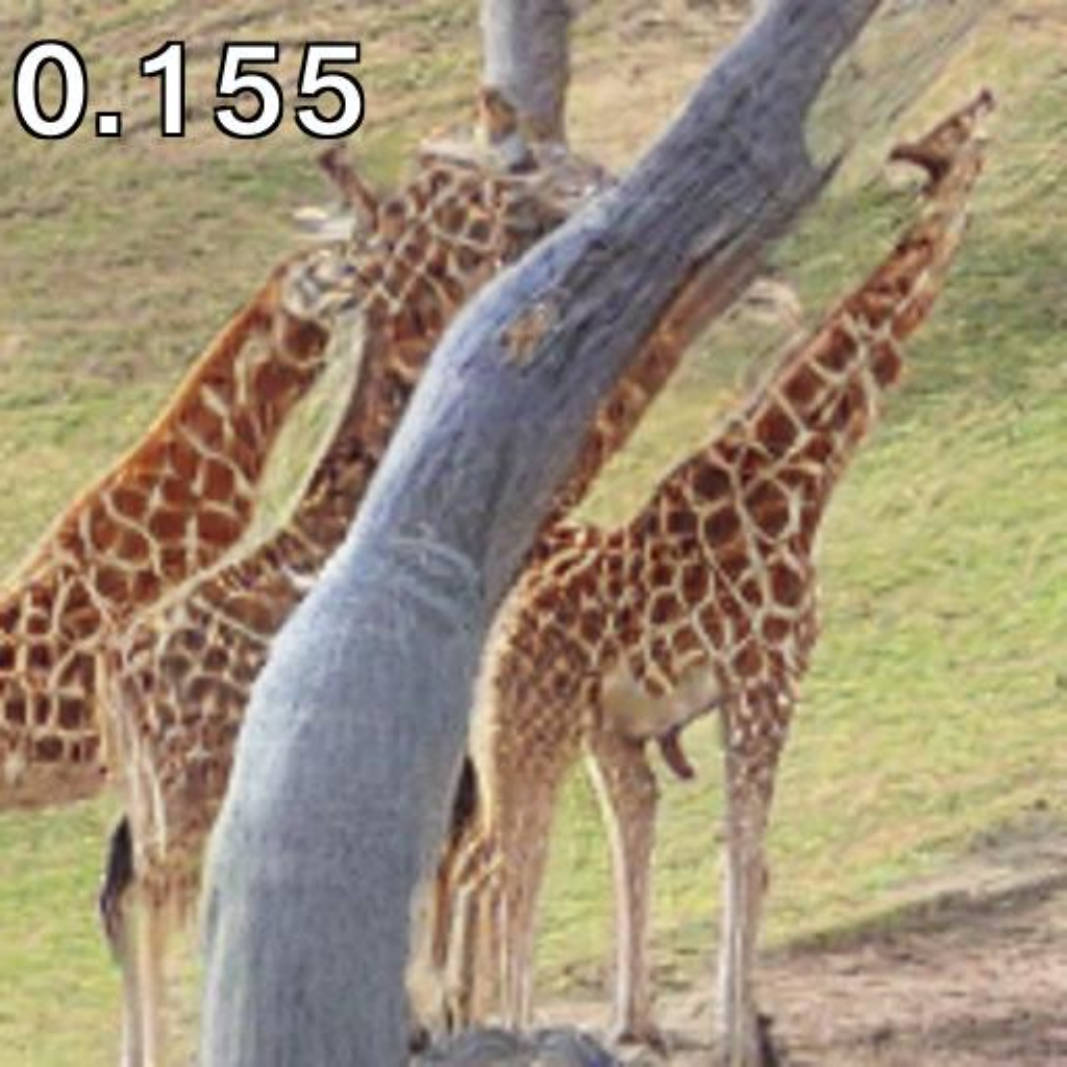}
  \includegraphics[width=1.0\linewidth]{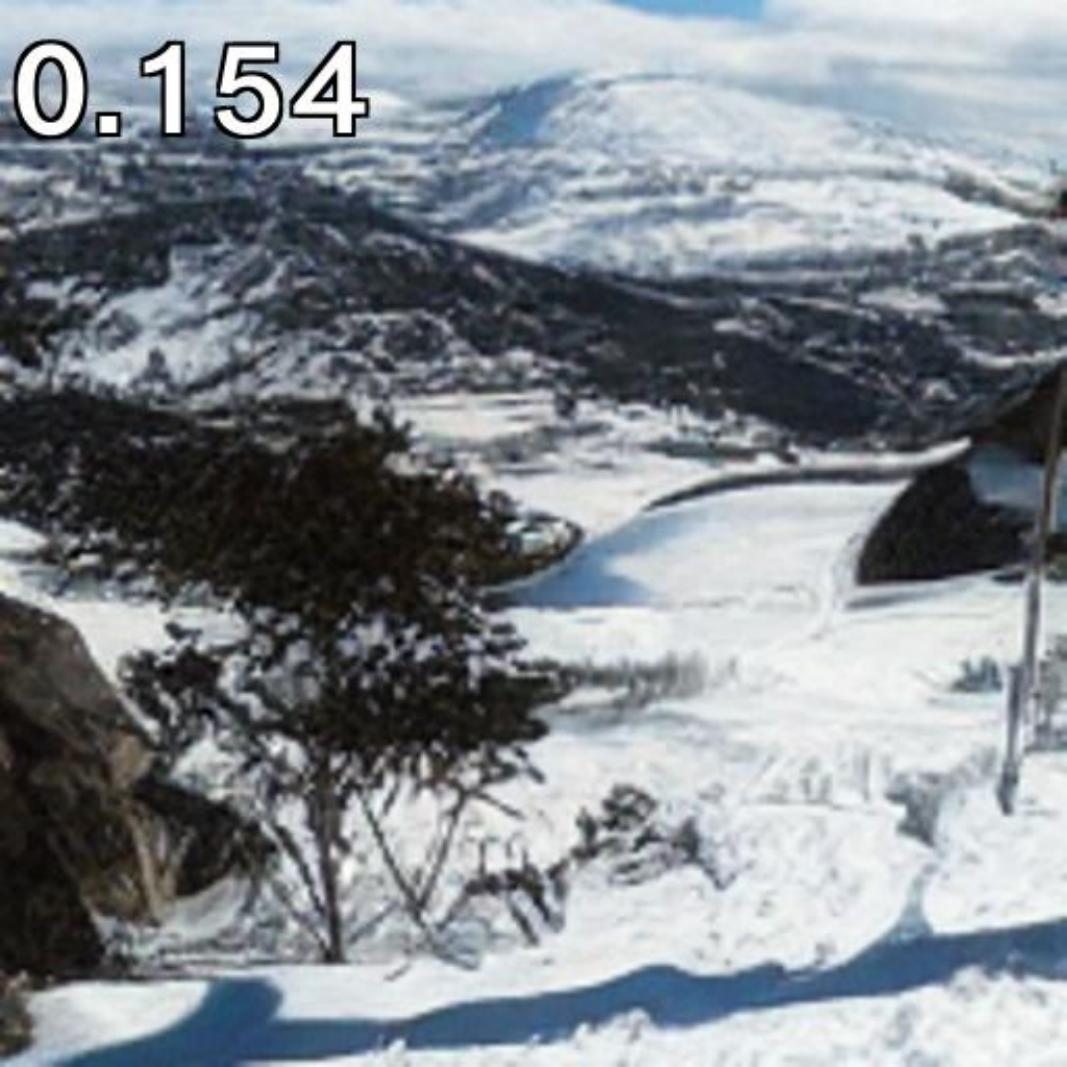}
    \includegraphics[width=1.0\linewidth]{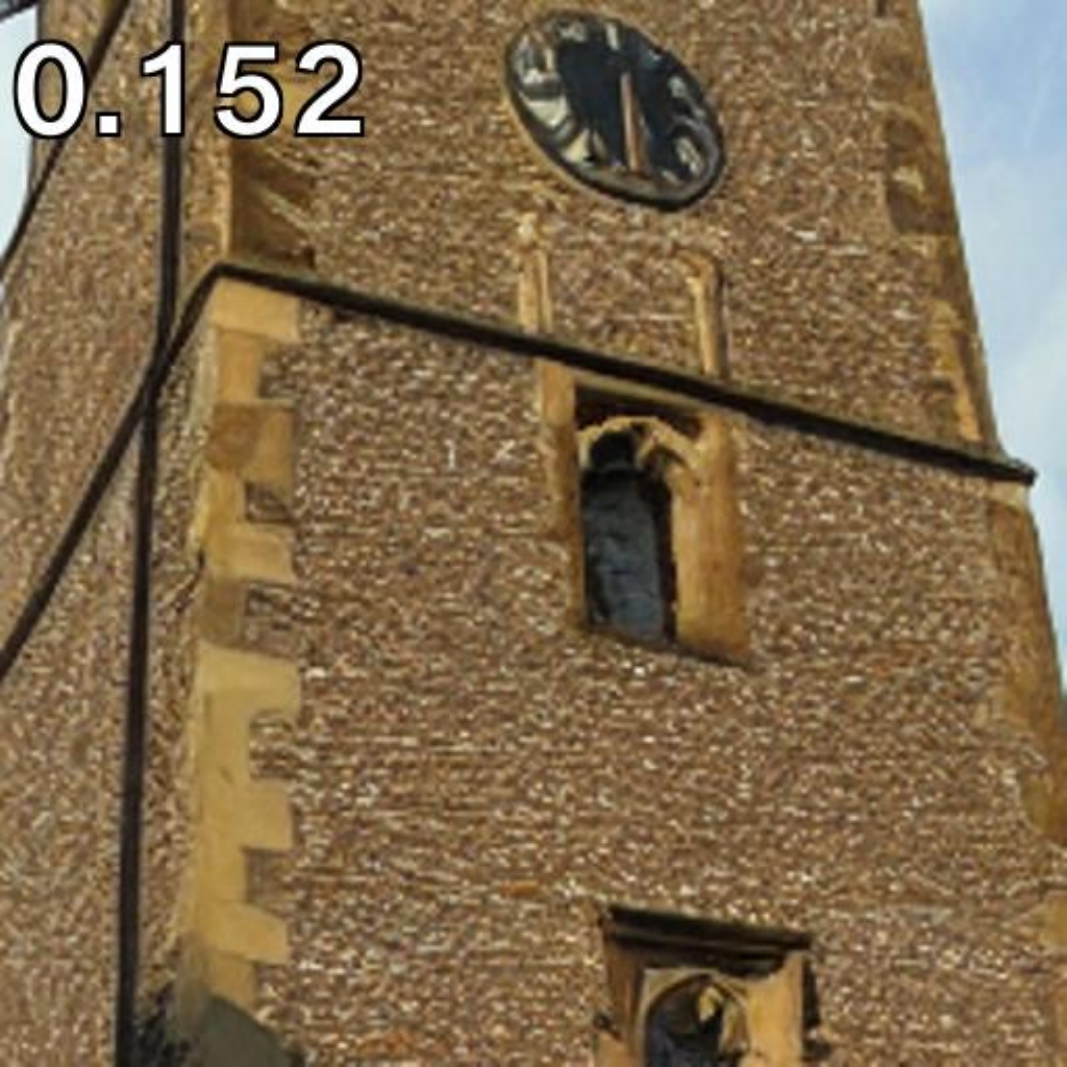}
    \includegraphics[width=1.0\linewidth]{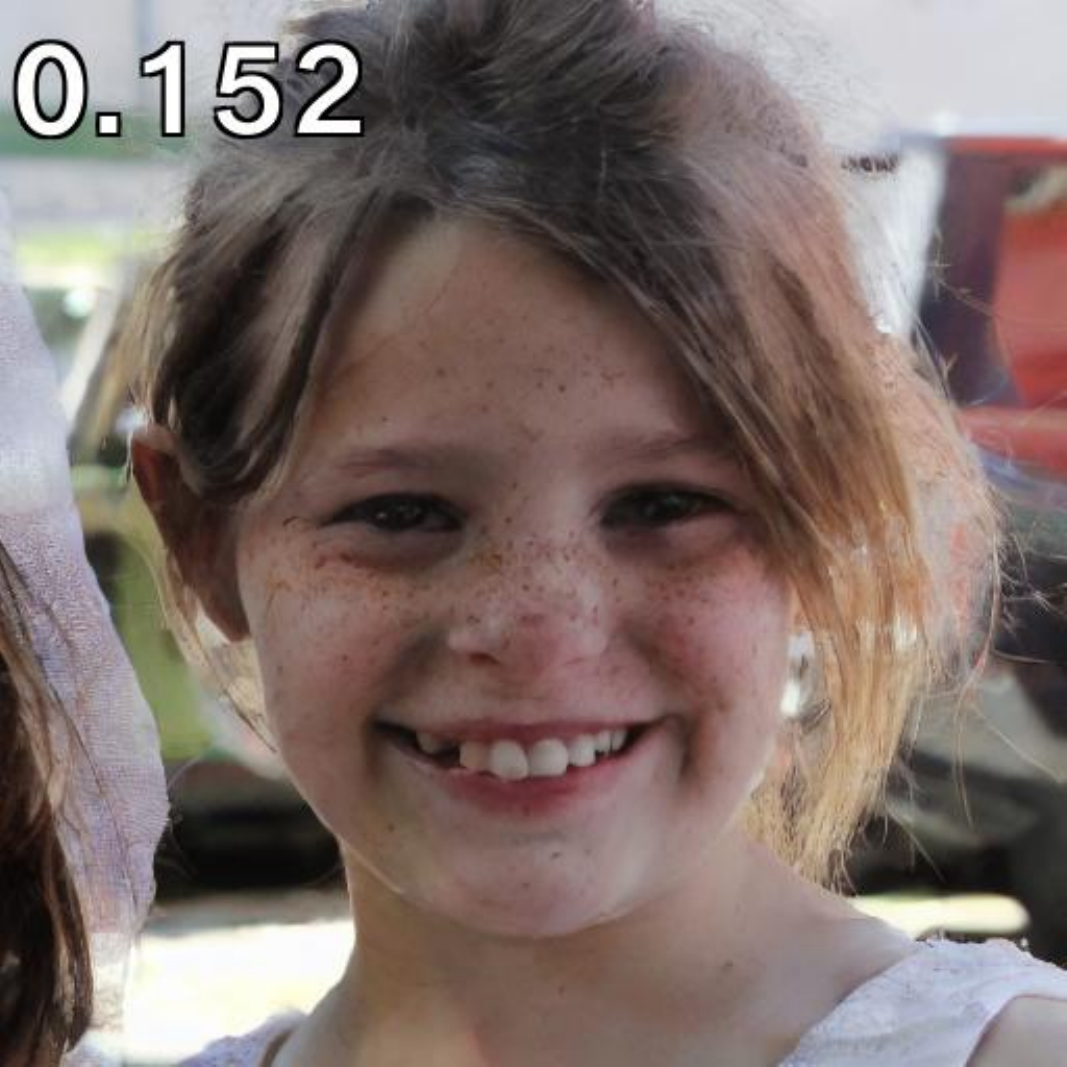}
         \includegraphics[width=1.0\linewidth]{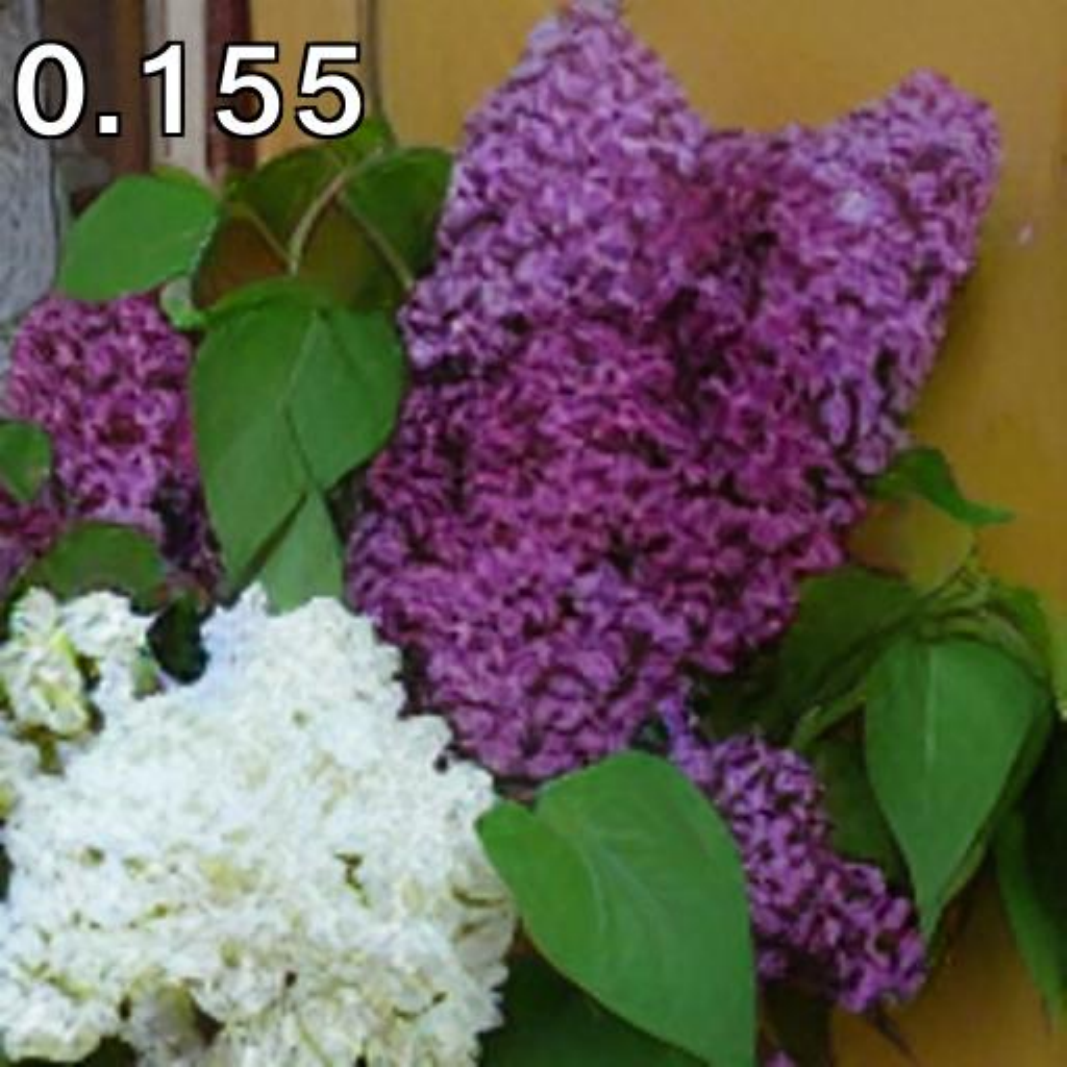}
  \caption{Ours}
  \label{dataset_i}
\end{subfigure}
\vspace{-0.3cm}
\caption{ The qualitative comparison of various image codecs at low bit-rate.} 
\label{figure6}
\vspace{-0.2cm}
\end{figure*}

\noindent \textbf{Qualitative Results.}
Figure \ref{figure6} demonstrates that our method produces superior visual results compared to previous state-of-the-art methods, even in extreme cases with a low bit rate. Most compression algorithms suffer from serious color blocks, ringing, blur or artifacts due to over-quantization and filtering, while our method effectively reconstructs fine textures, such as the tree trunks in the 1st row, the mountain in the 2nd row, the gravel in the 3rd row. In terms of medium-grained contour reconstruction, our method correctly captures the giraffe's head and legs in the 1st row, the clear shapes of the leaves in the 5th row, the sharp and distinct clock dial plate in the 3rd row, and the freckles on faces in the 4th row.
As for the high level semantic information preservation, we will discuss it in section \ref{section4-3}.

\noindent \textbf{Analysis of the Losses and the Hyper-parameter Patch Size.} We have also conducted experiments analyzing the losses of Eqn.\ref{eq12}, as shown in Table \ref{table:2}. Our pseudo-label generation and contrastive learning have significant benefits on the FID \cite{heusel2017gans} metric. This is because the FID \cite{heusel2017gans} metric uses the inception-v3 model to extract feature vectors of two datasets and calculates their distribution differences in the feature space. With the pretext task, our generated features become more discriminative, and the features become more compact, bringing their feature-space distribution closer to that of the real images.

\noindent \textbf{Analysis of the Hyper-parameter Patch Size.}
We also discussed the hyper-parameter patch size $N$ mentioned in section \ref{subsection4-1}. As shown in Table \ref{table:2}, the optimal patch size is 320. This may be due to the fact that excessively large blocks are heavily distorted during the palette diffusion process, while excessively small blocks cannot capture complete texture and semantic structure information.
\subsection{Evaluation for Machine Vision} \label{section4-3}
\begin{table*}[]
\clearpage
\small
\caption{Performance comparisons (\%) with different methods for machine vision tasks.}
\vspace{-0.3cm}
\centering
\begin{tabularx}{50em}{p{3em} |p{5em} |*{1}{>{\centering\arraybackslash}X}|*{1}{>{\centering\arraybackslash}X}|*{1}{>{\centering\arraybackslash}X}|*{1}{>{\centering\arraybackslash}X}|*{1}{>{\centering\arraybackslash}X}|*{1}{>{\centering\arraybackslash}X}||*{1}{>{\centering\arraybackslash}X}|*{1}{>{\centering\arraybackslash}X}|*{1}{>{\centering\arraybackslash}X}|*{1}{>{\centering\arraybackslash}X}} 
 \hline
  \multicolumn{2}{c|}{Dataset  }                                   & \multicolumn{6}{c||}{COCO 2017 \cite{lin2014microsoft}} & \multicolumn{4}{c}{WIDER FACE \cite{yang2016wider}}\\ [0.5ex] 
 \hline
 \multicolumn{2}{c|}{Task  }                                     & \multicolumn{3}{c|}{Detection}& \multicolumn{3}{c||}{Segmentation} & \multicolumn{4}{c}{Facial Landmark Detection}\\ [0.5ex] 
  \hline
   \multicolumn{2}{c|}{Metric  } & bpp&mAP & AR & bpp& mAP & AR & bpp & \multicolumn{3}{c}{mAP}\\ [0.5ex] 
\hline
  \multicolumn{2}{c|}{Setting  }                                   &  \multicolumn{6}{c||}{ IoU=0.50:0.95 | area=   all | maxDets=100 }  & &Easy & \centering Medium & Hard\\ [0.5ex] 
\hline
\hline
\multirow{8}{*}{method} 
&\multirow{1}{*}{JPEG}  \cite{wallace1992jpeg} & 0.153  & 18.1     & 27.8 & 0.153  & 15.8& 24.0     & 0.174&86.69   & 79.19 & 52.35\\  
 &\multirow{1}{*}{WebP} \cite{mukherjee2014webp} &  0.154  & 30.6       &41.8  & 0.154&27.0&  37.0     & 0.201&93.28   & 89.42 & 64.15\\ 
&\multirow{1}{*}{x265} \cite{ramachandran2013x265} &0.168    &23.1       &33.9  &0.168&  20.2& 29.4     &0.210& 77.46   & 74.97 & 53.51\\ 
&\multirow{1}{*}{IRN} \cite{xiao2023invertible} &0.154    & 26.0      & 36.0  & 0.154 & 23.0& 31.8     &0.180& 90.68   & 84.94 & 57.55 \\ 
&\multirow{1}{*}{VTM} \cite{bross2021overview} &0.161  &  36.0     & 47.6 & 0.161     &32.2& 42.7     &0.239& 93.71   & 90.46&71.24\\ 
&\multirow{1}{*}{HifiC} \cite{mentzer2020high} &0.156  &  35.2      &  46.3  &0.156& 31.5 &  41.5   &0.177 &94.46   & 91.74 & 72.14\\
&\multirow{1}{*}{Ours} &0.153  & \textbf{37.6}      & \textbf{48.6}  &0.153&\textbf{33.5} & \textbf{43.4}     & 0.154&\textbf{94.84}   & \textbf{92.10} &\textbf{75.10}\\
&\multirow{1}{*}{Original} &5.514  & 47.2      & 59.09  &5.514&42.05& 54.5    & 6.237 &95.48   & 94.04 &84.43\\
\hline
\end{tabularx}
\label{table:4}
\end{table*}
Furthermore, the machine vision performance of our method at extremely low-bitrate (0.15 bpp) is  presented in Table \ref{table:4}. We perform instance detection and segmentation \cite{wang2022internimage} on the original COCO 2017 validation \cite{lin2014microsoft} and the constructed dataset by JPEG \cite{wallace1992jpeg}, WebP \cite{mukherjee2014webp}, x265 \cite{ramachandran2013x265}, IRN \cite{xiao2023invertible}, VTM \cite{bross2021overview}, HifiC \cite{mentzer2020high} and our method. Facial landmark detection \cite{deng2020retinaface} is also performed on the original and constructed WIDER FACE validation \cite{yang2016wider}. Machine vision results on the original data are used as ground truth. It can be seen that our method achieves higher mean average precision at the similar bit-rate compared to other methods. In particular, on Easy and Medium face landmark detection task, our proposed method maintains semantic quality with only  0.64\% - 1.44\% degradation even with  39.9$\times$ further data compression based on the original dataset images, showing its robustness. We also found that different vision tasks require different compression ratios to preserve semantic information and emphasize the importance of scalable compression methods. Hard tasks such as instance detection and segmentation are more sensitive to bitrates than easy task facial landmark detection. We present the experimental results corresponding to more bit-rate points into the Appendix. 
\vspace{-0.3cm}
\subsection{Analysis of the Scalable Mechanism}  \label{section4-4}
Our scalable encoding provides a resource-friendly and adaptable solution, especially useful in scenarios with constraints on computational resources or varying task demands. In Fig. \ref{fig5}, our approach enables a smooth trade-off between bitrate and perceptual quality during testing. Table \ref{table:scalable} shows that more challenging tasks in Facial Landmark Detection require higher bitrates to meet the mAP criteria. Scalable encoding proves advantageous in such cases. Furthermore, Table \ref{table:scalable} reveals that different vision tasks demand varying bitrates to satisfy the same mAP criteria. This highlights the benefits of scalable encoding. Our decision to use scalable encoding allows dynamic adjustment of the bitrate, optimizing for various visual tasks to meet specific mAP requirements without unnecessarily using a higher bitrate. 
\begin{table}[h!]
\small
\caption{The relationship between bitrate and mAP loss.}
\vspace{-0.3cm}
\centering
\begin{tabularx}{30em}{*{1}{>{\centering\arraybackslash}X} | *{1}{>{\centering\arraybackslash}X}        |*{1}{>{\centering\arraybackslash}X}|*{1}{>{\centering\arraybackslash}X} |*{1}{>{\centering\arraybackslash}X}|*{1}{>{\centering\arraybackslash}X}| *{1}{>{\centering\arraybackslash}X} | *{1}{>{\centering\arraybackslash}X} | *{1}{>{\centering\arraybackslash}X}  } 
 \hline
\multicolumn{3}{c|}{Facial Landmark Easy  }    & \multicolumn{3}{c|}{Facial Landmark Med  } & \multicolumn{3}{c}{Facial Landmark Hard  } \\[0.5ex] 
\hline
bpp & mAP & Loss & bpp & mAP & Loss & bpp & mAP & Loss \\[0.5ex] 
 \hline\hline
 \textbf{0.154 }  & \centering 98.84   & 0.6\%  & \textbf{0.154 } & 92.10& 2\% & \textbf{0.154 } & 75.10 & 12\%   \\ 
 \hline \hline
 \multicolumn{3}{c|}{Facial Landmark Hard }    & \multicolumn{3}{c|}{Detection } & \multicolumn{3}{c}{Segmentation  } \\[0.5ex] 
\hline
bpp & mAP & Loss & bpp & mAP & Loss & bpp & mAP & Loss \\[0.5ex] 
 \hline\hline
 0.154   & 75.10   & \textbf{12\%}  & 0.3 & 45.54& \textbf{12\%} & 0.35 & 37.00 & \textbf{12\%}   \\ 
 \hline
\end{tabularx}
\label{table:scalable}
\vspace{-0.3cm}
\end{table}
\section{Conclusion}
We proposed a novel content-adaptive and scalable image feature compression method to satisfy both human and machine perception. Specifically, a collaborative texture-semantic feature extraction and pseudo-label generation technique is applied in self-supervised manner for discriminative feature learning, following with a content-adaptive Markov palette diffusion model to enable users to select the desired compression ratio, resulting in scalable feature compression. Finally, our experimental results on image reconstruction and machine tasks demonstrated the superiority of our approach.
\begin{acks}
This work received support from the National Natural Science Foundation of China (62088102, 61972129), the PKU-NTU Joint Research Institute (JRI) sponsored by the Ng Teng Fong Charitable Foundation, and the Basic and Frontier Research Project of PCL, Major Key Project of PCL.
\end{acks}

\bibliographystyle{ACM-Reference-Format}
\balance
\bibliography{sample-base}
\newpage
\appendix

\section{Supplementary Experiments}
Here, we present more detailed experimental results of machine vision tasks on COCO 2017 \cite{lin2014microsoft} and WIDER FACE \cite{yang2016wider} at various bitrates, as described in Section \ref{section4-3} of the main text.

Due to the varying degrees of sensitivity of different machine vision tasks' accuracy to bitrates, our goal is to evaluate the performance of various compression algorithms at different bitrates for these tasks. To achieve a fair comparison, we aligned the bitrates of different algorithms. Firstly, we established the high, medium, and low bitrate points of the non-scalable HifiC \cite{mentzer2018conditional}. Then, we traversed different qp values of other algorithms to make the experimental groups comparable at approximately the same bitrate for the accuracy comparison of machine vision tasks.

The experimental results indicate that different machine vision tasks require different bitrates to maintain their accuracy. This highlights the significance of scalable encoding and serves as motivation for further research in this direction.
  
\subsection{Supplementary Experiments on COCO 2017 \cite{lin2014microsoft} Detection}
\begin{figure} [h] 
\centering
\begin{subfigure}{0.48\linewidth}
\centering
\includegraphics[width=1.0\linewidth]{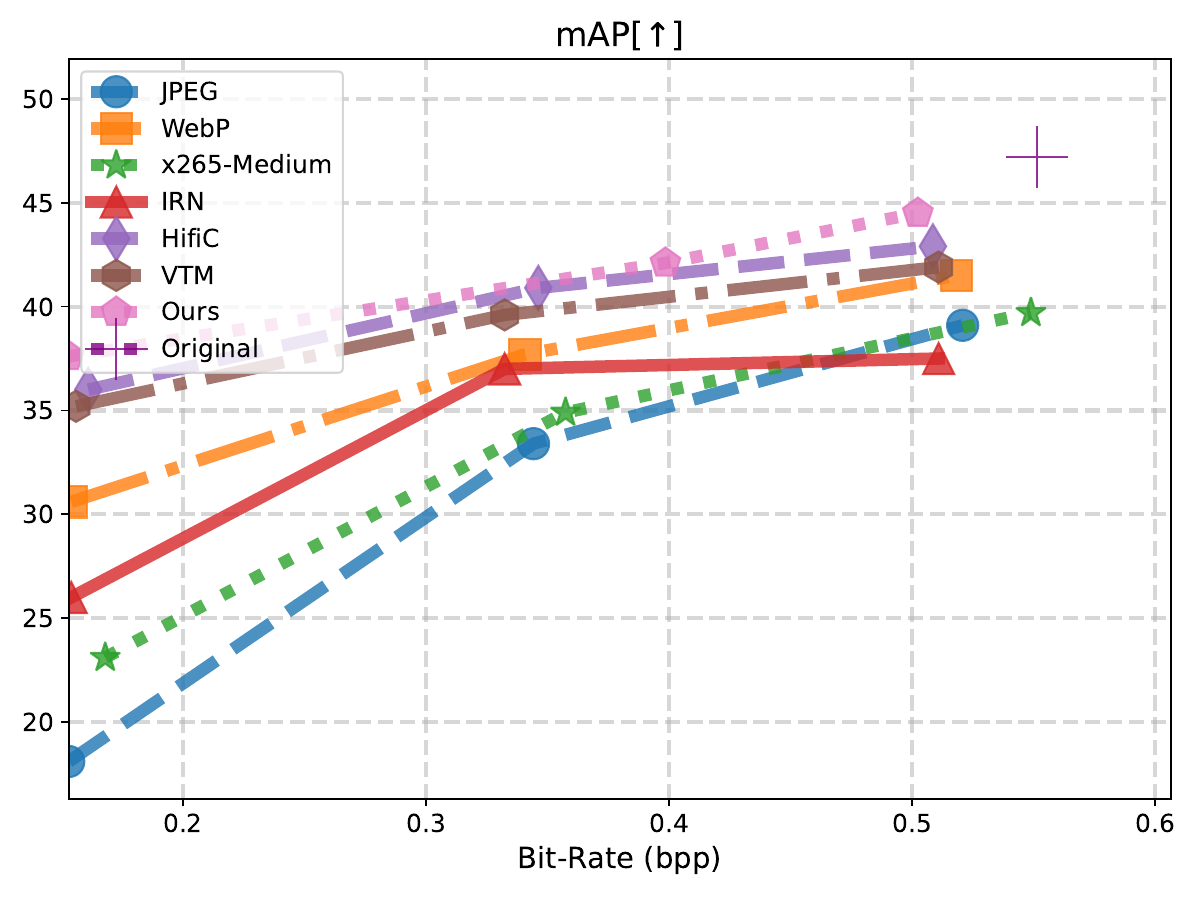}
\caption{}
\label{fig7_a}
\end{subfigure}
\begin{subfigure}{0.48\linewidth}
\centering
\includegraphics[width=1.0\linewidth]{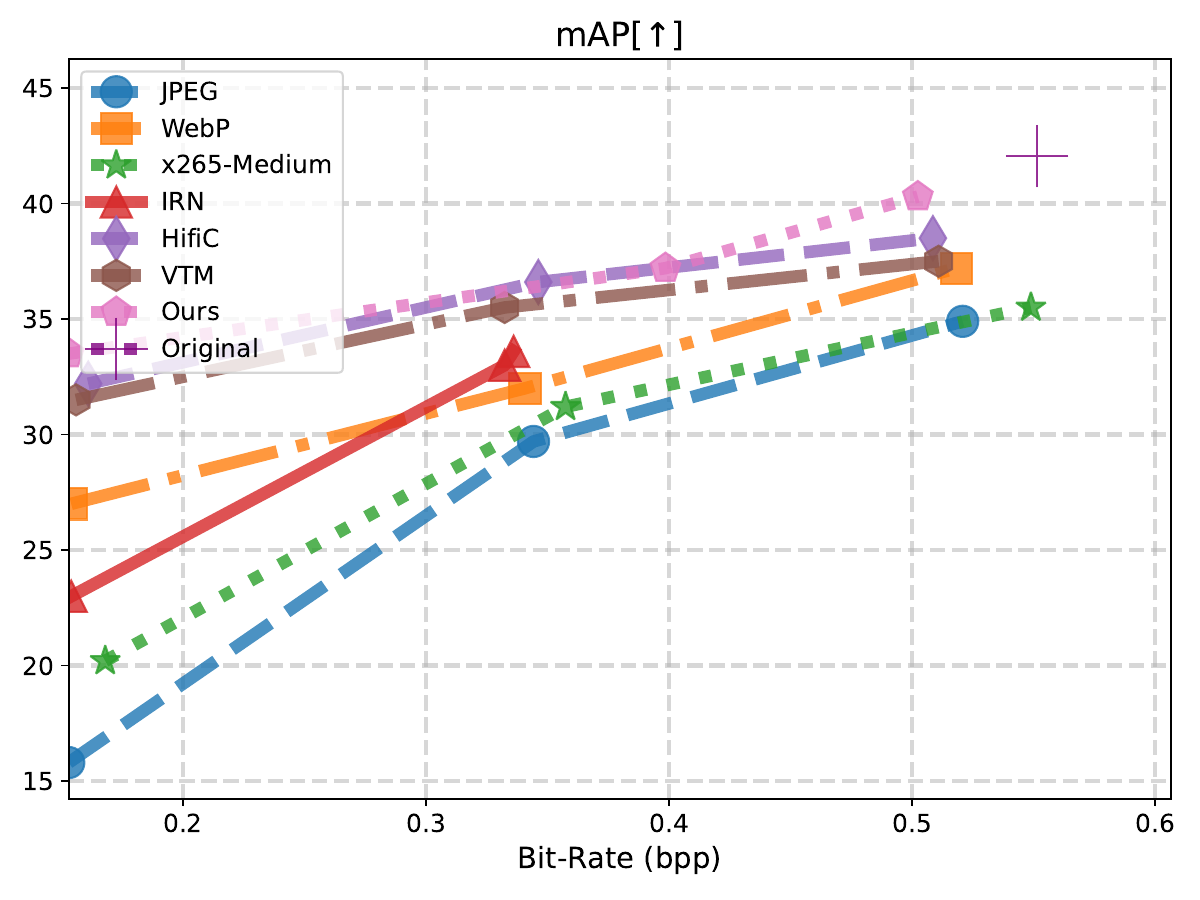}
\caption{}
\label{fig7_b}
\end{subfigure}
  \caption{ Bitrate and performance of CV tasks are shown in (a) as the curve of Bit-Rate and instance detection mAP on COCO 2017 \cite{lin2014microsoft}, and in (b) as the curve of Bit-Rate and instance segmentation mAP on COCO 2017 \cite{lin2014microsoft}.}
  \Description{Enjoying the baseball game from the third-base  seats. Ichiro Suzuki preparing to bat.}
  \label{fig7}
\end{figure}

Table \ref{table:4} presents the performance of various methods on COCO 2017 \cite{lin2014microsoft}, and Fig. \ref{fig7} (a) visually represents the Bit-Rate and mAP curve. Based on the results, our method achieves accuracy closest to the ground truth at the same bitrate. To include the ground truth chart in Fig. \ref{fig7} (a), we compressed its horizontal axis by a factor of 10. Notably, our algorithm only requires a 0.1x bitrate to achieve an mAP error of less than 5\%.
\begin{table*}[htbp]
\clearpage
\small
\caption{Performance comparisons (\%) with different methods for instance detection on COCO 2017 \cite{lin2014microsoft}.}
\centering
\begin{tabularx}{\linewidth}{p{3em} |p{5em} |*{1}{>{\centering\arraybackslash}X}|*{1}{>{\centering\arraybackslash}X}|*{1}{>{\centering\arraybackslash}X}|*{1}{>{\centering\arraybackslash}X}|*{1}{>{\centering\arraybackslash}X}||*{1}{>{\centering\arraybackslash}X}|*{1}{>{\centering\arraybackslash}X}|*{1}{>{\centering\arraybackslash}X}|*{1}{>{\centering\arraybackslash}X}|*{1}{>{\centering\arraybackslash}X}||*{1}{>{\centering\arraybackslash}X}|*{1}{>{\centering\arraybackslash}X}|*{1}{>{\centering\arraybackslash}X}|*{1}{>{\centering\arraybackslash}X}|*{1}{>{\centering\arraybackslash}X}} 
 \hline
   \multicolumn{2}{c|}{Metric  } & bpp&PSNR & SSIM  & mAP & AR & bpp& PSNR & SSIM  & mAP & AR & bpp& PSNR & SSIM  &  mAP & AR \\ [0.5ex] 
\hline
  \multicolumn{2}{c|}{Setting  }                                   &  \multicolumn{15}{c}{ IoU=0.50:0.95 | area=   all | maxDets=100 }   \\ [0.5ex] 
\hline
\hline
\multirow{8}{*}{Method} &\multirow{1}{*}{JPEG}  \cite{wallace1992jpeg} & 0.153  & 24.0     & 0.641  & 18.1  & 27.8     & 0.344 & 26.4   & 0.752 & 33.4  & 45.5     & 0.520 & 28.0   & 0.806 & 39.1 & 51.2 \\  
 &\multirow{1}{*}{WebP} \cite{mukherjee2014webp} &  0.154  & 26.3       & 0.709  & 30.6 & 41.8  & 0.340  &  28.2    & 0.796& 37.7   & 50.0 & 0.518 & 29.9  & 0.847    & 41.5 & 53.7   \\ 
&\multirow{1}{*}{x265} \cite{ramachandran2013x265} &0.168    &24.7       & 0.651 & 23.1  &  33.9  & 0.357     &26.6& 0.722  & 34.9 & 46.8 & 0.548 & 28.1    & 0.774 & 39.7  & 51.9 \\ 
&\multirow{1}{*}{IRN} \cite{xiao2023invertible} &0.154    & 25.4      & 0.688  & 26.0 & 36.0 & 0.332     &27.4&0.783   & 37.0 & 48.6 & 0.510  & 27.8    & 0.796 & 37.5  & 49.4 \\ 
&\multirow{1}{*}{VTM} \cite{bross2021overview} &0.161  &  28.7    & 0.790 & 36.0      &47.6& 0.346     &30.3& 0.850   & 40.9&53.1 & 0.508 & 31.7     & 0.886 & 42.9  & 55.1 \\ 
&\multirow{1}{*}{HifiC} \cite{mentzer2018conditional} &0.156  &  26.6      &  0.761   &35.2& 46.3 &  0.332   &27.86 &0.837   & 39.6 & 51.1 & 0.510  & 30.723    & 0.856 & 41.9  & 53.6 \\
&\multirow{1}{*}{Ours} &0.153  & 23.4     & 0.716   &\textbf{37.6}&\textbf{48.6} & 0.398     & 24.4& 0.697   & \textbf{42.1} & \textbf{54.9 }& 0.502 & 27.5     & 0.850 & \textbf{44.5}  & \textbf{57.6}\\
&\multirow{1}{*}{Original} &5.514  & -      & - &47.2&59.0& -    & - &-   & - &-& -  & -     & - & - & -\\
\hline
\end{tabularx}
\label{table:5}
\end{table*}
\subsection{Supplementary Experiments on COCO 2017 \cite{lin2014microsoft} Segmentation}
Table \ref{table2} presents the performance of various methods on COCO 2017 \cite{lin2014microsoft}, and Fig. \ref{fig7} (b) visually represents the Bit-Rate and mAP curve. To include the ground truth chart in Fig. \ref{fig7} (b), we compressed its horizontal axis by a factor of 10. Based on the results, our method achieves accuracy closest to the ground truth at the same bitrate. Notably, our method only requires a 0.1x bitrate to achieve an mAP error of less than 2\%.
\begin{table*}[htbp]
\clearpage
\small
\caption{Performance comparisons (\%) with different methods for instance segmentation on COCO 2017 \cite{lin2014microsoft}.}
\centering
\begin{tabularx}{\linewidth}{p{3em} |p{5em} |*{1}{>{\centering\arraybackslash}X}|*{1}{>{\centering\arraybackslash}X}|*{1}{>{\centering\arraybackslash}X}|*{1}{>{\centering\arraybackslash}X}|*{1}{>{\centering\arraybackslash}X}||*{1}{>{\centering\arraybackslash}X}|*{1}{>{\centering\arraybackslash}X}|*{1}{>{\centering\arraybackslash}X}|*{1}{>{\centering\arraybackslash}X}|*{1}{>{\centering\arraybackslash}X}||*{1}{>{\centering\arraybackslash}X}|*{1}{>{\centering\arraybackslash}X}|*{1}{>{\centering\arraybackslash}X}|*{1}{>{\centering\arraybackslash}X}|*{1}{>{\centering\arraybackslash}X}} 
 \hline
   \multicolumn{2}{c|}{Metric  } & bpp&PSNR & SSIM  & mAP & AR & bpp& PSNR & SSIM  & mAP & AR & bpp& PSNR & SSIM  &  mAP & AR \\ [0.5ex] 
\hline
  \multicolumn{2}{c|}{Setting  }                                   &  \multicolumn{15}{c}{ IoU=0.50:0.95 | area=   all | maxDets=100 }   \\ [0.5ex] 
\hline
\hline
\multirow{8}{*}{Method} &\multirow{1}{*}{JPEG}  \cite{wallace1992jpeg} & 0.153  & 24.0     & 0.641 & 15.8   & 24.0   & 0.344     & 26.4 & 0.752   & 29.7& 40.4 & 0.520  & 28.0     & 0.806 & 34.9  & 46.0 \\  
 &\multirow{1}{*}{WebP} \cite{mukherjee2014webp} &  0.154  & 26.3       &0.709    & 27.0 &37.0  &  0.340     &28.2 & 0.796   & 32.0 & 37.0 & 0.518  & 29.9     & 0.847& 37.2  & 48.3 \\ 
&\multirow{1}{*}{x265} \cite{ramachandran2013x265} &0.168    &24.7       &  0.168&  20.2& 29.4     &0.357& 26.6   & 0.722 & 31.2 & 41.9  & 0.548     & 28.1 & 0.774 & 35.5 & 46.7 \\ 
&\multirow{1}{*}{IRN} \cite{xiao2023invertible} &0.154    & 25.4      & 0.688   & 23.0  & 31.8& 0.332     &27.45& 0.783   & 33.0 & 43.8 & 0.510  & 27.8     & 0.796 & 33.6  & 44.5 \\ 
&\multirow{1}{*}{VTM} \cite{bross2021overview} &0.161  &  28.7     & 0.790 & 32.2      &42.7 & 0.346    &30.3& 0.850   & 36.6&48.0 & 0.508  & 31.75     & 0.886 & 38.5  & 49.8 \\ 
&\multirow{1}{*}{HifiC} \cite{mentzer2018conditional} &0.156  &  26.6      &  0.761   &31.5 & 41.5  & 0.332  &27.8 &0.837   & 35.5 & 45.9 & 0.510  & 30.7     & 0.856 & 37.5  & 48.4 \\
&\multirow{1}{*}{Ours} &0.153  & 23.4    & 0.716  &  33.5 & 43.4 &0.398    & 24.46 & 0.697   & 37.2 & 49.3 & 0.502  & 27.5     & 0.85 & \textbf{40.3}  & \textbf{51.2} \\
&\multirow{1}{*}{Original} &5.514  & -      & -  & 42.0  &54.5 & -   & - &-  &- & - & -  & -    & - & -  & - \\
\hline
\end{tabularx}
\label{table2}
\end{table*}
\subsection{Supplementary Experiments on WIDER FACE \cite{yang2016wider} Facial Landmark Detection}
Table \ref{table3} presents the performance of various methods on WIDER FACE \cite{yang2016wider}. Based on the results, our method achieves accuracy closest to the ground truth at the same bitrate. Notably, our method only requires a 0.063x bitrate to achieve an mAP error of less than 1.7\% on the hard facial landmark detection task.
\begin{table*}[htbp]
\clearpage
\small
\caption{Performance comparisons (\%) with different methods for facial landmark  detection on WIDER FACE \cite{yang2016wider}.}
\centering
\begin{tabularx}{\linewidth}{p{4.5em} |*{1}{>{\centering\arraybackslash}X}|*{1}{>{\centering\arraybackslash}X}|*{1}{>{\centering\arraybackslash}X}|*{1}{>{\centering\arraybackslash}X}|p{3em}|p{2em}||*{1}{>{\centering\arraybackslash}X}|*{1}{>{\centering\arraybackslash}X}|*{1}{>{\centering\arraybackslash}X}|*{1}{>{\centering\arraybackslash}X}|p{3em}|p{2em}||*{1}{>{\centering\arraybackslash}X}|*{1}{>{\centering\arraybackslash}X}|*{1}{>{\centering\arraybackslash}X}|*{1}{>{\centering\arraybackslash}X}|p{3em}|p{2em}} 
 \hline
\multirow{2}{*}{Method} & \multirow{2}{*}{bpp} & \multirow{2}{*}{\centering PSNR}& \multirow{2}{*}{\centering SSIM}& \multicolumn{3}{c||}{mAP}& \multirow{2}{*}{bpp} & \multirow{2}{*}{\centering PSNR}& \multirow{2}{*}{\centering SSIM}& \multicolumn{3}{c||}{mAP}& \multirow{2}{*}{bpp} & \multirow{2}{*}{\centering PSNR}& \multirow{2}{*}{\centering SSIM}& \multicolumn{3}{c}{mAP} \\ \cline{5-7}  \cline{11-13}  \cline{17-19} 

 &&&& Easy & Medium & Hard &&&& Easy & Medium & Hard & &&&Easy & Medium & Hard \\    
\hline
\hline
\multirow{1}{*}{JPEG}  \cite{wallace1992jpeg} & 0.174    & 25.1   & 0.685  & 86.6  & \centering 79.1   & \centering 52.3      & 0.354 & 27.6   & 0.804& 94.2 &  \centering91.8 &  \centering 77.1  & 0.408  & 32.5 & 0.895 &  94.9 &  \centering92.9 &  81.1  \\  
\multirow{1}{*}{WebP} \cite{mukherjee2014webp} &  0.201  &  28.3       & 0.774  &93.2  &\centering 89.4 &  \centering 64.1     & 0.317 & 30.3   & 0.869 & 95.1 & \centering92.7  & \centering 77.3  &0.404 &34.8&0.927&95.2 &\centering93.4 &81.0  \\ 
\multirow{1}{*}{x265} \cite{ramachandran2013x265} &0.210    &29.1       &0.766  &77.4&  \centering 74.9& \centering 53.5     &0.283& 31.2   & 0.851 & 94.2 & \centering91.4  & \centering 74.8  & 0.410 & 33.1 & 0.889 & 94.9   &\centering92.7  & 78.9\\ 
\multirow{1}{*}{IRN} \cite{xiao2023invertible} &0.180    & 29.9      & 0.787  & 90.6  & \centering 84.9 & \centering 57.5      &0.264& 32.2   & 0.896 & 94.7 & \centering92.5  & \centering 77.1 & 0.387 & 32.84 & 0.906 & 94.8 &  \centering92.6    &77.8   \\ 
\multirow{1}{*}{VTM} \cite{bross2021overview} &0.239   &  33.9     & 0.872 & 93.7      &\centering 90.4 & \centering 71.2      &0.272& 35.9   & 0.939&95.3 & \centering \textbf{93.5}  & \centering 81.3 &0.412 & 36.5 & 0.948 &  95.2 & \centering93.6 & 82.5    \\ 
\multirow{1}{*}{HifiC} \cite{mentzer2018conditional} &0.177  &  32.8     &  0.821  &  94.4 & \centering 91.7  &  \centering 72.1    & 0.264 & 34.3  &  0.903 & \centering95.2 & \centering93.3  & \centering 80.3  & 0.387 & 35.2  & 0.937 & 94.9  &\centering92.6 & 77.8\\
\multirow{1}{*}{Ours} &0.154  & 24.0     & 0.688  &\centering \textbf{94.8} & \centering \textbf{92.1}      & \centering \textbf{75.1} & 0.264 & 26.1 & 0.759  & 95.3 & \centering93.2  & \centering \textbf{81.7} & 0.394 & 28.6 & 0.839 & \textbf{95.3} & \centering {\textbf{93.7}}  & \textbf{82.7}     \\
\multirow{1}{*}{Original} &6.237 & -      & -   &95.4 &\centering 94.0 & \centering 84.4     & - & -   & - & -& -  & -  & - & -   & - & -& \centering-  & - \\
\hline
\end{tabularx}
\label{table3}
\end{table*}
\section{Computational Complexity}
In this section, we present details regarding the computational complexity of model components and the sampling cost.
\subsection{Computational Complexity of U-Net} The benchmarking in Table \ref{table7} was conducted on an Nvidia 3080 Ti with CUDA 11.1. We performed 1,000 forward passes on square inputs of size 256 $\times$ 256 and 1024 $\times$ 1024, and recorded the inference time using CUDA events. However, we excluded the time for entropy coding and Hierarchical K-means ($O(n^2 \log n)$) from our analysis.
\subsection{Forward Complexity}
Encoder time = Encoder + Hierarchical K-means $O(n^2 log n)$ for different bpp.

Mathematically, we can directly write out the expression for the t-th step in the forward propagation process to save time: 
\begin{equation}
\begin{aligned}
   z_{t-1} &= z_t - C(R(z_t,t),t) +C(R(z_t,t),t-1) \\
    &= D(z_0,t) -C(R(z_t,t)) + C(R(z_t)t-1) \\
    &= z_0 + s\cdot e-R(x_t,t)- s\cdot e +R(z_t,t) +(s-1)e \\
    &=  z_0 + (s-1)e \\
    &=D(z_0, S-1),
\end{aligned}
\end{equation}
here, $z$ represents the features, $t$ represents the timestep, $R$ represents the recovery operator, $C$ represents the Degradation operator (Hierarchical K-means with $O(n^2 log n)$ complexity), and it satisfies the condition $C(z, 0) = z$. 
\subsection{Backward Complexity}
Backward complexity (Decoder time): Due to the iterative sampling mechanism of diffusion, the decoding time is negatively correlated with the bit rate and linearly positively correlated with the sampling step length. The decoder time of different DDPMs equals to (U-Net inference time) $\times$ sample step.
\subsection{Fast Sampling of Diffusion}
While the current technical approach suffers from the common issue of time consumption, we would like to emphasize that despite the inherent computational demands of diffusion models, we still find them highly applicable and effective in our study. Although effective sampling is not our main focus, researchers are actively exploring ways to sample effectively in diffusion models.

\noindent \textbf{Knowledge Distillation.} Approaches that use knowledge distillation can significantly improve the sampling speed of diffusion models. Specifically, in Progressive Distillation, the authors \cite{salimans2022progressive, luhman2021knowledge, meng2023distillation} propose distilling the full sampling process into a faster sampler that requires only half as many steps.

\noindent \textbf{Truncated Diffusion.} 
One can improve sampling speed by truncating the forward and reverse diffusion processes \cite{lyu2022accelerating, zheng2022truncated}. The key idea is to halt the forward diffusion process early on, after just a few steps, and to begin the reverse denoising process with a non-Gaussian distribution.

\noindent \textbf{Optimized Discretization.}  Given a pre-trained diffusion model, optimized discretization approaches \cite{watson2021learning, dockhorn2022genie} put forth a strategy for finding the optimal discretization scheme by selecting the best time steps to maximize the training objective for DDPMs.

\noindent \textbf{Engineering Techniques.} Finally, engineering techniques \cite{hong2020efficient, balle2018integer} aimed at lightening network models, such as fixed-point inference and integer quantization, may help alleviate issues related to time consumption.
\begin{table}[h!]
\clearpage
\small
\caption{Parameter count and time consumption of diffusion U-Net}
\centering
\begin{tabularx}{\linewidth}{*{1}{>{\centering\arraybackslash}X} | *{1}{>{\centering\arraybackslash}X}|*{1}{>{\centering\arraybackslash}X}} 
 \hline
Parameter count  & Runtime 256 (ms)	 & Runtime 1024 (ms)\\[0.5ex] 
 \hline
 \hline
 108.4M & 19.3 & 132.5   \\  
 \hline
\end{tabularx}
\label{table7}
\end{table}
\end{document}